\documentclass[10pt,twocolumn,letterpaper]{article}

\usepackage[pagenumbers]{cvpr}

\usepackage{microtype}
\renewcommand{\paragraph}[1]{\vspace{.5em}\noindent\textbf{#1.}}
\usepackage{nicefrac}
\usepackage{multicol, multirow}
\usepackage[table]{xcolor}
\definecolor{1st}{RGB}{102,194,164}
\definecolor{2nd}{RGB}{178,226,226}
\definecolor{3rd}{RGB}{217,248,241}
\definecolor{1stText}{RGB}{57,146,116}

\usepackage{dsfont}
\usepackage{enumitem}
\usepackage{bm}
\usepackage{cuted}
\usepackage{tikz}
\tikzset{
  ovlabel/.style={inner sep=1.5pt, font=\scriptsize,
                  text=white, fill=black, fill opacity=0.5, text opacity=1},
  ovpanel/.style={inner sep=0, outer sep=0, anchor=south west},
}
\newcommand{\FLIP}{\protect\reflectbox{F}LIP\xspace}

\definecolor{cvprblue}{rgb}{0.21,0.49,0.74}
\usepackage[pagebackref,breaklinks,colorlinks,allcolors=cvprblue]{hyperref}

\title{Compact Neural Appearance Models for Efficient Gaussian Splatting}
\author{
Florian Hahlbohm$^{1}$ \quad Jorge Condor$^{2}$ \quad Linus Franke$^{3}$ \quad Martin Eisemann$^{1}$ \quad Marcus Magnor$^{1, 4}$\\
\small
$^1$TU Braunschweig \quad $^2$Università della Svizzera italiana \quad $^3$University of Würzburg \quad $^4$University of New Mexico\\
\small\textbf{\url{https://fhahlbohm.github.io/efficient-gaussian-appearance}}
}

\begin{document}
\maketitle
\begin{abstract}
Explicit primitive-based radiance fields such as 3D Gaussian Splatting typically model view-dependent appearance using low-order spherical harmonics (SH).
Although efficient to evaluate, SH coefficients dominate per-primitive storage and memory traffic, while their band-limited basis restricts angular detail.
We present a thorough, end-to-end comparison of SH and recent spherical appearance models and introduce an implicit alternative that decodes compact per-primitive latent codes using a tiny shared MLP.
We integrate all models into the same optimized pipeline, fusing their forward and backward passes into a differentiable CUDA rasterizer and provide a portable WebGL viewer for laptop and mobile GPUs.
Our evaluation across reconstruction quality, memory use, and optimization and rendering performance shows that recent spherical models offer the strongest overall quality--efficiency trade-off.
Our neural representation is the most compact model evaluated and, compared to third-degree SH, reduces the per-primitive appearance footprint from 192 to 28 bytes, accelerates optimization by 1.3\texttimes, while improving reconstruction quality.
We further analyze how appearance parametrization shapes optimization, identifying differences in recovered geometry and the tendency of expressive models to absorb non-static scene content.
Together, our framework and analysis provide practical guidance for replacing SH beyond what image metrics alone can capture.
\end{abstract}

\section{Introduction}
\label{sec:introduction}
Radiance fields have driven rapid progress in 3D reconstruction and novel view synthesis, with applications in robotics, virtual and augmented reality, and interactive graphics.
Early methods represented scenes primarily with neural networks~\cite{mildenhall2020nerf}, while subsequent work increasingly adopted explicit representations for more efficient optimization and rendering~\cite{hedman2021snerg, yu2021plenoxels, mueller2022instant, Chen2022tensorf, barron2023ICCV, hahlbohm2024inpc}.
This trend culminated in 3D Gaussian Splatting (3DGS)~\cite{kerbl3Dgaussians}, whose explicit anisotropic primitives and differentiable rasterizer combine high reconstruction quality with real-time rendering.
3DGS has since become the blueprint for many primitive-based radiance field methods~\cite{huang20242d, mai2024ever, held2025convexsplatting, govindarajan2025radfoam, von2025linprim, held2025trianglesplatting, condor2026nht}.

Across these representations, modeling view-dependent appearance remains essential for high-quality novel view synthesis.
3DGS uses low-order spherical harmonics (SH), which are inexpensive to evaluate, straightforward to optimize, and support a coarse-to-fine schedule that suppresses view dependence early in training~\cite{kerbl3Dgaussians}.
However, third-degree SH requires 48 coefficients per primitive, constituting most parameters stored by each Gaussian~\cite{3DGSzip2024}.
Appearance thus dominates model size, optimizer state, and memory traffic during optimization and rendering.
At high primitive counts, updating SH coefficients alone accounts for up to 50\% of optimization time~\cite{hahlbohm2026fastergs}.
Moreover, low-degree SH are band-limited and therefore unable to represent high-frequency angular appearance.

Recent work has proposed replacing SH with more expressive spherical models based on beta kernels~\cite{liu2025betasplatting}, soft Voronoi diagrams~\cite{disario2026sphericalvoronoi}, as well as anisotropic spherical Gaussian or Gabor kernels~\cite{miazga2026nasgabor}.
Although these methods report better quality at equal or lower parameter counts, their practical trade-offs remain difficult to assess without a controlled evaluation under a common optimization protocol, consistently optimized end-to-end implementations, and measurements across desktop and mobile GPUs.
Existing evaluations also focus primarily on image metrics, which reveal little about how appearance parametrization affects the geometry recovered during optimization.

We therefore present a controlled, end-to-end comparison of view-dependent appearance models for Gaussian splatting.
We integrate SH and recent spherical alternatives into Faster-GS~\cite{hahlbohm2026fastergs}, using runtime code generation~\cite{tiny-cuda-nn} to JIT-compile each model's forward and backward passes into its differentiable rasterizer, retaining extensibility while avoiding separate kernel launches and global memory round-trips.
We further implement all evaluated models in a portable WebGL viewer, extending the comparison to laptop and mobile GPUs, where appearance evaluation faces different computational and memory constraints.

\begin{figure*}[t]
\centering
\includegraphics[width=1.0\linewidth]{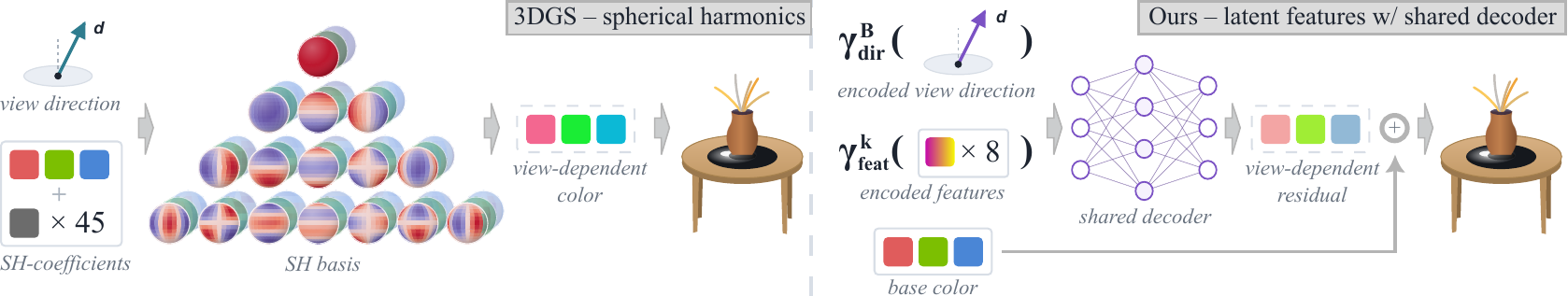}
\caption{%
3DGS~\cite{kerbl3Dgaussians} employs third-degree spherical harmonics, totaling 45 view-dependent appearance parameters per Gaussian (left). Our model (right) achieves faster training$\slash$rendering with similar quality using only 8 features per Gaussian and a tiny MLP with shared weights.
}\label{fig:sh_vs_mlp}
\end{figure*}

We also introduce a basis-free alternative that decodes compact per-primitive latent codes using a tiny direction-conditioned MLP shared across the scene (\cf \cref{fig:sh_vs_mlp}).
Unlike explicit spherical models, the decoder is not restricted to a predefined basis.
This approach has received little attention in explicit radiance fields, likely due to the belief that network evaluation is too expensive in real-time rendering pipelines.
We show that this assumption no longer holds on modern consumer GPUs: the fused decoder executes efficiently on Tensor Cores, while on mobile GPUs, inexpensive FMA operations and the cache locality of weights shared across primitives yield competitive rendering performance.

Our evaluation reveals distinct trade-offs among the considered models: recent spherical representations provide strong overall quality and efficiency compared to classical spherical harmonics. Our neural representation is, on the other hand, the most compact, and comparable to them in speed and quality.
We further analyze how appearance parametrization shapes optimization, identifying differences in recovered geometry and a tendency of expressive models to absorb non-static scene content.
In summary, we contribute:
\begin{itemize}
\item a controlled comparison of SH and recent parametric appearance models in a common optimized pipeline, spanning quality, memory use, optimization, and rendering,
\item a compact, basis-free neural appearance model based on per-primitive latent codes and a shared MLP, fused into the differentiable rasterizer, and
\item an analysis of how appearance parametrization shapes optimization, including its effects on recovered geometry and the absorption of non-static scene content.
\end{itemize}
Our full implementation is publicly available.
Its appearance models and neural architecture remain configurable beyond the settings evaluated in this paper, facilitating future work on alternative representations and platform-specific quality--efficiency trade-offs.

\section{Related Work}
\label{sec:related_work}

\paragraph{Radiance Field Representations}
Radiance field methods span a spectrum from implicit neural fields to fully explicit scene representations.
NeRF~\cite{mildenhall2020nerf} and its derivatives~\cite{barron2022mipnerf360} use MLPs to represent density and view-dependent radiance, while hybrid approaches accelerate optimization and rendering using explicit spatial structures such as hash grids~\cite{mueller2022instant, barron2023ICCV}, voxel grids~\cite{sun2022dvgo}, or factorized tensors~\cite{Chen2022tensorf}.
Fully explicit radiance fields eliminate the neural scene representation entirely~\cite{yu2021plenoxels, kerbl3Dgaussians}.
In particular, 3D Gaussian Splatting (3DGS)~\cite{kerbl3Dgaussians} established differentiable rasterization of explicit primitives as an efficient paradigm for reconstruction and rendering, inspiring variants based on surfels~\cite{huang20242d}, triangles~\cite{held2025trianglesplatting, held2026meshsplatting}, convexes~\cite{held2025convexsplatting}, tetrahedra~\cite{von2025linprim, mai2026radmesh}, and Voronoi cells~\cite{govindarajan2025radfoam}, as well as ray-traced alternatives~\cite{condor2025dontsplat, mai2024ever, moenne2024gaussiantracer}.
Complementary compression methods reduce the storage of these representations by pruning, quantizing, or vector-quantizing Gaussian parameters, including their SH coefficients~\cite{3DGSzip2024}.
Our work instead studies which appearance parametrization should be used in the first place, retaining the explicit geometry and rasterization pipeline of 3DGS.

\paragraph{View-Dependent Appearance}
Explicit radiance fields commonly model angular appearance using spherical functions.
Plenoxels~\cite{yu2021plenoxels} and 3DGS~\cite{kerbl3Dgaussians} employ spherical harmonics (SH), whose inexpensive evaluation and simple coarse-to-fine optimization have made them the de facto standard.
Recent work proposes more expressive alternatives based on von Mises--Fisher distributions~\cite{blanc2025raygauss}, beta kernels~\cite{liu2025betasplatting}, soft Voronoi diagrams~\cite{disario2026sphericalvoronoi}, as well as anisotropic spherical Gaussian and Gabor kernels~\cite{miazga2026nasgabor}.
Reflection-oriented methods instead introduce specialized parametrizations based on reflected or secondary rays to model challenging specular appearance~\cite{verbin2022refnerf, verbin2024nerfcasting, xie2025envgs, poirier2025editable}.
These methods primarily target novel-view quality, whereas we compare general-purpose appearance models under a common optimization and implementation framework, including their memory and rendering costs across desktop and mobile GPUs.
We additionally study how their parametrization and expressiveness affect optimization beyond image metrics.

\paragraph{Neural Appearance Evaluation}
Neural appearance models differ in where they are evaluated within the rendering pipeline.
Deferred neural rendering~\cite{thies2019deferred} decodes appearance in image space after rasterization using CNNs or per-pixel MLPs~\cite{hedman2021snerg, chen2023mobilenerf, reiser2023merf, aliev2020npbg, kopanas2021perviewopt, ruckert2022adop, franke2024trips, hahlbohm2024inpc, condor2026nht}, tying evaluation cost to rendering resolution.
Forward neural rendering instead evaluates appearance at sampled 3D positions before compositing~\cite{mildenhall2020nerf, mueller2022instant, Chen2022tensorf}.
Scaffold-GS~\cite{lu2024scaffoldgs} similarly decodes primitive attributes from compact latent features before rasterization.
In contrast, our neural appearance model follows the forward paradigm but retains the standard per-primitive 3DGS representation, decoding only view-dependent color once per visible Gaussian as a direct replacement for spherical appearance models.
To achieve practical frame rates, we rely on efficient on-GPU inference: tiny-cuda-nn~\cite{tiny-cuda-nn} introduced fully fused networks and runtime code generation, enabling compact MLPs to be inlined into CUDA kernels for applications ranging from radiance caching~\cite{mueller2021neuralpath} and layered material models~\cite{zeltner2024realtimeneural} to visibility prediction~\cite{zoomers2026nvgs}.
Building on these developments, we integrate the forward and backward passes of our neural appearance model directly into a differentiable Gaussian rasterizer.

\section{Preliminaries}
\label{sec:method}
We will first provide an overview of the standard view dependent appearance model in 3D Gaussian Splatting, and then further expand to recently proposed alternatives.

\subsection{Appearance Modelling in Gaussian Splatting}
\label{sec:appearance_models}
In 3D Gaussian Splatting (3DGS)~\cite{kerbl3Dgaussians}, a scene is represented by a set of anisotropic Gaussians with position, covariance, opacity, and view-dependent color.
Rendering proceeds in two stages.
A per-primitive \emph{preprocess} kernel projects each Gaussian onto the image plane, determines the overlapped 16\texttimes16 pixel tiles, and computes its view-dependent color.
A subsequent tile-based \emph{blend} kernel alpha-composites the Gaussians in each pixel in depth-sorted order.

\paragraph{Spherical Harmonics}
Standard 3DGS represents appearance with spherical harmonics (SH).
With the normalized direction $\mathbf{d}$ from the camera center to a Gaussian mean and RGB coefficients $\mathbf{c}_{\ell}^{m}\in\mathbb{R}^{3}$, the view-dependent color is
\begin{equation}
\label{eq:sh_color_eval}
\mathbf{c}(\mathbf{d}) = \operatorname{ReLU}\Big(
\underbrace{0.5 + Y_0\, \mathbf{c}_0}_{\text{base color } \mathbf{c}_\mathrm{b}}
+ \underbrace{\textstyle\sum_{\ell=1}^{L} \sum_{m=-\ell}^{\ell} Y_\ell^m(\mathbf{d})\, \mathbf{c}_\ell^m}_{\text{view-dependent residual } \mathbf{c}_\mathrm{r}}
\Big).
\end{equation}
Training begins with $L=0$ and activates one additional band every 1000 iterations, usually up to $L=3$.
Degree-3 SH requires 48 scalar coefficients per Gaussian, of which 45 describe view dependence.
The degree-zero coefficient is initialized from the SfM point color $\mathbf{c}_{\mathrm{sfm}}\in[0,1]^3$ as $\mathbf{c}_0=(\mathbf{c}_{\mathrm{sfm}}-0.5)/Y_0$, while the remaining coefficients are initialized to zero.
For a controlled comparison, we express all representations through a common interface
\begin{equation}
\label{eq:common_appearance}
\mathbf{c}(\mathbf{d}) = \phi\left(\mathbf{c}_0 + \mathcal{A}_{\psi}(\mathbf{d})\right),
\end{equation}
where $\mathcal{A}_{\psi}$ is the view-dependent appearance model and $\phi$ the color activation, including any constant shift.
For standard 3DGS, $\phi(\hat{\mathbf{c}}) = \operatorname{ReLU}(\hat{\mathbf{c}}+0.5)$ and $\mathcal{A}_{\psi}$ is the SH expansion over degrees $\ell\geq1$.
We drop the constant factor $Y_0$ from the forward pass, initializing $\mathbf{c}_0=\phi^{-1}(\mathbf{c}_{\mathrm{sfm}})$ and multiplying its learning rate by $Y_0$ instead.
Since the commonly used optimizer Adam~\cite{kingma2014adam} normalizes the gradient magnitude per parameter, both formulations take practically the same step in color space.
All models retain the base color exactly as in the baseline, including its initialization and densification handling, and replace only $\mathcal{A}_{\psi}$.
The color activation $\phi$ also affects optimization independently of $\mathcal{A}_{\psi}$.
Beyond the baseline ReLU, natural alternatives include a smooth rectifier $\phi(\hat{\mathbf{c}})=\operatorname{softplus}(\hat{\mathbf{c}}+0.5)$ ($\beta=10$)~\cite{mai2024ever}, a scaled sigmoid $\phi(\hat{\mathbf{c}})=\sigma(4\hat{\mathbf{c}})$, and a hard sigmoid $\phi(\hat{\mathbf{c}})=\operatorname{clamp}(\hat{\mathbf{c}}+0.5,\,0,\,1)$.
All four map zero to $0.5$, the midpoint of the target RGB range, and are scaled to closely match the baseline ReLU near this operating point.
All view-dependent parameters are initialized to produce a zero residual.
Unless stated otherwise, we use the baseline ReLU for all appearance models in this work.
However, as we will show in our experiments, the choice of color activation has a significant effect on optimization dynamics.

\subsection{Alternative Explicit Models}
\paragraph{Spherical Voronoi}
Spherical Voronoi (SV)~\cite{disario2026sphericalvoronoi} represents appearance as a soft interpolation over a set of sites $\{\mathbf{s}_i\}$ on the sphere with associated RGB values $\{\mathbf{a}_i\}$,
\begin{equation}
\label{eq:spherical_voronoi}
\mathcal{A}_{\psi}(\mathbf{d})
=
\sum_i
\frac{\exp\left(-\tau_i\rho(\mathbf{d},\mathbf{s}_i)\right)}
{\sum_j\exp\left(-\tau_j\rho(\mathbf{d},\mathbf{s}_j)\right)}
\,\mathbf{a}_i,
\end{equation}
where $\rho(\mathbf{d},\mathbf{s}_i)$ is the distance between the viewing direction and site $i$, and $\tau_i$ controls the sharpness of its partition.
Each site is parametrized by $(\mathbf{a}_i, \mathbf{s}_i, \tau_i) \in \mathbb{R}^{7}$.
Since the softmax assigns a single site unit weight regardless of direction, at least two are needed for view-dependent variation.
Note that the reference implementation does not separate a base color, but since the weights sum to unity, a shared base color can be factored out of the site values, a property that we use to align SV with the common interface in \cref{eq:common_appearance}.

\paragraph{Normalized Anisotropic Spherical Gaussians}
Normalized anisotropic spherical Gaussians (NASG), originally introduced for neural path guiding~\cite{huang2024nasg} and adapted to primitive-based reconstruction by Miazga~\etal~\cite{miazga2026nasgabor}, represent appearance as a weighted sum of spherical kernels,
\begin{equation}
\label{eq:nasg}
\mathcal{A}_{\psi}(\mathbf{d})
=
\sum_i G_i(\mathbf{d})\,\tanh(\mathbf{a}_i),
\end{equation}
where each $G_i$ has a learned orientation $\in\mathbb{R}^3$ and anisotropic angular extent $\in\mathbb{R}^2$, and $\mathbf{a}_i\in\mathbb{R}^3$ is its RGB amplitude, resulting in eight parameters per lobe.
Normalization decouples a kernel's contribution from its angular extent, while anisotropy allows different rates of variation along its two principal tangent directions.
The $\tanh$ bounds each lobe's color contribution, constraining the residual to a range compatible with the base color in \cref{eq:common_appearance}.
Miazga~\etal further extend $G_i$ with a non-negative cosine carrier, yielding NASGabor with kernel $\tilde{G}_i$ and one additional frequency parameter per lobe.
The carrier allows a single kernel to represent a multimodal signal while maintaining a compact memory footprint.

For all explicit models, we follow the parametrizations and initialization of the respective reference works, adapting them to the common interface in \cref{eq:common_appearance}.
We provide more detailed definitions and discuss all implementation-level differences in \cref{sec:suppl_appearance_model_details,sec:suppl_implementation_details}.

\section{Neural View-Dependent Appearance}
\label{sec:neural_appearance}
In contrast to explicit, parametric models, which impose constraints or assumptions on the angular signal to be modelled in order to achieve their compactness, we replace the view-dependent parameters with a per-Gaussian latent code and decode appearance with a small network shared across the entire scene, on a per-primitive basis.

Each Gaussian stores a feature vector $\mathbf{f}\in\mathbb{R}^{F}$, and a shared MLP with parameters $\theta$ maps $\mathbf{f}$ and the viewing direction to an RGB residual.
Its input is
\begin{equation}
\label{eq:mlp_input}
\mathbf{x}
=
\left(
\gamma_{\mathrm{dir}}^{\mathcal{B}}(\mathbf{d}),
\gamma_{\mathrm{feat}}^{k}(\mathbf{f})
\right)
\in\mathbb{R}^{D},
\end{equation}
where $\gamma_{\mathrm{dir}}^{\mathcal{B}}$ evaluates the SH basis functions of degrees $\mathcal{B}$ and $\gamma_{\mathrm{feat}}^{k}$ applies a frequency encoding~\cite{mildenhall2020nerf} independently to each feature component.
For a scalar $f$, the latter is
\begin{equation}
\gamma_{\mathrm{feat}}^{k}(f)
=
\left(
\sin(2^j f),\cos(2^j f)
\right)_{j=0}^{k-1}.
\end{equation}
How the network inputs are split between the two encodings is the central design choice.
The direction encoding is computed on the fly and costs no storage, but determines the \textit{angular resolution} of the view dependence.
The features are stored per Gaussian and determine the \textit{conditioning capacity}, \ie, how much the shared decoder's response can differ between primitives: a small $F$ forces all Gaussians into a low-dimensional code space, while a large $F$ makes the model behave more like classic per-primitive coefficients at a linear storage cost.
Efficient evaluation on Tensor Cores requires the network input to be a multiple of 16.
Since the degree-3 direction encoding inherited from the baseline already occupies 16 inputs, $D=32$ is the smallest viable input size that accommodates per-Gaussian features.
The remaining 16 dimensions encode $F=8$ feature values with $k=1$.
The decoder is a fully connected MLP with two hidden layers of 16 neurons, three outputs, ReLU hidden activations, and no biases.
Bounding its predicted residual with a $\tanh$, we compute
\begin{equation}
\label{eq:neural_color_eval}
\mathcal{A}_{\psi}(\mathbf{d})
=
\tanh\left(
\operatorname{MLP}_{\theta}(\mathbf{x})
\right).
\end{equation}
The bounded residual discourages the decoder from compensating for an arbitrarily displaced base color.
To retain the baseline's progressive optimization schedule, we initially disable the residual and enable the directional encoding bands according to the SH schedule.
Inactive inputs are set to zero and consequently receive no gradient.
The model stores three base-color values and eight latent features per Gaussian, while the shared MLP adds 816 parameters for the entire scene.
Moreover, since both the forward and backward pass carry out all computations in half precision, MLP weights and features can be deployed in 16-bit without quality loss.
With 32-bit base colors, the deployed appearance footprint is 28 bytes per Gaussian, compared with 192 bytes for 48 single-precision SH coefficients.

\paragraph{CUDA Implementation}
\label{sec:cuda_implementation}
We expose all appearance models through a common device interface whose forward and backward evaluations are inlined into the differentiable CUDA rasterizer.
Using the runtime code-generation facilities of tiny-cuda-nn~\cite{tiny-cuda-nn}, the interface is specialized for each model and JIT-compiled with NVRTC on first use.
This allows the appearance model to be selected via configuration without sacrificing compile-time optimizations, and makes the framework readily extensible to new representations while retaining the state-of-the-art training performance of Faster-GS~\cite{hahlbohm2026fastergs}.
Appearance is evaluated only for primitives that survive culling, directly within the preprocess kernel, requiring no separate pass or global-memory round trip.
Directional gradients from the appearance evaluation are propagated to the Gaussian positions, preserving the appearance-dependent positional gradients used by the baseline's densification procedure.
We additionally retain a reference path that evaluates appearance in a separate pass and use it to validate each fused implementation.
The remainder of Faster-GS, including tile assignment, sorting, blending, densification, and pruning, is unchanged.
We provide implementation details, including optimization hyperparameters, in \cref{sec:suppl_appearance_model_details,sec:suppl_implementation_details}.

\begin{table*}[t]
\caption{%
Quantitative results on established benchmark datasets using MCMC densification~\cite{kheradmand243dgsmcmc}. We employ PPISP~\cite{deutsch2026ppisp} on real scenes to handle photometric variations and measure frames per second (FPS) at the native scene resolution. Since image metrics vary between runs, we report average values across five runs in this table to improve reproducibility.
}\label{tab:training}
\centering
\setlength\tabcolsep{1.6pt}
\scriptsize
\begin{tabular}{lccccccccccccccccccc}
\toprule
 & & \multicolumn{6}{c}{Mip-NeRF~360~\cite{barron2022mipnerf360}} & \multicolumn{6}{c}{Tanks \& Temples~\cite{Knapitsch2017}~$\slash$~Deep Blending~\cite{hedman2018deep}} & \multicolumn{6}{c}{NeRF Synthetic~\cite{mildenhall2020nerf}} \\
\cmidrule(lr){3-8} \cmidrule(lr){9-14} \cmidrule(lr){15-20}
Basis & Bytes$\slash$G & PSNR$^\uparrow$ & SSIM$^\uparrow$ & LPIPS$^\downarrow$ & Train$^\downarrow$ & VRAM$^\downarrow$ & FPS$^\uparrow$ & PSNR$^\uparrow$ & SSIM$^\uparrow$ & LPIPS$^\downarrow$ & Train$^\downarrow$ & VRAM$^\downarrow$ & FPS$^\uparrow$ & PSNR$^\uparrow$ & SSIM$^\uparrow$ & LPIPS$^\downarrow$ & Train$^\downarrow$ & VRAM$^\downarrow$ & FPS$^\uparrow$\\ 
\midrule
None     & 12  & 27.73 & 0.825 & 0.234 & 3m32s & 4.6GiB & 443.1 & 27.65 & 0.888 & 0.237 & 2m17s & 3.3GiB & 763.3 & 31.46 & 0.959 & 0.048 & 57s   & 1.3GiB & 1734 \\
\midrule
SH       & 192 & 28.36 & \cellcolor{1st}0.836 & \cellcolor{1st}0.222 & 6m12s & 6.7GiB & 378.9 & 27.96 & \cellcolor{1st}0.893 & \cellcolor{2nd}0.228 & 4m10s & \cellcolor{3rd}4.9GiB & 683.2 & \cellcolor{2nd}34.01 & \cellcolor{1st}0.971 & \cellcolor{1st}0.036 & \cellcolor{3rd}1m22s & \cellcolor{2nd}1.5GiB & 1696 \\
SV       & 208 & \cellcolor{1st}28.64 & \cellcolor{2nd}0.834 & \cellcolor{3rd}0.225 & 6m30s & 6.8GiB & 385.0 & \cellcolor{2nd}27.98 & \cellcolor{1st}0.893 & \cellcolor{3rd}0.232 & 4m20s & 5.0GiB & 696.6 & \cellcolor{3rd}33.45 & \cellcolor{3rd}0.968 & \cellcolor{2nd}0.038 & 1m28s & \cellcolor{3rd}1.6GiB & 1576 \\
NASG     & \cellcolor{2nd}44  & 28.50 & \cellcolor{1st}0.836 & \cellcolor{2nd}0.223 & \cellcolor{1st}3m54s & \cellcolor{1st}4.9GiB & \cellcolor{2nd}439.9 & \cellcolor{3rd}27.97 & \cellcolor{1st}0.893 & \cellcolor{1st}0.227 & \cellcolor{1st}2m36s & \cellcolor{1st}3.6GiB & \cellcolor{1st}757.4 & 32.69 & 0.966 & 0.043 & \cellcolor{1st}59s   & \cellcolor{1st}1.3GiB & \cellcolor{3rd}1826 \\
NASGabor & \cellcolor{3rd}48  & \cellcolor{3rd}28.54 & \cellcolor{1st}0.836 & \cellcolor{2nd}0.223 & \cellcolor{2nd}4m00s & \cellcolor{2nd}5.0GiB & \cellcolor{3rd}436.9 & 27.96 & \cellcolor{1st}0.893 & \cellcolor{1st}0.227 & \cellcolor{2nd}2m39s & \cellcolor{1st}3.6GiB & \cellcolor{3rd}743.6 & 32.95 & 0.967 & \cellcolor{3rd}0.041 & \cellcolor{1st}59s   & \cellcolor{1st}1.3GiB & \cellcolor{2nd}1852 \\
Neural   & \cellcolor{1st}28  & \cellcolor{2nd}28.55 & \cellcolor{3rd}0.831 & 0.227 & \cellcolor{3rd}4m17s & \cellcolor{3rd}5.3GiB & \cellcolor{1st}443.6 & \cellcolor{1st}28.13 & \cellcolor{1st}0.893 & \cellcolor{2nd}0.228 & \cellcolor{3rd}2m56s & \cellcolor{2nd}3.9GiB & \cellcolor{2nd}743.8 & \cellcolor{1st}34.04 & \cellcolor{2nd}0.970 & \cellcolor{2nd}0.038 & \cellcolor{2nd}1m02s & \cellcolor{1st}1.3GiB & \cellcolor{1st}1869 \\
\bottomrule
\end{tabular}
\end{table*}

\begin{table*}[t]
\caption{%
Average frame time at 720p in milliseconds across scenes, devices, and rendering backends. CPU-side depth sorting is excluded. For WebGL, we load the same checkpoint for all models, \ie, geometry is identical, assigning dummy values for the appearance parameters, and exclude the CPU-side depth sorting as well as application of the PPISP module. Values rounded roughly \wrt to measurement precision.
}\label{tab:inference}
\centering
\setlength\tabcolsep{2.2pt}
\scriptsize
\begin{tabular}{llccccccccccccccc}
\toprule
 & & \multicolumn{5}{c}{Bicycle (6M Gaussians)} & \multicolumn{5}{c}{Stump (4.5M Gaussians)} & \multicolumn{5}{c}{Bonsai (1.5M Gaussians)} \\
\cmidrule(lr){3-7} \cmidrule(lr){8-12} \cmidrule(lr){13-17}
 & Device & ~~~SH~~~ & ~~~SV~~~ & ~NASG~~~ & NASGabor & ~Neural~ & ~~~SH~~~ & ~~~SV~~~ & ~NASG~~~ & NASGabor & ~Neural~ & ~~~SH~~~ & ~~~SV~~~ & ~NASG~~~ & NASGabor & ~Neural~ \\
\midrule
\multirow{3}{*}{\rotatebox{90}{CUDA}}
 & Desktop (RTX 5090) & 2.11 & 2.16 & \cellcolor{1st}1.85 & \cellcolor{2nd}1.87 & \cellcolor{3rd}1.97 & 2.27 & 2.29 & \cellcolor{3rd}2.11 & \cellcolor{2nd}2.00 & \cellcolor{1st}1.98 & 1.46 & 1.42 & \cellcolor{2nd}1.32 & \cellcolor{1st}1.28 & \cellcolor{3rd}1.36 \\
 & Desktop (RTX 4090) & 2.75 & 2.80 & \cellcolor{1st}2.25 & \cellcolor{2nd}2.26 & \cellcolor{3rd}2.38 & 2.95 & 3.00 & \cellcolor{3rd}2.55 & \cellcolor{2nd}2.43 & \cellcolor{1st}2.36 & 1.76 & 1.71 & \cellcolor{2nd}1.51 & \cellcolor{1st}1.46 & \cellcolor{3rd}1.55 \\
 & Desktop (RTX 3090) & 4.27 & 4.36 & \cellcolor{1st}3.82 & \cellcolor{2nd}3.83 & \cellcolor{3rd}4.06 & 4.54 & 4.66 & \cellcolor{3rd}4.28 & \cellcolor{2nd}4.06 & \cellcolor{1st}4.02 & 2.90 & 2.85 & \cellcolor{2nd}2.66 & \cellcolor{1st}2.60 & \cellcolor{3rd}2.75 \\
\midrule
\multirow{4}{*}{\rotatebox{90}{WebGL}}
 & Desktop (RTX 4090) & \cellcolor{1st}4.0  & \cellcolor{2nd}4.5  & \cellcolor{1st}4.0  & \cellcolor{1st}4.0  & \cellcolor{1st}4.0  & \cellcolor{1st}3.6  & \cellcolor{3rd}4.0  & \cellcolor{1st}3.6  & \cellcolor{1st}3.6  & \cellcolor{2nd}3.9  & \cellcolor{2nd}1.3  & \cellcolor{3rd}1.4  & \cellcolor{1st}1.2  & \cellcolor{2nd}1.3  & \cellcolor{1st}1.2  \\
 & MacBook (M1 Pro)   & \cellcolor{2nd}12.2 & \cellcolor{3rd}13.1 & \cellcolor{1st}12.0 & \cellcolor{1st}12.0 & 13.2 & \cellcolor{2nd}11.1 & 12.1 & \cellcolor{1st}11.0 & \cellcolor{1st}11.0 & \cellcolor{3rd}11.9 & \cellcolor{1st}6.1  & \cellcolor{2nd}6.4  & \cellcolor{1st}6.1  & \cellcolor{1st}6.1  & \cellcolor{2nd}6.4  \\
 & iPad (M2 Pro)      & \cellcolor{2nd}35.9 & 40.0 & \cellcolor{1st}35.6 & \cellcolor{1st}35.6 & \cellcolor{3rd}38.1 & \cellcolor{2nd}31.7 & 34.8 & \cellcolor{1st}31.3 & \cellcolor{1st}31.3 & \cellcolor{3rd}33.4 & \cellcolor{2nd}15.6 & 16.6 & \cellcolor{1st}15.5 & \cellcolor{1st}15.5 & \cellcolor{3rd}16.0 \\
 & iPhone (A18 Pro)   & \cellcolor{2nd}45.0 & OOM  & \cellcolor{1st}44.3 & \cellcolor{1st}44.3 & \cellcolor{3rd}45.1 & \cellcolor{3rd}41.0 & OOM  & \cellcolor{2nd}40.5 & \cellcolor{1st}40.4 & 41.4 & \cellcolor{2nd}17.7 & 19.7 & \cellcolor{1st}17.5 & \cellcolor{1st}17.5 & \cellcolor{3rd}18.5 \\
\bottomrule
\end{tabular}
\end{table*}

\paragraph{WebGL Viewer}
\label{sec:webgl}
To evaluate deployment beyond CUDA-capable hardware, we implement all appearance models in a WebGL viewer built from scratch on Three.js.
Depth sorting proceeds asynchronously on the CPU, matching the approach of state-of-the-art web viewers such as SuperSplat~\cite{supersplat2024}.
At the beginning of each frame, a fragment-shader prepass evaluates the view-dependent color for each Gaussian and writes the result to a texture consumed by the splatting pass, isolating appearance cost from splat blending.
For the established SH model, the view-dependent coefficients are quantized to the compact bit-packed layout used by Spark~\cite{spark2025}, compressing a full degree-3 parametrization to 40 bytes per Gaussian.
The neural model requires no additional quantization: features and MLP weights are stored and evaluated in their native half precision, and the view-independent contribution of the first MLP layer is precomputed per Gaussian at load time.
Note that because the network is shared by all primitives, its weights exhibit high cache locality throughout the pre-pass.
For SV, NASG, and NASGabor, all parameters are stored in half precision and view-independent quantities are precomputed at export time, leaving only the direction-dependent evaluation for the per-frame prepass.
This enables a consistent performance comparison across desktop, laptop, and mobile GPUs.

\section{Evaluation}
\label{sec:evaluation}
Our aim is to provide a thorough evaluation on all possible axes of the appearance modelling problem in primitive-based novel view synthesis: reconstruction quality, VRAM usage, and rendering performance of the implemented appearance models on established benchmarks.
We will further discuss on the specific advantages and disadvantages of explicit \vs implicit models, and finally report WebGL viewer performance for a mobile perspective.

\subsection{Setup}
We evaluate on a total of 21 scenes from Mip-NeRF~360~\cite{barron2022mipnerf360}, Tanks \& Temples~\cite{Knapitsch2017}, Deep Blending~\cite{hedman2018deep}, and NeRF Synthetic~\cite{mildenhall2020nerf}, following the image resolutions and train$\slash$test splits of the original 3DGS paper~\cite{kerbl3Dgaussians}.
All configurations are optimized with Faster-GS~\cite{hahlbohm2026fastergs} using MCMC densification~\cite{kheradmand243dgsmcmc} with established scene-specific primitive counts~\cite{liu2025betasplatting} and identical hyperparameters apart from the appearance model.
In particular, optimization and densification strictly follow the protocol of the baseline~\cite{hahlbohm2026fastergs, kheradmand243dgsmcmc, deutsch2026ppisp}, with no appearance-model-specific tuning of non-appearance parameters.
For SV we use seven sites, and for NASG and NASGabor one lobe, matching the default configurations of the respective reference implementations.
While the reference implementations of SV and NASGabor default to beta kernels~\cite{liu2025betasplatting}, our evaluation uses Gaussian kernels throughout, as they are more widely adopted and we found no meaningful quality advantage with beta kernels.
For detailed comparisons of alternative spherical functions, we refer to Miazga~\etal~\cite{miazga2026nasgabor}.
To prevent more expressive appearance models from absorbing photometric variations in training images as a function of viewing direction, we use per-image signal processing (PPISP)~\cite{deutsch2026ppisp} for the quality comparison.
The WebGL performance evaluation omits PPISP, as evaluating its web deployment is outside the scope of this work.

\newlength{\blkgap}     \setlength{\blkgap}{1pt}   %
\newlength{\blksep}     \setlength{\blksep}{1pt}   %
\newlength{\blklabelw}  \setlength{\blklabelw}{6.5pt}
\newlength{\blkrefextra}\setlength{\blkrefextra}{12pt}
\newlength{\blktmp}     \setlength{\blktmp}{\dimexpr\textwidth-\blklabelw-\blksep-6\blkgap-\blkrefextra\relax}
\newlength{\blkpanelw}  \setlength{\blkpanelw}{0.142857\blktmp}
\newlength{\blkrefw}    \setlength{\blkrefw}{\dimexpr\blkpanelw+\blkrefextra\relax}
\newcommand{\blkimg}[1]{\includegraphics[width=\blkpanelw]{resources/bonsai_block/#1}}
\newlength{\blkpanelh}  \settoheight{\blkpanelh}{\blkimg{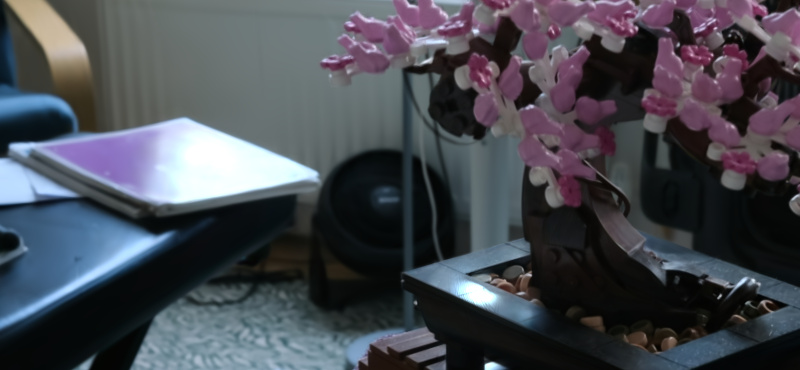}}
\newlength{\blkextrah}  \settoheight{\blkextrah}{\blkimg{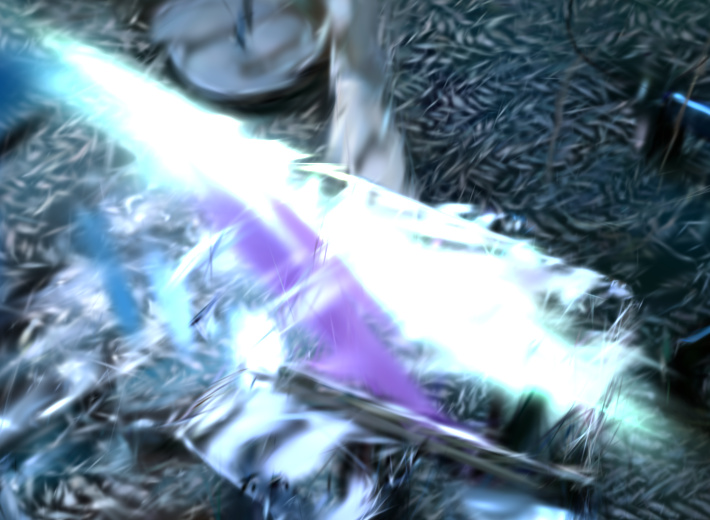}}
\newlength{\blkrefh}    \setlength{\blkrefh}{\dimexpr3\blkpanelh+\blkextrah+3\blkgap\relax}
\newcommand{\blkref}{\includegraphics[width=\blkrefw,height=\blkrefh]{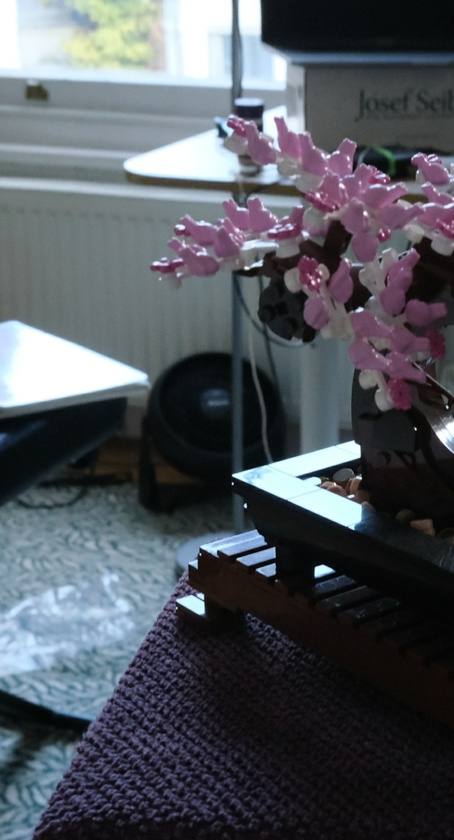}}
\newcommand{\blkheader}[2]{\node[anchor=south, inner sep=2pt, font=\scriptsize] at (#1.north) {#2};}
\newcommand{\blkrowlabel}[2]{%
\node[rotate=90, anchor=south, inner sep=1pt, font=\tiny, text height=3.5pt, text depth=1pt] at (#1.west) {#2};}

\begin{figure*}[t]
\centering
\noindent\makebox[\textwidth][c]{%
\begin{tikzpicture}
\node[anchor=north west, inner sep=0pt, minimum width=\blklabelw, minimum height=\blkrefh] (lab) at (0,0) {};
\node[ovpanel, anchor=north west, xshift=\blksep] (r0c0) at (lab.north east) {\blkimg{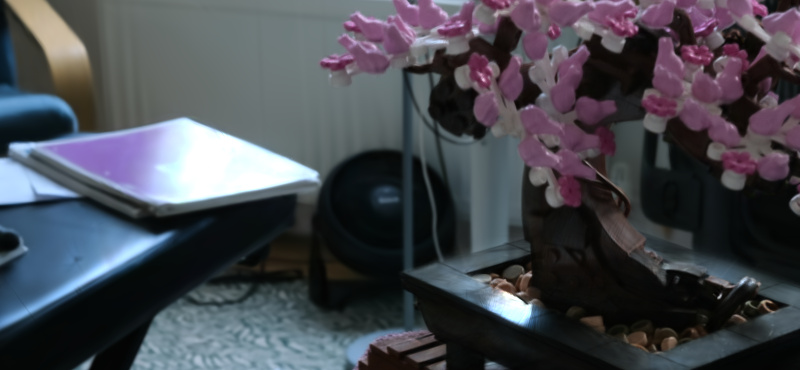}};
\node[ovpanel, anchor=north west, xshift=\blkgap] (r0c1) at (r0c0.north east) {\blkimg{sh_full_crop.jpg}};
\node[ovpanel, anchor=north west, xshift=\blkgap] (r0c2) at (r0c1.north east) {\blkimg{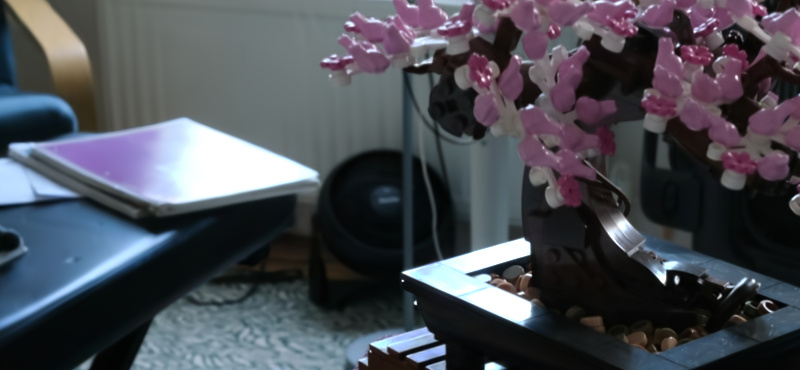}};
\node[ovpanel, anchor=north west, xshift=\blkgap] (r0c3) at (r0c2.north east) {\blkimg{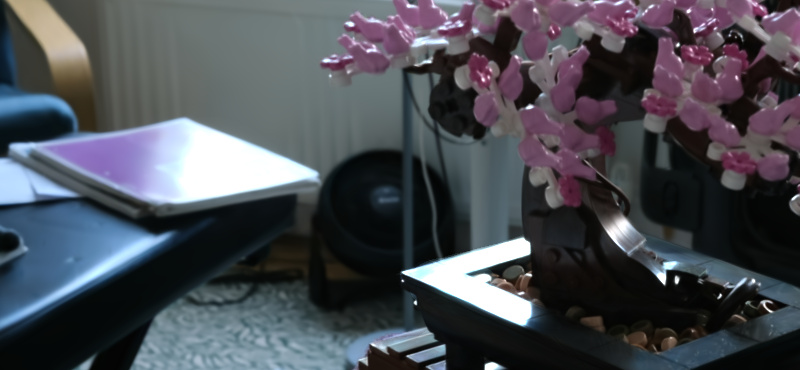}};
\node[ovpanel, anchor=north west, xshift=\blkgap] (r0c4) at (r0c3.north east) {\blkimg{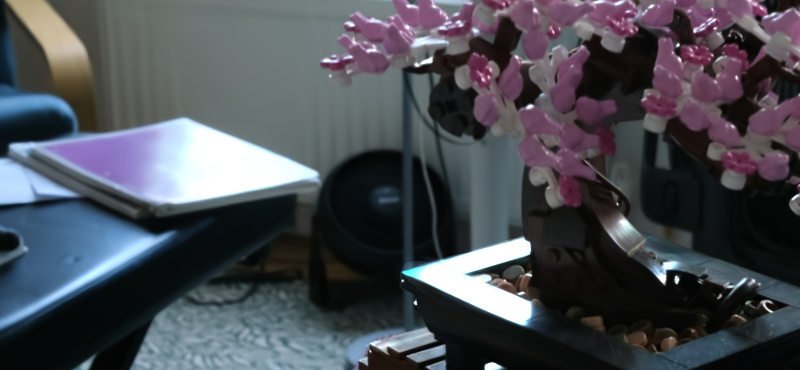}};
\node[ovpanel, anchor=north west, xshift=\blkgap] (r0c5) at (r0c4.north east) {\blkimg{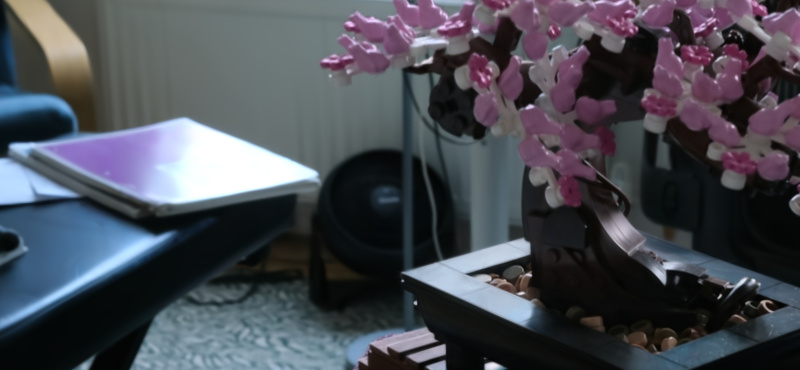}};
\node[ovpanel, anchor=north west, yshift=-\blkgap] (r1c0) at (r0c0.south west) {\blkimg{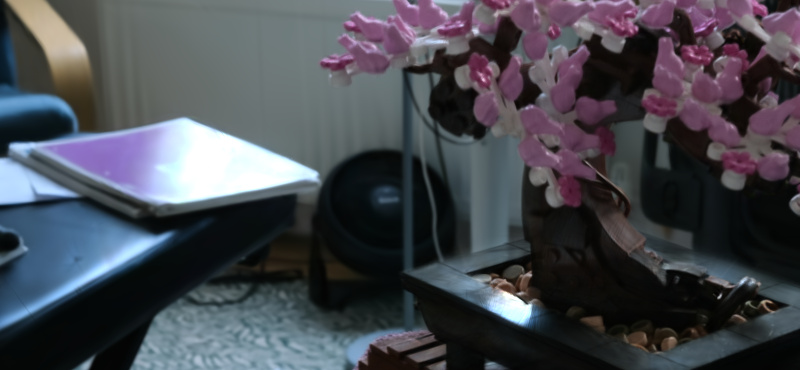}};
\node[ovpanel, anchor=north west, xshift=\blkgap] (r1c1) at (r1c0.north east) {\blkimg{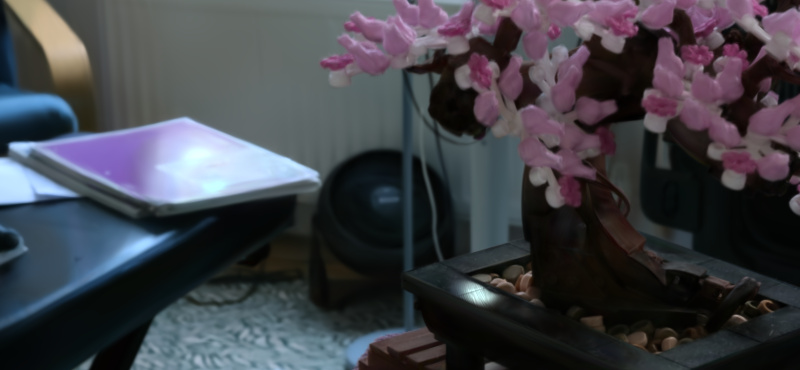}};
\node[ovpanel, anchor=north west, xshift=\blkgap] (r1c2) at (r1c1.north east) {\blkimg{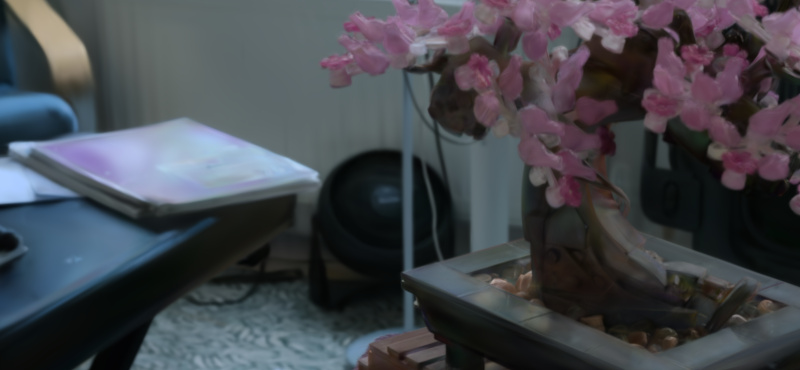}};
\node[ovpanel, anchor=north west, xshift=\blkgap] (r1c3) at (r1c2.north east) {\blkimg{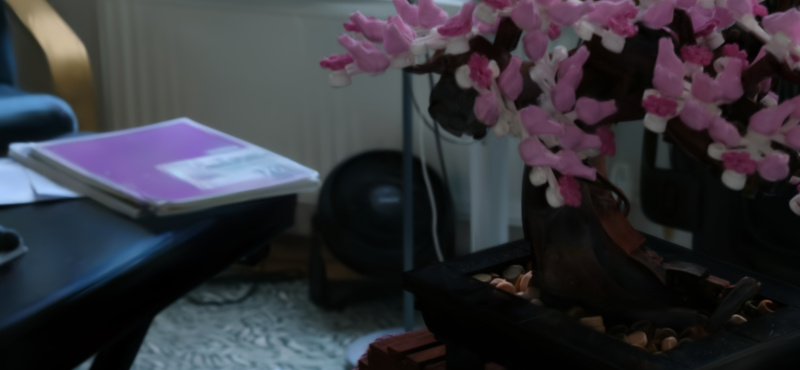}};
\node[ovpanel, anchor=north west, xshift=\blkgap] (r1c4) at (r1c3.north east) {\blkimg{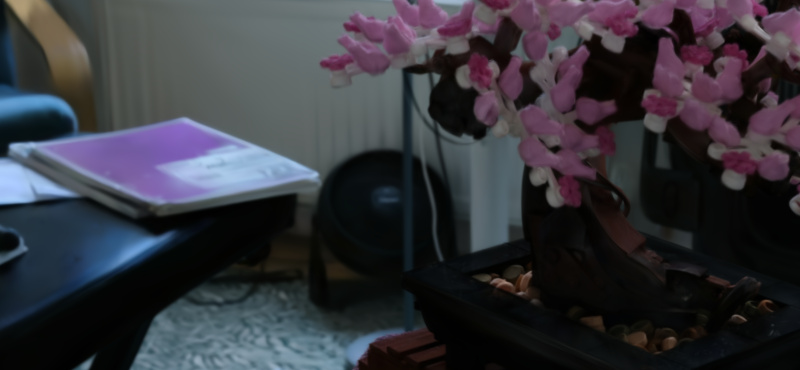}};
\node[ovpanel, anchor=north west, xshift=\blkgap] (r1c5) at (r1c4.north east) {\blkimg{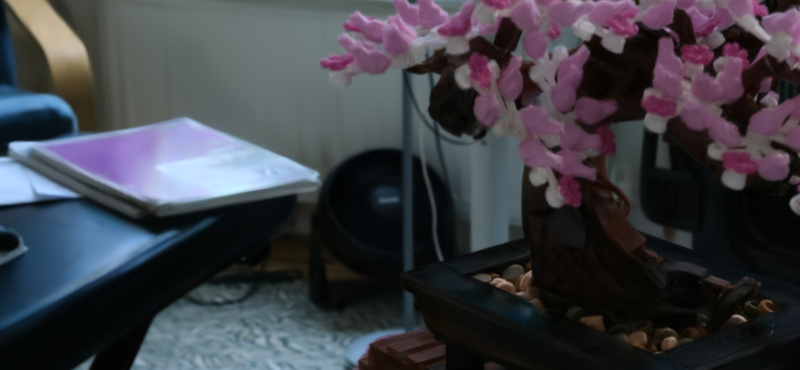}};
\node[ovpanel, anchor=north west, yshift=-\blkgap] (r2c0) at (r1c0.south west) {\blkimg{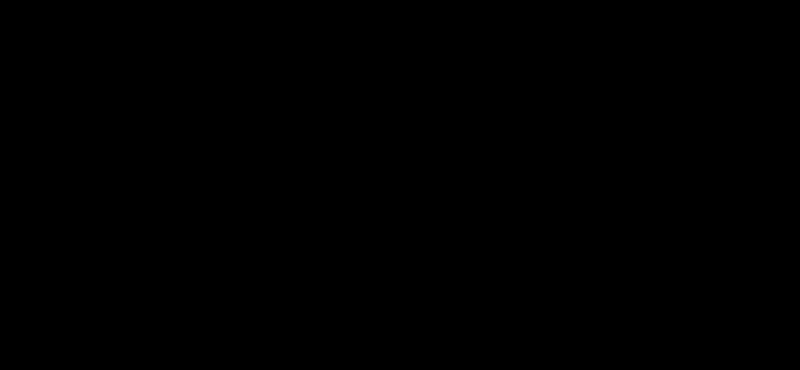}};
\node[ovpanel, anchor=north west, xshift=\blkgap] (r2c1) at (r2c0.north east) {\blkimg{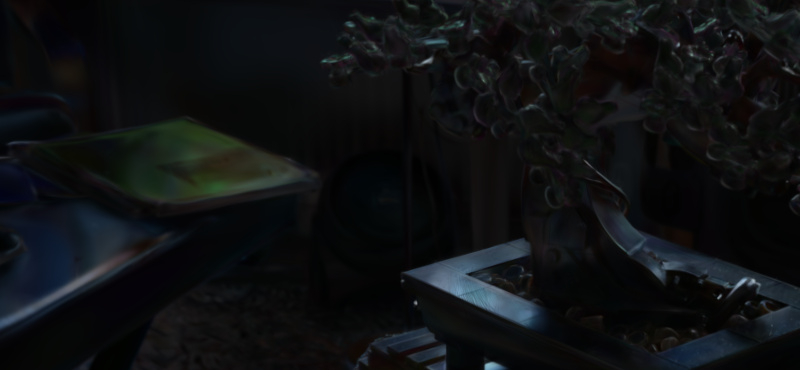}};
\node[ovpanel, anchor=north west, xshift=\blkgap] (r2c2) at (r2c1.north east) {\blkimg{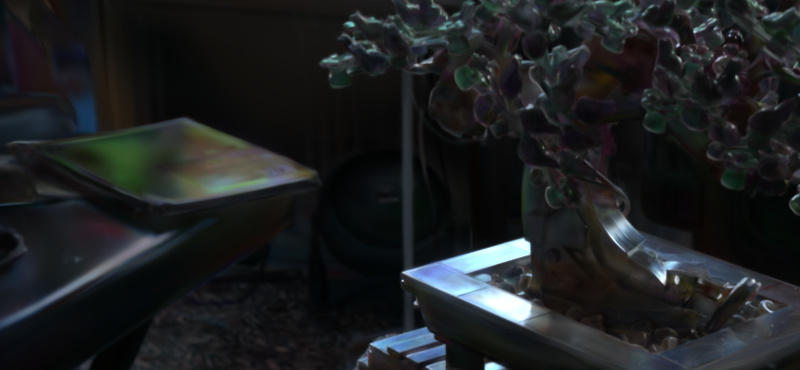}};
\node[ovpanel, anchor=north west, xshift=\blkgap] (r2c3) at (r2c2.north east) {\blkimg{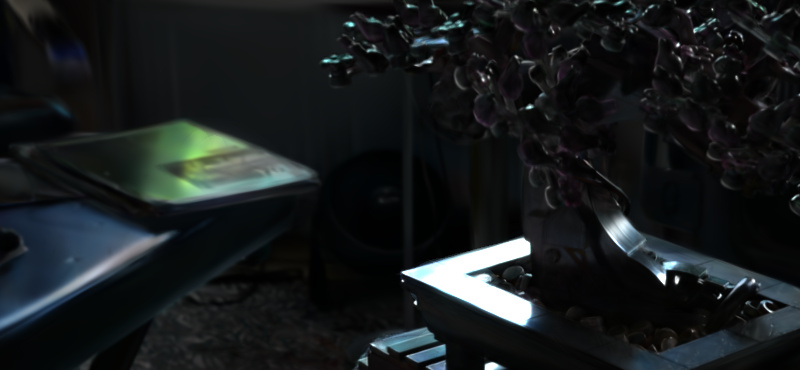}};
\node[ovpanel, anchor=north west, xshift=\blkgap] (r2c4) at (r2c3.north east) {\blkimg{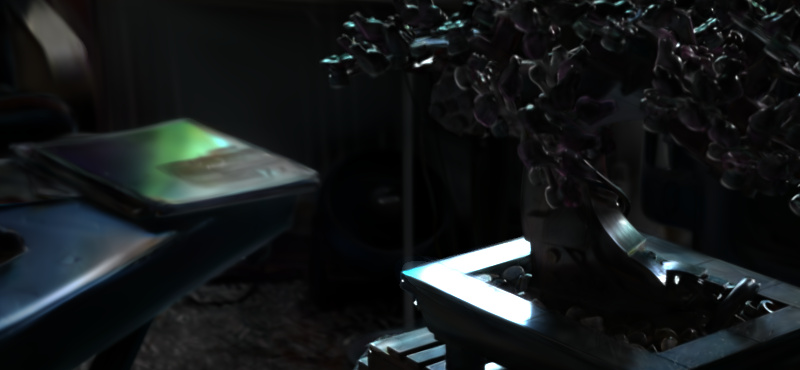}};
\node[ovpanel, anchor=north west, xshift=\blkgap] (r2c5) at (r2c4.north east) {\blkimg{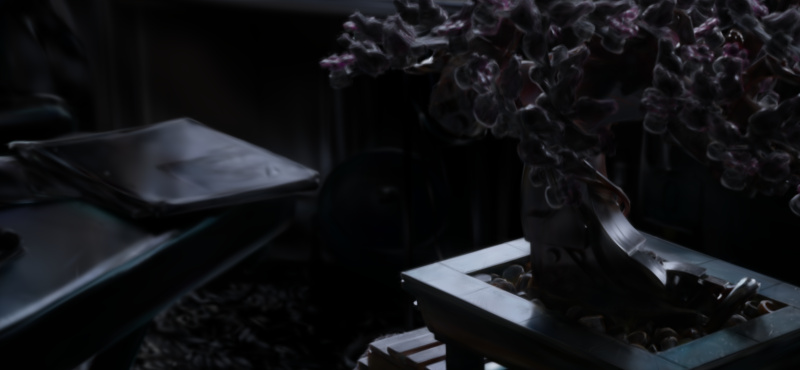}};
\node[ovpanel, anchor=north west, yshift=-\blkgap] (r3c0) at (r2c0.south west) {\blkimg{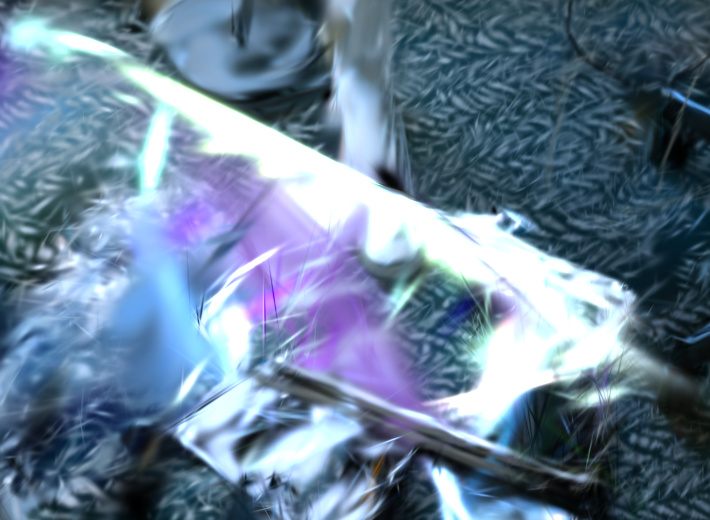}};
\node[ovpanel, anchor=north west, xshift=\blkgap] (r3c1) at (r3c0.north east) {\blkimg{sh_extrapolated_crop.jpg}};
\node[ovpanel, anchor=north west, xshift=\blkgap] (r3c2) at (r3c1.north east) {\blkimg{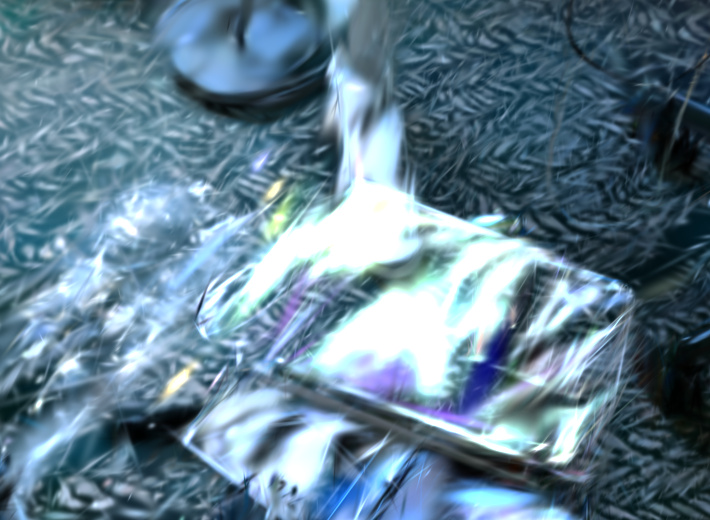}};
\node[ovpanel, anchor=north west, xshift=\blkgap] (r3c3) at (r3c2.north east) {\blkimg{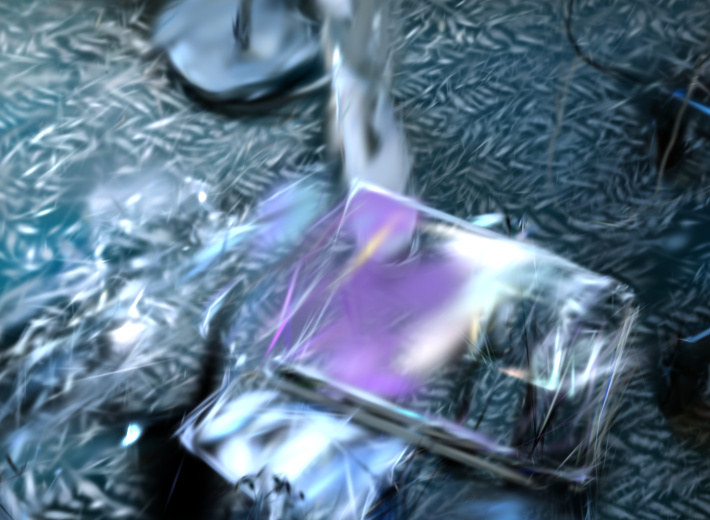}};
\node[ovpanel, anchor=north west, xshift=\blkgap] (r3c4) at (r3c3.north east) {\blkimg{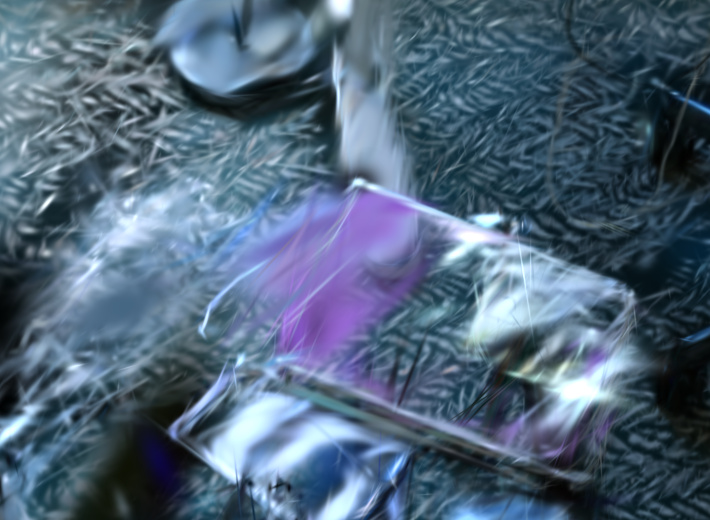}};
\node[ovpanel, anchor=north west, xshift=\blkgap] (r3c5) at (r3c4.north east) {\blkimg{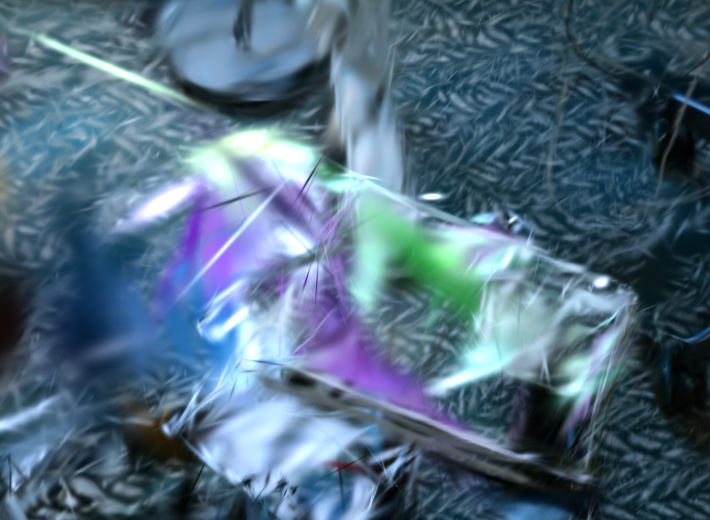}};
\node[ovpanel, anchor=north west, xshift=\blkgap] (ref) at (r0c5.north east) {\blkref};
\blkheader{ref}{Reference}
\blkheader{r0c0}{None}
\blkheader{r0c1}{SH}
\blkheader{r0c2}{SV}
\blkheader{r0c3}{NASG}
\blkheader{r0c4}{NASGabor}
\blkheader{r0c5}{Neural}
\blkrowlabel{r0c0}{Full}
\blkrowlabel{r1c0}{Base}
\blkrowlabel{r2c0}{Residual}
\blkrowlabel{r3c0}{Extrapolated View}
\end{tikzpicture}}%
\caption{%
Greater angular expressivity can reduce geometric artifacts. Models with high angular expressivity separate view-dependent reflections from diffuse color, whereas less expressive models compensate through geometry. SH uses elongated white Gaussians to fake the reflection on the sketch block and misses the reflection on the bonsai pot, while NASGabor achieves the best diffuse$\slash$specular decomposition.
}\label{fig:bonsai_block}
\end{figure*}
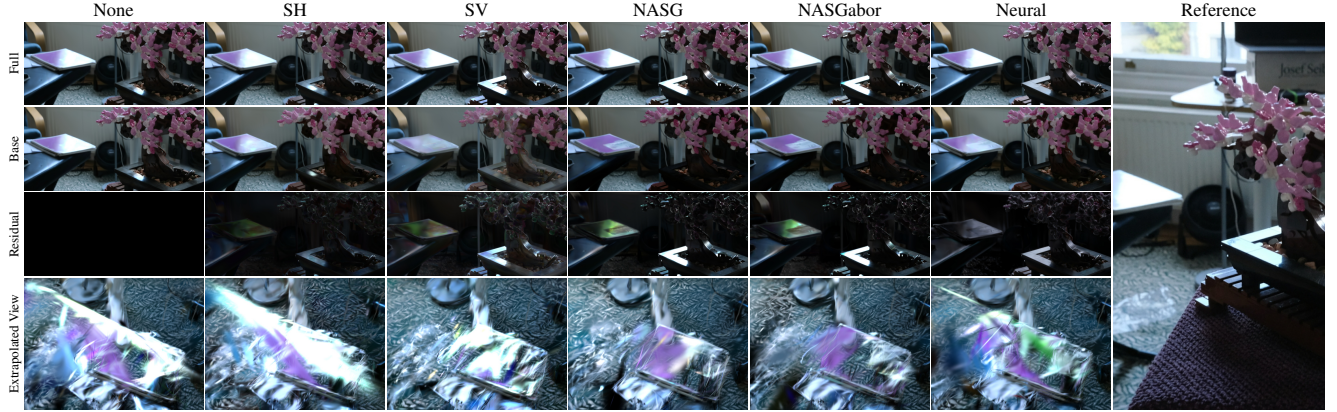

\newcommand{\vgood}{$\bm{++}$}
\newcommand{\good}{$+$}
\newcommand{\neutral}{$\circ$}
\newcommand{\bad}{$-$}
\newcommand{\vbad}{$\bm{--}$}
\begin{table*}[t]
\caption{Study of the evaluated view-dependent appearance models. Ranking goes from best (\vgood{}) to worst (\vbad{}).
}\label{tab:qualitative_analysis}
\centering
\setlength\tabcolsep{1.9pt}
\scriptsize
\begin{tabular}{lccccccl}
\toprule
Method & t\textsubscript{Train} & t\textsubscript{Render} & Angular Res. & Robustness & Simplicity & Compactness & Key Takeaway \\
\midrule
SH & \bad & \bad & \vbad & \vgood & \good & \bad & Overfits the least (strong SSIM$\slash$LPIPS); geometry must compensate for sharp reflections.\textsuperscript{1} \\
SV & \bad & \bad & \good & \bad & \bad & \bad & Parameter-efficient for sharp diffuse reflections; least robust method in our evaluation.\textsuperscript{2} \\
NASG(abor) & \good & \good & \vgood & \good & \bad & \good & Best quality$\slash$speed trade-off; its sharp lobes generalize well and can explain capture artifacts.\textsuperscript{3} \\
Neural & \good & \neutral & \good & \neutral & \vgood & \vgood & Most compact in general settings; hard to regularize per scene.\textsuperscript{4} \\
\midrule
\multicolumn{8}{l}{\parbox{.99\textwidth}{\textsuperscript{1}Costly Adam updates must also touch Gaussians invisible in the current iteration; rendering incurs many memory fetches.}} \\
\multicolumn{8}{l}{\parbox{.99\textwidth}{\textsuperscript{2}Ineffective with few sites (softmax requires $\geq 2$), making it slow and a poor fit for compact representations; explicit many-lobe models are hard to optimize via gradient descent.}} \\
\multicolumn{8}{l}{\parbox{.99\textwidth}{\textsuperscript{3}Weaker with a single lobe when primitives are few and lighting is controlled (\eg, NeRF Synthetic); Gabor term is not worth it outside standard benchmarks.}} \\
\multicolumn{8}{l}{\parbox{.99\textwidth}{\textsuperscript{4}Fits training images far better than explicit spherical functions; capacity reduction through regularization helps low view-dependence scenes but worsens view-dependent ones.}} \\ %
\bottomrule
\end{tabular}
\end{table*}

\subsection{Results}
In general, our results showcase the bottleneck imposed by heavy appearance models, substantially diminishing training and inference performance.
While on the other hand, compact yet flexible parametric models, and our proposed neural model, outperform them by virtue of their smaller memory footprint, while still managing comparable quality to the most expressive explicit models.
We include a detailed qualitative analysis across the different models in \cref{tab:qualitative_analysis}. 
All representations support rendering the view-independent base color and the view-dependent residual separately (\cf \cref{fig:bonsai_block}), a property enabled by the common interface in \cref{eq:common_appearance}.
Unlike prior work, we visualize the residual as $|\mathbf{c}(\mathbf{d}) - \phi\left(\mathbf{c}_0\right)|$ rather than $\phi\left(\mathcal{A}_{\psi}(\mathbf{d})\right)$, since the residual is signed and, \eg, $\phi=\operatorname{ReLU}$ would hide negative contributions.

\paragraph{Reconstruction Quality}
\cref{tab:training} reports reconstruction quality, optimization time, peak memory, and CUDA rendering speed on an RTX 4090.
All appearance models achieve similar quality across the evaluated datasets, with differences largely within run-to-run variation.
The neural representation matches or exceeds degree-3 SH while storing 28 instead of 192 bytes per Gaussian, resulting in faster optimization, lower peak memory, and higher rendering speed due to reduced memory traffic during parameter updates.

\paragraph{WebGL Performance}
\cref{tab:inference} reports frame times at 720p, averaged over multiple renders per test view.
In CUDA, NASG$\slash$NASGabor are the fastest models, rendering up to 18\% faster than SH on the RTX 4090 due to reduced memory traffic, with the neural model ranking third.
For the WebGL measurements, we load the same checkpoint for all models so that geometry is identical and differences stem exclusively from appearance evaluation.
Without Tensor Cores, differences narrow and all representations except SV render within a few percent of each other.
SV is the slowest model on all devices and runs out of memory on the iPhone for the two largest scenes.

\newlength{\actgap}
\setlength{\actgap}{1pt}
\newlength{\actpanelw}
\setlength{\actpanelw}{\dimexpr(\textwidth-3\actgap)/4\relax}
\newcommand{\actimg}[1]{\includegraphics[width=\actpanelw]{resources/activations/#1}}
\newlength{\actpanelh}
\settoheight{\actpanelh}{\actimg{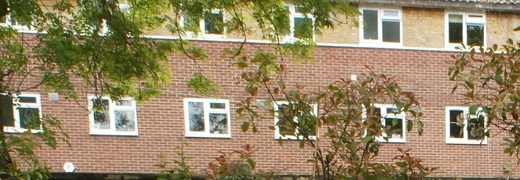}}
\newlength{\acttallh}
\setlength{\acttallh}{\dimexpr(3\actpanelh+\actgap)/2\relax}
\newcommand{\acttallimg}[1]{\includegraphics[width=\actpanelw,height=\acttallh]{resources/activations/#1}}
\newcommand{\actheader}[2]{\node[anchor=south, inner sep=2pt, font=\scriptsize] at (#1.north) {#2};}
\newcommand{\actlabelfont}{\fontsize{6}{7}\selectfont\mdseries}
\tikzset{actlabel/.style={ovlabel, anchor=north west, font=\actlabelfont,
                          text height=5pt, text depth=2pt}}
\newcommand{\actlabel}[2]{\node[actlabel] at (#1.north west) {#2};}

\begin{figure*}[t]
\centering
\noindent\makebox[\textwidth][c]{%
\begin{tikzpicture}
\node[ovpanel, anchor=north west] (rA) at (0,0) {\actimg{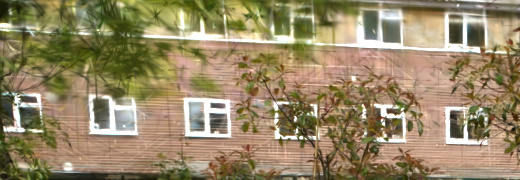}};
\node[ovpanel, anchor=north west, xshift=\actgap] (rAb) at (rA.north east) {\actimg{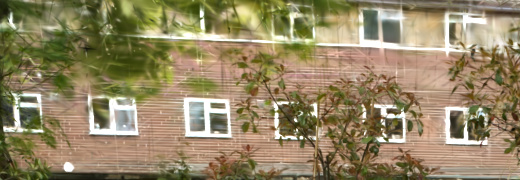}};
\node[ovpanel, anchor=north west, xshift=\actgap] (rAc) at (rAb.north east) {\actimg{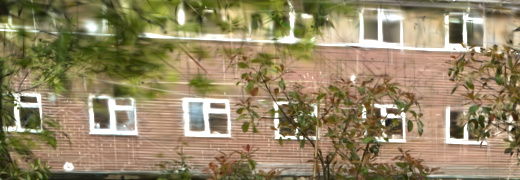}};
\node[ovpanel, anchor=north west, yshift=-\actgap] (rB) at (rA.south west) {\actimg{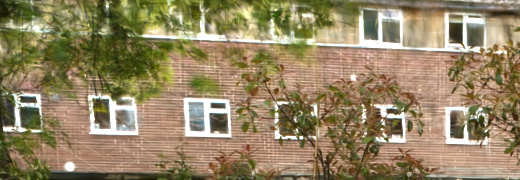}};
\node[ovpanel, anchor=north west, xshift=\actgap] (rBb) at (rB.north east) {\actimg{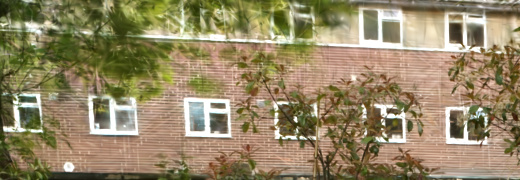}};
\node[ovpanel, anchor=north west, xshift=\actgap] (rBc) at (rBb.north east) {\actimg{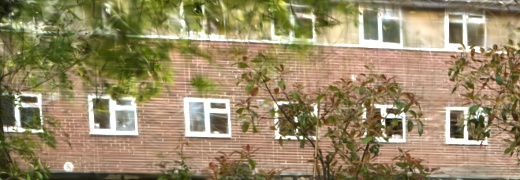}};
\node[ovpanel, anchor=north west, yshift=-\actgap] (rC) at (rB.south west) {\actimg{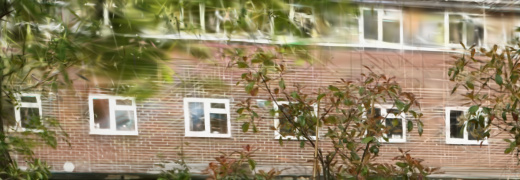}};
\node[ovpanel, anchor=north west, xshift=\actgap] (rCb) at (rC.north east) {\actimg{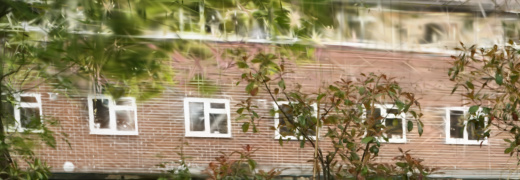}};
\node[ovpanel, anchor=north west, xshift=\actgap] (rCc) at (rCb.north east) {\actimg{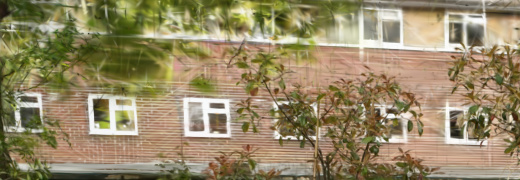}};
\node[ovpanel, anchor=north west, yshift=-\actgap] (rD) at (rC.south west) {\actimg{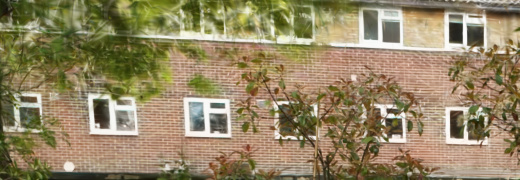}};
\node[ovpanel, anchor=north west, xshift=\actgap] (rDb) at (rD.north east) {\actimg{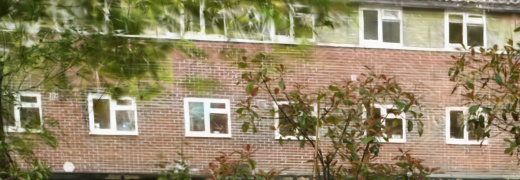}};
\node[ovpanel, anchor=north west, xshift=\actgap] (rDc) at (rDb.north east) {\actimg{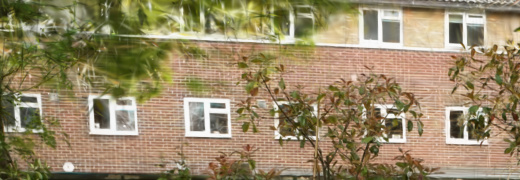}};
\node[ovpanel, anchor=north west, xshift=\actgap] (phA) at (rAc.north east) {\acttallimg{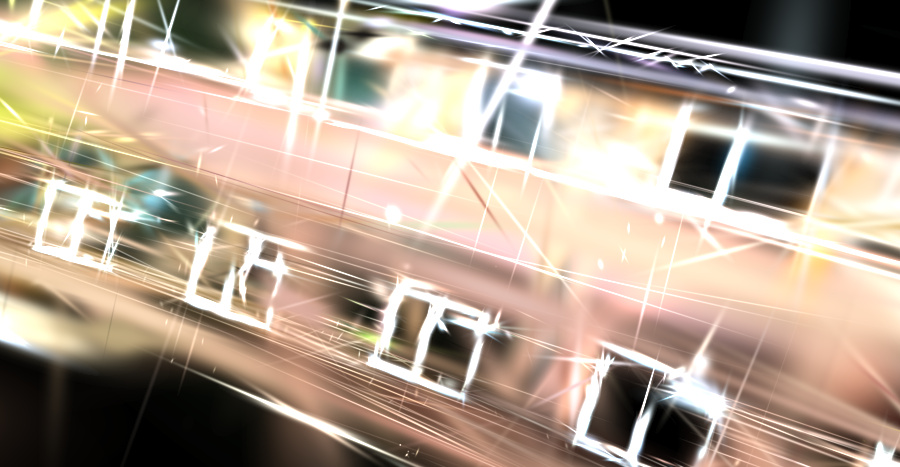}};
\node[ovpanel, anchor=north west, yshift=-\actgap] (phB) at (phA.south west) {\acttallimg{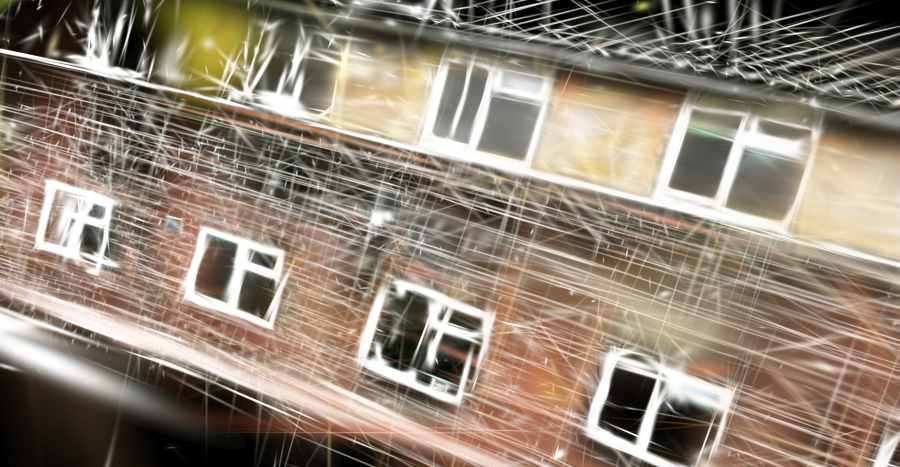}};
\node[ovpanel, anchor=north west, yshift=-\actgap] (gt)  at (phB.south west) {\actimg{garden_gt.jpg}};
\actheader{rA}{SH}
\actheader{rAb}{NASGabor}
\actheader{rAc}{Neural}
\actlabel{phA}{SH, ReLU, lr=0.05}
\actlabel{phB}{Neural, Sigmoid, lr=0.025}
\actlabel{gt}{Reference}
\actlabel{rA}{ReLU, lr=0.05}      \actlabel{rAb}{ReLU, lr=0.05}      \actlabel{rAc}{ReLU, lr=0.05}
\actlabel{rB}{Softplus, lr=0.025}      \actlabel{rBb}{Softplus, lr=0.025}      \actlabel{rBc}{Softplus, lr=0.025}
\actlabel{rC}{Sigmoid, lr=0.05}      \actlabel{rCb}{Sigmoid, lr=0.05}      \actlabel{rCc}{Sigmoid, lr=0.05}
\actlabel{rD}{Sigmoid, lr=0.025}      \actlabel{rDb}{Sigmoid, lr=0.025}      \actlabel{rDc}{Sigmoid, lr=0.025}
\end{tikzpicture}}%
\caption{%
Effect of color activation and opacity learning rate on a test view of the \emph{garden} scene. The choice of parametrization has strong  influence on how well the optimization recovers meaningful geometry as evident from the significantly improved details of the brick wall.
}\label{fig:activations_qualitative}
\end{figure*}

\begin{table}[t]
\caption{%
Quality metrics under different color activations and opacity learning rates across Mip-NeRF~360 scenes~\cite{barron2022mipnerf360} using MCMC densification and PPISP. The variant with lr=auto adjusts the opacity learning rate between 0.05 and 0.025 based on the scene-specific training view distribution as proposed by Miazga~\etal~\cite{miazga2026nasgabor}. 
}\label{tab:activations}
\centering
\setlength\tabcolsep{3.3pt}
\scriptsize
\begin{tabular}{lcccccc}
\toprule
 & \multicolumn{2}{c}{SH} & \multicolumn{2}{c}{NASGabor} & \multicolumn{2}{c}{Neural} \\
\cmidrule(lr){2-3} \cmidrule(lr){4-5} \cmidrule(lr){6-7}
Method                                   & PSNR$^\uparrow$ & SSIM$^\uparrow$ & PSNR$^\uparrow$ & SSIM$^\uparrow$ & PSNR$^\uparrow$ & SSIM$^\uparrow$ \\
\midrule
ReLU, lr=0.05                            & 28.35           & \cellcolor{2nd}0.836           & 28.51           & \cellcolor{2nd}0.836           & 28.53           & \cellcolor{2nd}0.831           \\
ReLU, lr=0.025                           & 28.33           & 0.831           & \cellcolor{3rd}28.59           & 0.833           & \cellcolor{3rd}28.54           & 0.827           \\
Softplus, lr=0.05                        & \cellcolor{1st}28.42           & \cellcolor{1st}0.837           & 28.55           & \cellcolor{1st}0.837           & \cellcolor{3rd}28.54           & \cellcolor{2nd}0.831           \\
Softplus, lr=0.025                       & \cellcolor{2nd}28.41           & 0.833           & \cellcolor{2nd}28.60           & 0.834           & \cellcolor{2nd}28.56           & 0.827           \\
Hardsigmoid, lr=0.05                     & 28.31           & \cellcolor{3rd}0.834           & 28.49           & \cellcolor{3rd}0.835           & 28.50           & \cellcolor{2nd}0.831           \\
Hardsigmoid, lr=0.025                    & 28.30           & 0.829           & 28.52           & 0.832           & 28.48           & 0.826           \\
Sigmoid, lr=0.05                         & 28.38           & \cellcolor{1st}0.837           & 28.43           & \cellcolor{3rd}0.835           & 28.53           & \cellcolor{1st}0.832           \\
Sigmoid, lr=0.025                        & \cellcolor{3rd}28.40           & 0.833           & 28.57           & 0.833           & 28.52           & 0.827           \\
\midrule
ReLU, lr=auto~\cite{miazga2026nasgabor}  & 28.38           & \cellcolor{3rd}0.834           & \cellcolor{1st}28.66           & \cellcolor{3rd}0.835           & \cellcolor{1st}28.58           & \cellcolor{3rd}0.829           \\
\bottomrule
\end{tabular}
\end{table}

\paragraph{Color Activations}
\cref{tab:activations} evaluates the color activations introduced in \cref{sec:appearance_models} across SH, NASGabor, and the neural model on Mip-NeRF~360.
As shown in \cref{fig:activations_qualitative}, differences manifest not only in metrics but in how well the optimization recovers geometric detail, \eg the brick wall in the \emph{garden} scene.
Sigmoid achieves the strongest results in SSIM and qualitative sharpness, as its bounded range prevents the degenerate solution, where unbounded activations fit opaque white regions with low opacity and colors far exceeding the unit value.
However, its vanishing gradients may cause worse fits in bright regions such as a white sky, where the source pixels are clipped to one.
When investigating per-scene metrics, we find that the optimal configuration varies strongly across scenes: higher opacity learning rates generally favor SSIM, whereas a lower ones favor PSNR, indicating that the color activation and the opacity parametrization interact during optimization.
This underscores that primitive-based radiance field methods require per-scene hyperparameter tuning to maximize quality, consistent with the findings of Miazga~\etal~\cite{miazga2026nasgabor}, whose per-scene adaptive learning rate mitigates this and achieves the best PSNR for NASGabor and the neural model.

\section{Discussion}
\label{sec:discussion}
\newlength{\bbgap}       \setlength{\bbgap}{1pt}      %
\newlength{\bbsep}       \setlength{\bbsep}{4pt}      %
\newlength{\bbvpad}      \setlength{\bbvpad}{0pt}     %
\newlength{\bbtextsep}   \setlength{\bbtextsep}{2pt}  %
\newlength{\bbrefdrop}   \setlength{\bbrefdrop}{20pt}
\newlength{\bbtmp}       \setlength{\bbtmp}{\dimexpr\textwidth-6.5pt\relax}
\newlength{\bbcellw}     \setlength{\bbcellw}{0.11281\bbtmp}
\newlength{\bbcellh}     \setlength{\bbcellh}{0.54054\bbcellw}
\newlength{\bbgridh}     \setlength{\bbgridh}{\dimexpr6\bbcellh+5\bbgap\relax}
\newlength{\bbrefh}      \setlength{\bbrefh}{\dimexpr\bbgridh-\bbrefdrop\relax}
\newcommand{\bbimg}[1]{\includegraphics[width=\bbcellw]{resources/bonsai_book/#1.jpg}}
\newcommand{\bbref}[1]{\includegraphics[height=\bbrefh]{resources/bonsai_book/#1.jpg}}
\newlength{\bbcompw}    \setlength{\bbcompw}{7pt}   %
\tikzset{bblabel/.style={rotate=90, anchor=south, inner sep=1pt, font=\tiny,
                         text height=3.5pt, text depth=1pt}}
\newcommand{\bbcomplabel}[2]{\node[bblabel] at (#1.west) {#2};}
\newcommand{\bbmethodlabel}[2]{\node[bblabel] at ([xshift=-\bbcompw]#1.west) {#2};}
\newcommand{\bbreftext}[2]{\node[anchor=north, font=\scriptsize, inner sep=0pt, yshift=-\bbtextsep] at (#1.south) {#2};}

\begin{figure*}[t]
\centering
\noindent\makebox[\textwidth][c]{%
\begin{tikzpicture}
\node[ovpanel, anchor=north west, yshift=-\bbvpad] (ref0) at (0,0) {\bbref{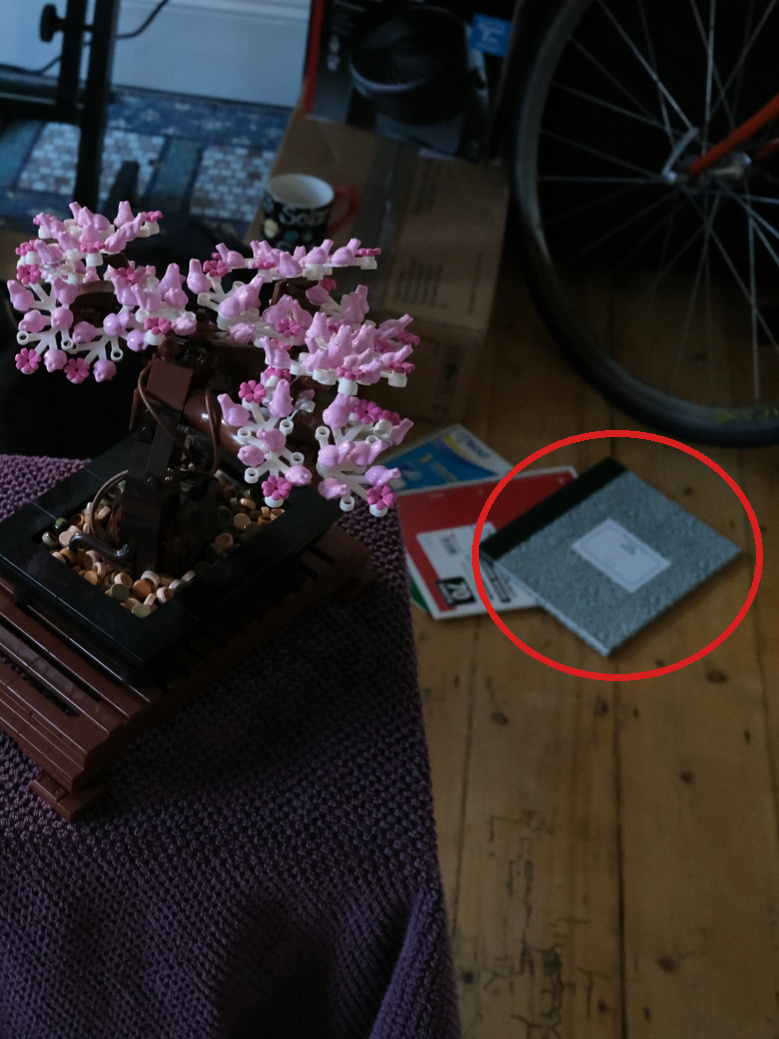}};
\bbreftext{ref0}{Reference View 0}
\node[anchor=north west, inner sep=0pt, xshift=\bbsep, minimum width=14pt, minimum height=\bbgridh]
      (lab) at (ref0.north east) {};
\node[ovpanel, anchor=north west, xshift=\bbsep] (sf) at (lab.north east) {\bbimg{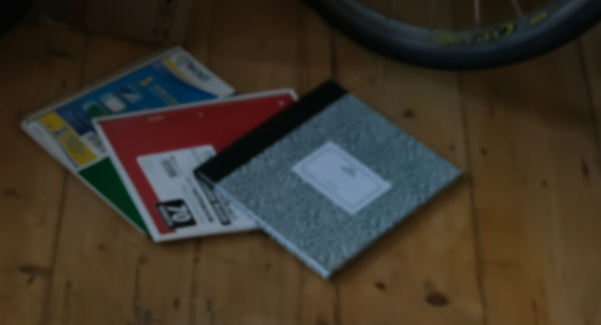}};
\node[ovpanel, anchor=north west, xshift=\bbgap] (sf1) at (sf.north east) {\bbimg{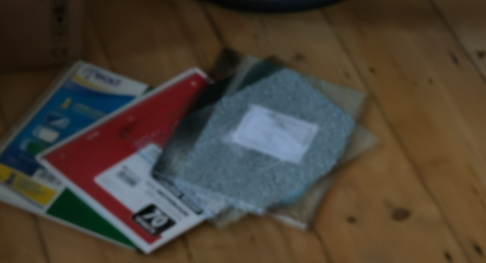}};
\node[ovpanel, anchor=north west, xshift=\bbgap] (sf2) at (sf1.north east) {\bbimg{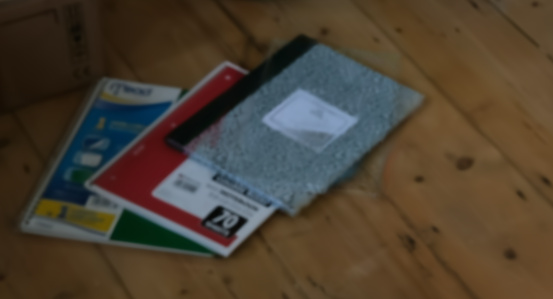}};
\node[ovpanel, anchor=north west, xshift=\bbgap] (sf3) at (sf2.north east) {\bbimg{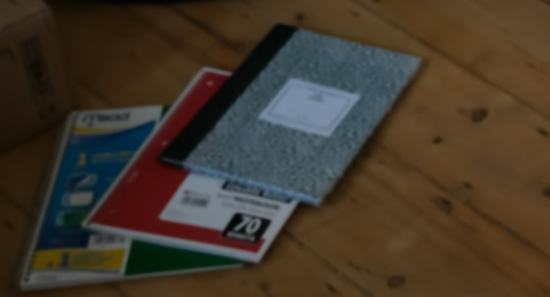}};
\node[ovpanel, anchor=north west, yshift=-\bbgap] (sb) at (sf.south west) {\bbimg{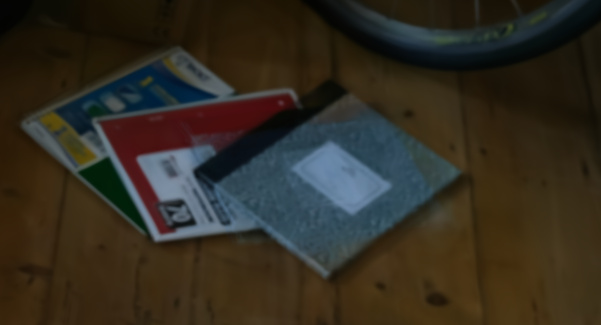}};
\node[ovpanel, anchor=north west, xshift=\bbgap] (sb1) at (sb.north east) {\bbimg{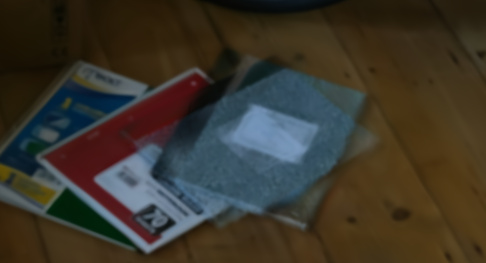}};
\node[ovpanel, anchor=north west, xshift=\bbgap] (sb2) at (sb1.north east) {\bbimg{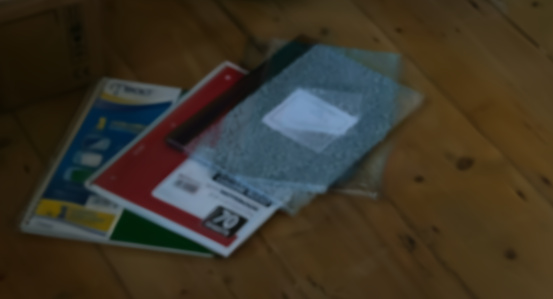}};
\node[ovpanel, anchor=north west, xshift=\bbgap] (sb3) at (sb2.north east) {\bbimg{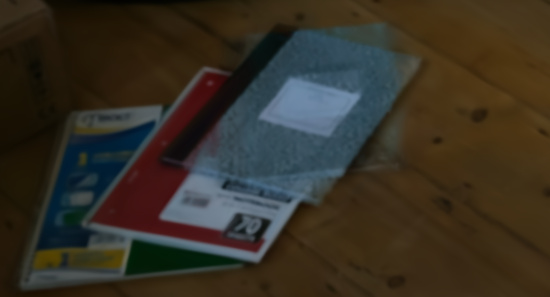}};
\node[ovpanel, anchor=north west, yshift=-\bbgap] (sr) at (sb.south west) {\bbimg{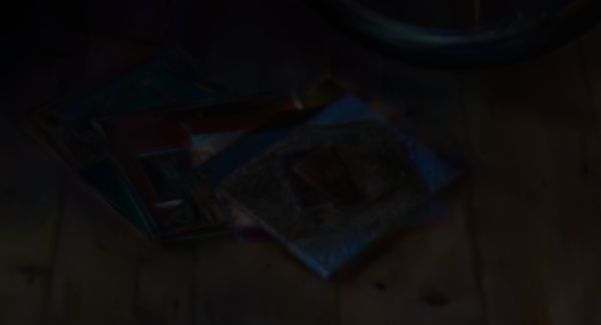}};
\node[ovpanel, anchor=north west, xshift=\bbgap] (sr1) at (sr.north east) {\bbimg{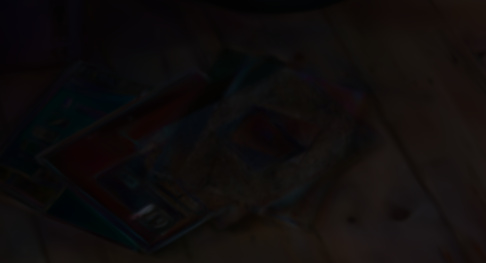}};
\node[ovpanel, anchor=north west, xshift=\bbgap] (sr2) at (sr1.north east) {\bbimg{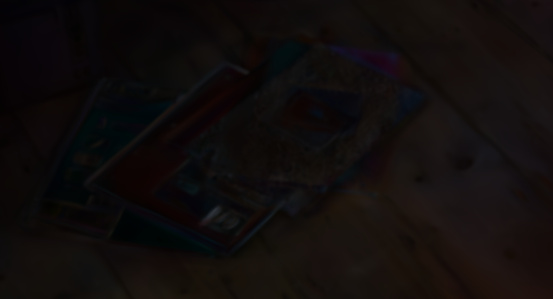}};
\node[ovpanel, anchor=north west, xshift=\bbgap] (sr3) at (sr2.north east) {\bbimg{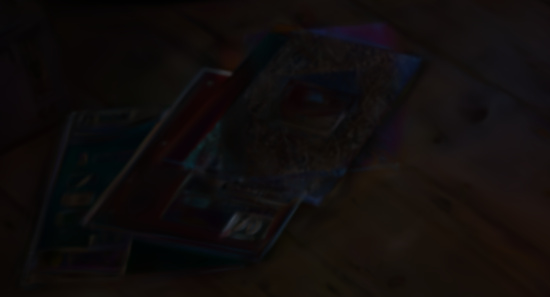}};
\node[ovpanel, anchor=north west, yshift=-\bbgap] (nf) at (sr.south west) {\bbimg{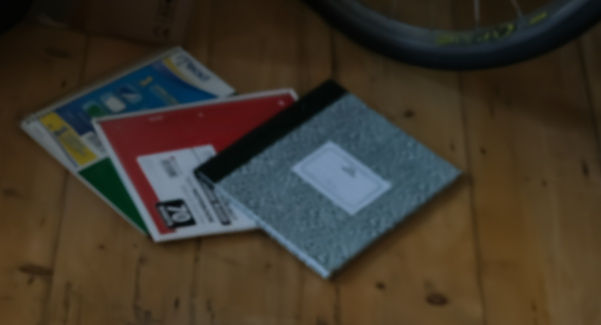}};
\node[ovpanel, anchor=north west, xshift=\bbgap] (nf1) at (nf.north east) {\bbimg{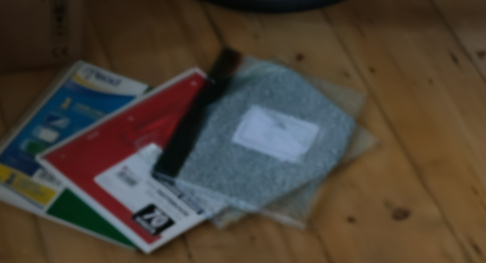}};
\node[ovpanel, anchor=north west, xshift=\bbgap] (nf2) at (nf1.north east) {\bbimg{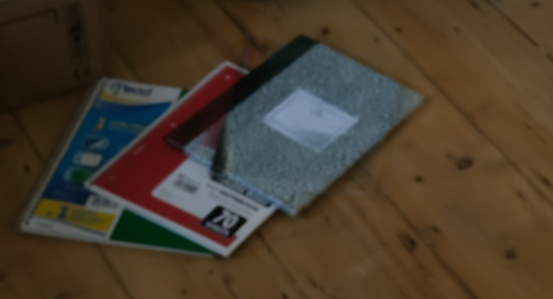}};
\node[ovpanel, anchor=north west, xshift=\bbgap] (nf3) at (nf2.north east) {\bbimg{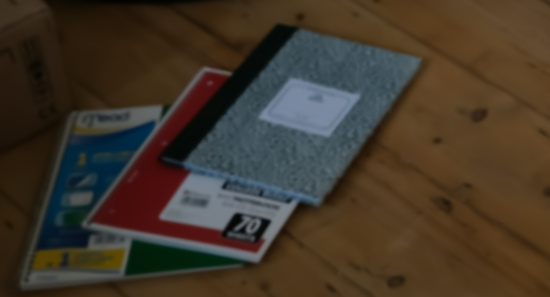}};
\node[ovpanel, anchor=north west, yshift=-\bbgap] (nb) at (nf.south west) {\bbimg{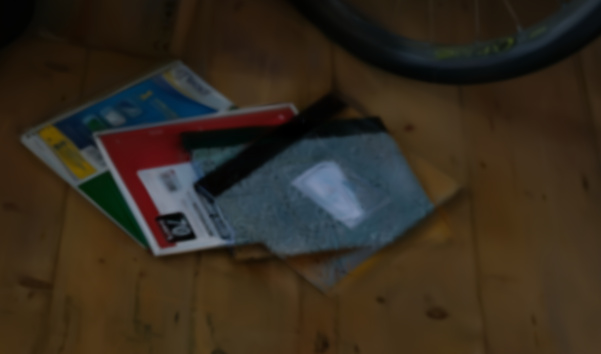}};
\node[ovpanel, anchor=north west, xshift=\bbgap] (nb1) at (nb.north east) {\bbimg{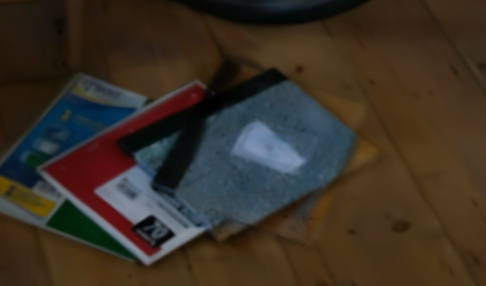}};
\node[ovpanel, anchor=north west, xshift=\bbgap] (nb2) at (nb1.north east) {\bbimg{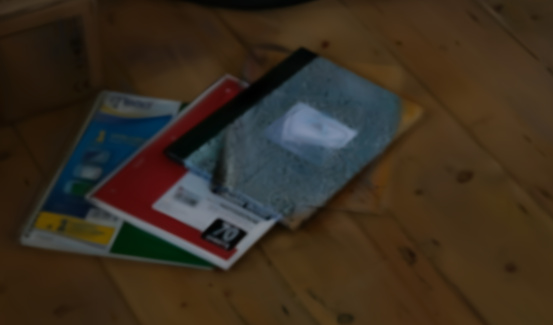}};
\node[ovpanel, anchor=north west, xshift=\bbgap] (nb3) at (nb2.north east) {\bbimg{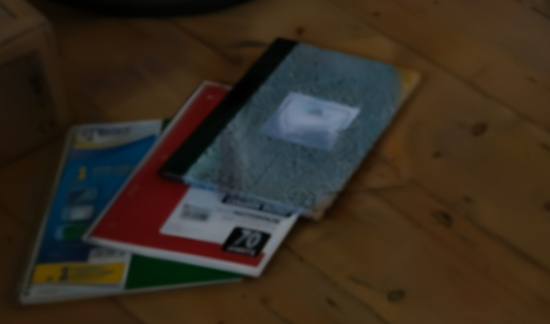}};
\node[ovpanel, anchor=north west, yshift=-\bbgap] (nr) at (nb.south west) {\bbimg{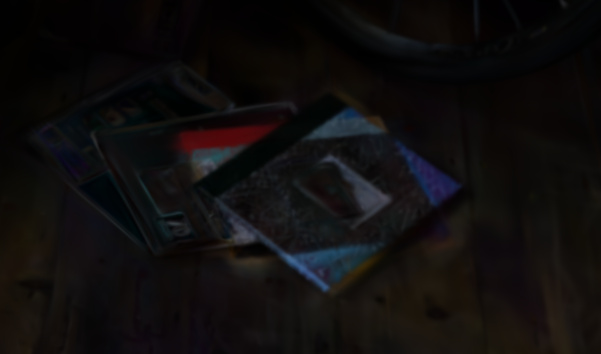}};
\node[ovpanel, anchor=north west, xshift=\bbgap] (nr1) at (nr.north east) {\bbimg{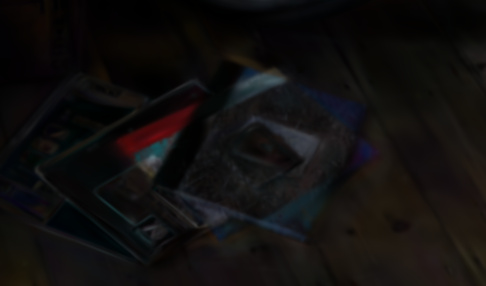}};
\node[ovpanel, anchor=north west, xshift=\bbgap] (nr2) at (nr1.north east) {\bbimg{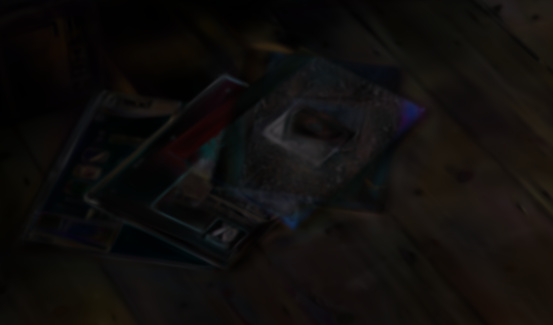}};
\node[ovpanel, anchor=north west, xshift=\bbgap] (nr3) at (nr2.north east) {\bbimg{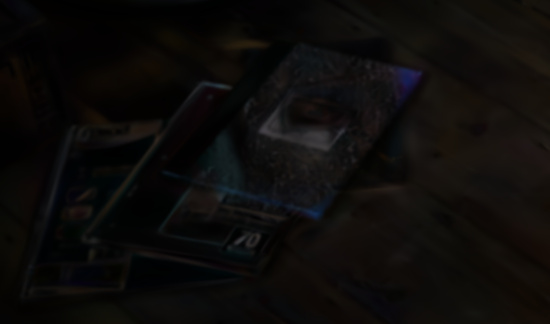}};
\node[ovpanel, anchor=north west, xshift=\bbsep, yshift=-\bbvpad] (ref1) at (sf3.north east) {\bbref{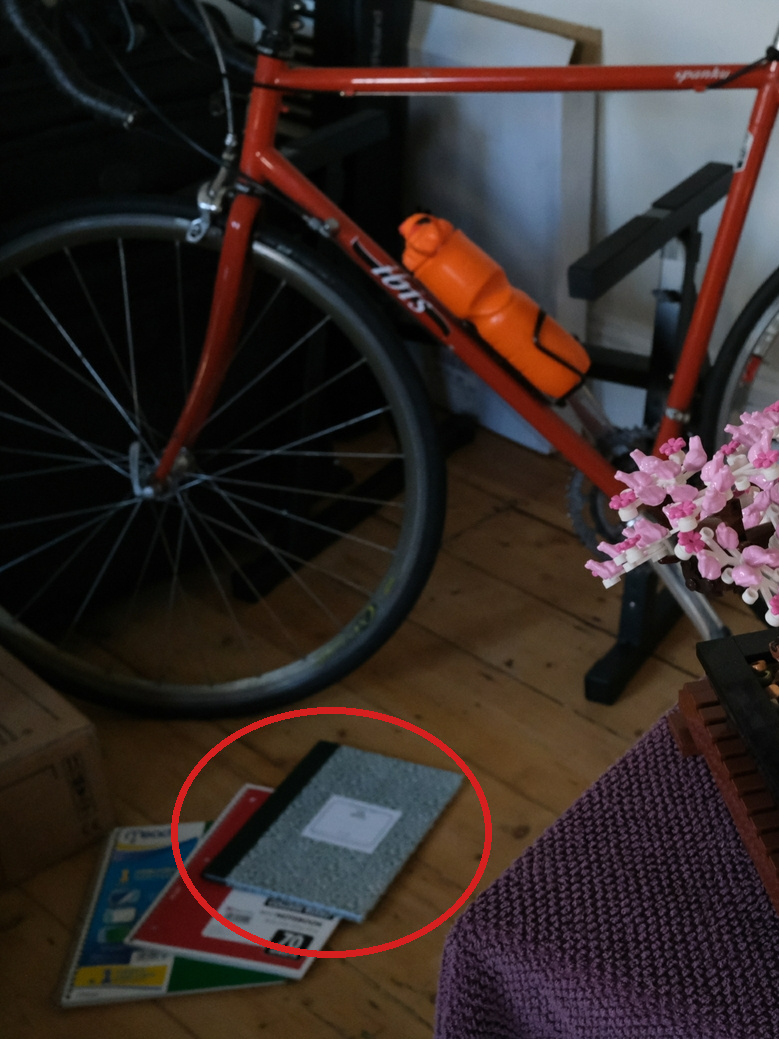}};
\bbreftext{ref1}{Reference View 1}
\bbcomplabel{sf}{Full}
\bbcomplabel{sb}{Base}
\bbcomplabel{sr}{Residual}
\bbcomplabel{nf}{Full}
\bbcomplabel{nb}{Base}
\bbcomplabel{nr}{Residual}
\bbmethodlabel{sb}{SH}
\bbmethodlabel{nb}{NASGabor}
\end{tikzpicture}}%
\caption{%
Base color and view-dependent residual along a camera path between two training views of the \emph{bonsai} scene, cropped to the books on the floor. The first and last crop show the reference views, the others show in-between views. The topmost book moved during the data capture. NASGabor's expressivity allows partially fitting this artifact as a view-dependent effect, while SH requires the geometry to compensate through popping~\cite{radl2024stopthepop}, an insight that motivates employing expressive appearance models with pixel-sorted approaches~\cite{hahlbohm2025htgs, steiner2025aaags}.
}\label{fig:bonsai_book}
\end{figure*}

\newlength{\trailgap}
\setlength{\trailgap}{1pt}
\newlength{\trailpanelw}
\setlength{\trailpanelw}{\dimexpr(\columnwidth-2\trailgap)/3\relax}
\newcommand{\trailimg}[1]{\includegraphics[width=\trailpanelw]{resources/garden_trail/#1}}
\newcommand{\traillabelfont}{\fontsize{6}{7}\selectfont\mdseries}
\tikzset{traillabel/.style={ovlabel, anchor=north west, font=\traillabelfont,
                            text height=5pt, text depth=2pt}}

\begin{figure}[t]
\centering
\noindent\makebox[\columnwidth][c]{%
\begin{tikzpicture}
\node[ovpanel, anchor=north west] (p1) at (0,0) {\trailimg{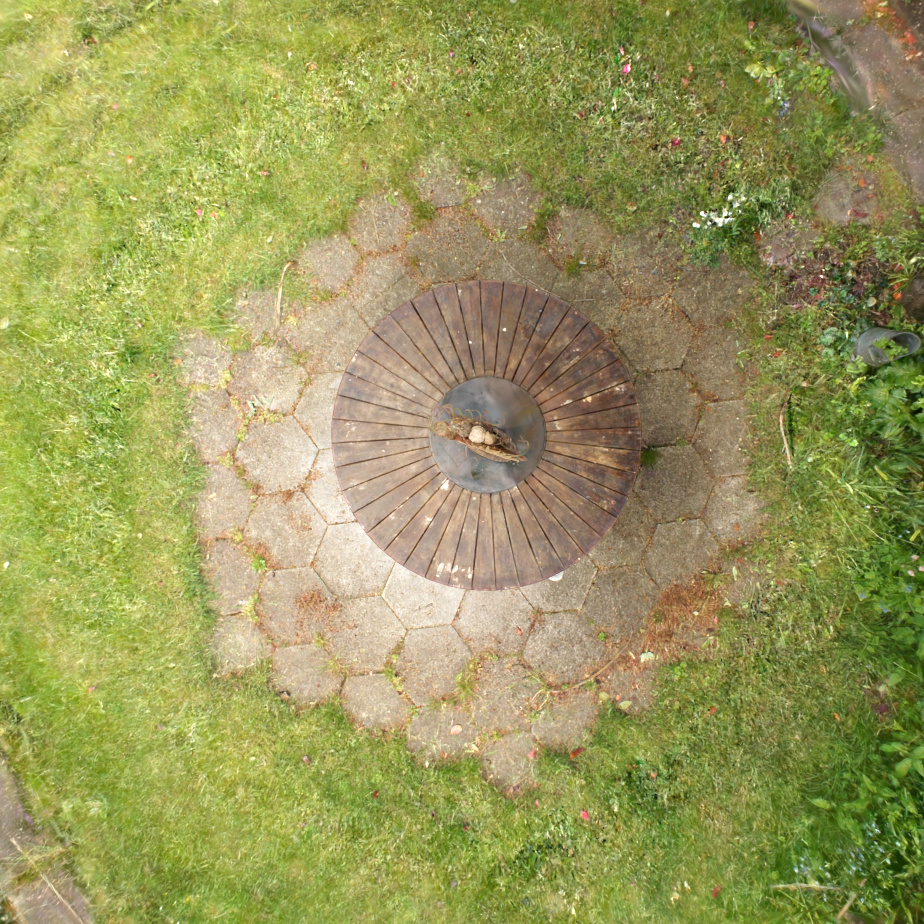}};
\node[ovpanel, anchor=north west, xshift=\trailgap] (p2) at (p1.north east) {\trailimg{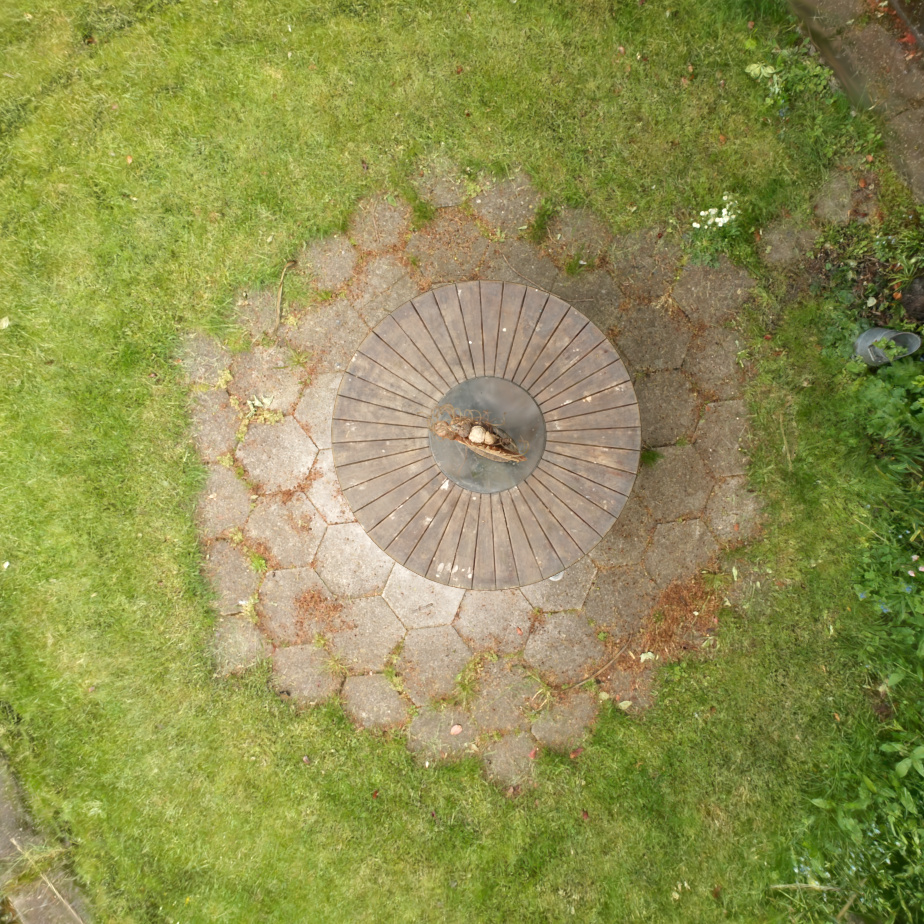}};
\node[ovpanel, anchor=north west, xshift=\trailgap] (p3) at (p2.north east) {\trailimg{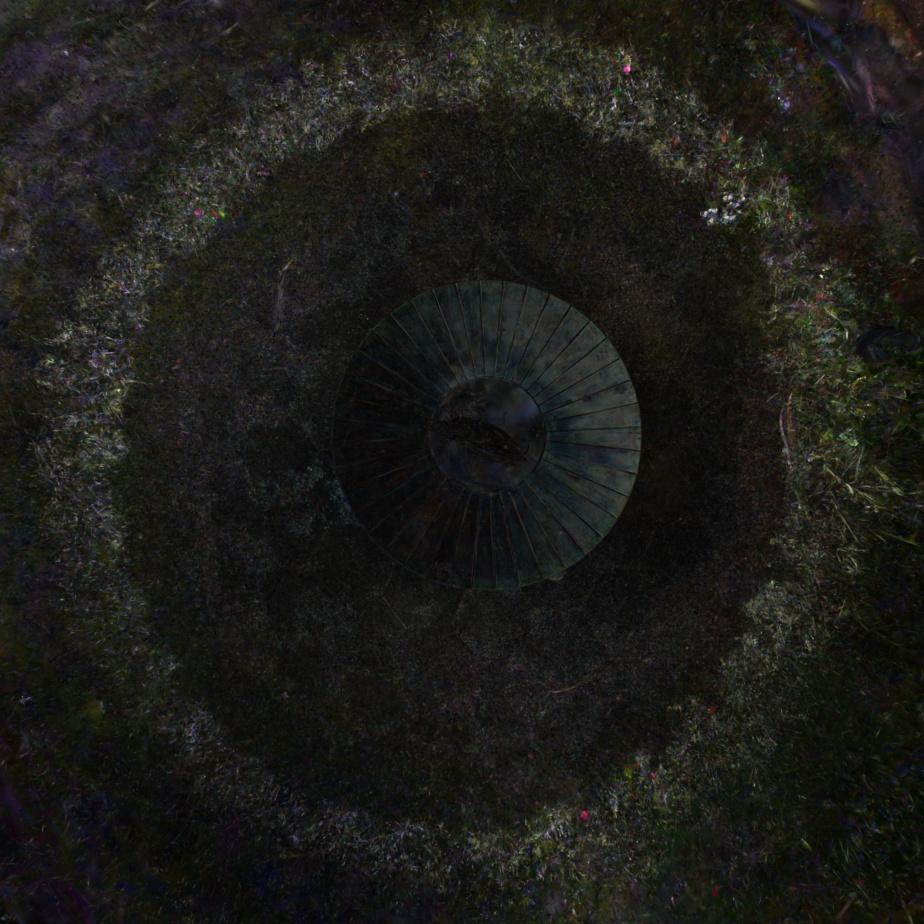}};
\node[traillabel] at (p1.north west) {Full};
\node[traillabel] at (p2.north west) {Base};
\node[traillabel] at (p3.north west) {Residual};
\end{tikzpicture}}%
\caption{%
Base color and view-dependent residual with SH appearance rendered separately for a top-down view of the \emph{garden} scene. The residual reveals where the camera operator inadvertently modified the scene by stepping on the grass during the capture, \ie, the optimizer fit changes in geometry through the appearance model. 
}\label{fig:garden_trail}
\end{figure}
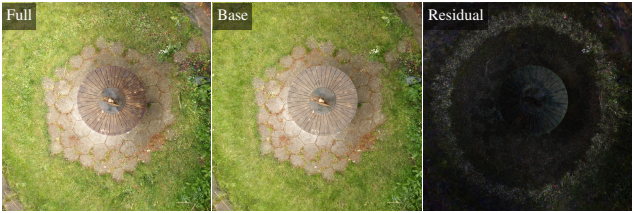

While our evaluation provides a broad comparison among appearance models for novel-view synthesis, several factors complicate a definitive assessment of their relative strengths. 
In particular, we argue that \emph{view-dependent appearance} should not necessarily be equated with purely modelling outgoing radiance.
On real captures, view-dependent residuals can instead encode inconsistencies in the observations. These can be caused by changing illumination, shadows, exposure, scene motion, or imperfect geometry. 
An expressive appearance model can absorb these inconsistencies since, along a finite camera trajectory, temporal or spatial changes may be correlated with viewing direction.

This effect is visible, \eg, in Mip-NeRF~360 scenes (\cf \cref{fig:bonsai_book,fig:garden_trail}).
In \emph{bonsai}, the top-most book moved during capture. NASGabor's expressive angular model can partially represent this discrepancy as a view-dependent effect, whereas the more constrained SH representation has less capacity to do so and consequently needs to abuse deficiencies of the rendering algorithm such as popping~\cite{radl2024stopthepop}.
Similarly, in \emph{garden}, the view-dependent SH residual reveals a region where the camera operator stepped on the grass during capture.
Here, the residual is not a reflectance property, but rather the representation of a scene change that happened to correlate with the camera trajectory.

This raises a broader question: to what extent does the advantage of expressive models on real-world benchmarks measure capacity to model outgoing radiance, or does it simply measure robustness to violations of the static-scene assumption?
In other words, a model may obtain better novel-view synthesis by being better at explaining inconsistent observations, without necessarily recovering a more faithful representation of the underlying scene.
This ambiguity is largely absent from synthetic benchmarks with fixed geometry, lighting, and camera parameters, but these introduce a different bias.
Their predominantly diffuse, hard-surface content can penalize explicit representations whose primitives have limited angular expressivity, unless additional parametric lobes are introduced.

Consequently, current benchmarks make it difficult to isolate the intrinsic quality of an appearance representation.
We therefore believe that future evaluations should distinguish \emph{reflectance and lighting-dependent angular variation} from \emph{observation inconsistency}, rather than treating both simply as view-dependent appearance.
Despite this ambiguity in image quality, our memory and performance measurements provide a more direct metric: compact spherical-parametric and implicit representations offer a substantial practical advantage over storing increasingly expressive but heavy appearance models explicitly per primitive.

\paragraph{Limitations}
All appearance models in this comparison share the limitations inherent to spherical color functions in 3DGS: they cannot accurately model mirror reflections, which require explicit tracing of secondary rays~\cite{verbin2024nerfcasting, xie2025envgs}.
For the neural representation, the shared decoder introduces an additional constraint: unlike an explicit spherical basis, its response cannot be analytically rotated, complicating scene composition and editing~\cite{schuetz2025splatshop}.
This can be addressed by applying the inverse model transformation to the view direction before querying the network.
The shared decoder may also become a capacity bottleneck as scene scale increases. 
One possible solution is to store multiple sets of decoder weights, as in SMERF~\cite{duckworth2023smerf}, which partitions the scene into a grid and trilinearly interpolates per-cell weights at the camera origin. Besides increasing capacity, such a formulation naturally supports scene composition.
Finally, WebGL does not expose the hardware acceleration that makes neural decoding inexpensive in CUDA, making parametric models faster in this setting. 
As mobile hardware increasingly incorporates neural accelerators and APIs such as WebGPU provide access to them, we expect this gap to narrow.

\paragraph{Future Work}
The compact neural representation is particularly attractive for feedforward reconstruction, since predicting an 8-dimensional latent code is substantially simpler than predicting 48 SH coefficients, while the decoder itself can be shared across scenes.
More broadly, our unified framework makes it possible to study the interaction between appearance representation, color activation, and densification strategy in a controlled manner. 
We believe this joint design space, rather than treating these components independently, is a promising direction for future work.

\section{Conclusion}
\label{sec:conclusion}
We presented a controlled comparison of view-dependent appearance models for Gaussian splatting, integrating spherical harmonics, spherical Voronoi, normalized anisotropic spherical Gaussians, NASGabor, and a compact neural decoder into a common pipeline with optimized CUDA and WebGL implementations.
Our evaluation shows that recent spherical models provide strong quality--efficiency trade-offs, while our neural representation is the most compact, reducing the per-primitive appearance footprint from 192 to 28 bytes compared to third-degree SH while matching or improving reconstruction quality.
Beyond image quality, we find that appearance parametrization and color activation influence optimization and, consequently, recovered geometry, highlighting the need for controlled comparisons when evaluating this design space.
At the same time, non-static scenes and transient capture artifacts in common benchmarks complicate the interpretation of view-dependent appearance, as models may explain such inconsistencies through appearance or geometry.
Because all evaluated models operate on per-primitive view directions, our findings transfer directly to other explicit primitive-based radiance-field representations, extending their relevance beyond Gaussian splatting.
We release our configurable framework to facilitate further research into appearance representations and their quality, memory, and platform-specific trade-offs.

\section*{Acknowledgments}
We thank Timon Scholz for his contributions to the software infrastructure supporting this work and Jannis Möller for identifying and helping us fix a bug in tiny-cuda-nn.
This work was partially funded by the DFG projects ``Real-Action VR'' (ID 523421583) and ``Increasing Realism of Omnidirectional Videos in Virtual Reality'' (ID 491805996).

{
    \small
    \bibliographystyle{ieeenat_fullname}
    \bibliography{main}
}
\appendix
\section{Appearance Model Details}
\label{sec:suppl_appearance_model_details}
This section complements \cref{sec:appearance_models,sec:neural_appearance} with the parametrization, initialization, and optimization of each appearance model, and lists where our implementation deviates from the respective reference.
We intentionally restrict all model-specific adaptations to the appearance parameters, \ie, the base color and the view-dependent parameters.
Changes that the reference implementations make to the geometry, such as different learning rates or schedules for means, scales, rotations, and opacities, are omitted.
This isolates the effect of the appearance model, at the cost of not reproducing every reference recipe exactly.
Beyond integrating each model into the fused rasterizer, we improve upon the reference implementations at the kernel level, which we point out in the respective subsections.
\cref{tab:learning_rates} summarizes the appearance learning rates and schedules.
All other hyperparameters are shared and described in \cref{sec:suppl_optimization}.
\begin{table}[t]
\caption{%
Learning rates and schedules of the appearance parameters. \emph{Cosine} follows Miazga~\etal~\cite{miazga2026nasgabor}: decay from iteration 7k toward zero at 37k, \ie, about 13\% of the initial rate remain at the end of optimization. \emph{Delayed cosine} follows Di~Sario~\etal~\cite{disario2026sphericalvoronoi}: frozen until 1k, decay to 0.1\texttimes\ at 27k, frozen after. \emph{Warm-up} follows Mip-NeRF~\cite{barron2021mipnerf}: sinusoidal warm-up over 3k, then log-linear decay to 0.1\texttimes\ at 30k. $\bar{d}$ is the mean distance between each training camera and its three nearest neighbors.
}\label{tab:learning_rates}
\centering
\scriptsize
\setlength\tabcolsep{6.8pt}
\begin{tabular}{llll}
\toprule
Model & Parameter & Learning rate & Schedule \\
\midrule
\multirow{2}{*}{SH}         & base color   & $0.0025 \cdot Y_0$                & constant \\
                            & coefficients & 0.000125                          & constant \\
\midrule
\multirow{4}{*}{SV}         & base color   & 0.0008                            & constant \\
                            & site colors  & 0.0008                            & constant \\
                            & sites        & 0.002                             & delayed cosine \\
                            & temperatures & 0.006                             & delayed cosine \\
\midrule
\multirow{2}{*}{NASG(abor)} & base color   & 0.00025                           & cosine \\
                            & lobes        & $0.0025 \cdot (0.3536/\bar{d})^2$ & cosine \\
\midrule
\multirow{3}{*}{Neural}     & base color   & 0.00025                           & cosine \\
                            & features     & 0.008                             & cosine \\
                            & weights      & 0.003                             & warm-up \\
\bottomrule
\end{tabular}
\end{table}

\subsection{Spherical Harmonics}
The SH model is the standard 3DGS~\cite{kerbl3Dgaussians} parametrization in the common interface of \cref{eq:common_appearance}: $\mathbf{c}_0=\phi^{-1}(\mathbf{c}_\mathrm{sfm})$ and 45 zero-initialized coefficients for degrees one to three, activated at iterations 1000, 2000, and 3000.
As described in \cref{sec:appearance_models}, the constant $Y_0=\nicefrac{1}{\sqrt{4\pi}}\approx0.282$ is folded into the base color, whose learning rate is scaled by $Y_0$ accordingly, while the coefficients use the standard rate of 0.000125.
Both stay constant during optimization.
The model without view-dependent appearance in \cref{tab:training} is this model restricted to degree zero.

\subsection{Spherical Voronoi}
Each site stores seven raw parameters, \ie, an unnormalized site direction, a temperature, and an RGB value, which are mapped to $\mathbf{s}_i=\operatorname{normalize}(\cdot)$, $\tau_i=\exp(\cdot)$, and $\mathbf{a}_i$ with the identity residual activation.
The distance in \cref{eq:spherical_voronoi} is the chord distance $\rho(\mathbf{d},\mathbf{s}_i)=\lVert\mathbf{s}_i-\mathbf{d}\rVert$.
We improve the evaluation of the sites in two respects.
First, the reference implementation~\cite{disario2026sphericalvoronoi} normalizes the sites and exponentiates the temperatures in PyTorch before every rasterizer call, with automatic differentiation handling their gradients in separate passes.
Our kernels instead take the raw parameters and fuse both activations and their derivatives into the forward and backward pass.
Second, the reference evaluates the numerically stable softmax in three passes over the sites, \ie, maximum, normalizer, and weighted sum, each recomputing the distances.
Our forward pass reads the site parameters once and maintains a running maximum, and our backward pass requires two passes.
Following the reference, sites are initialized on a Fibonacci lattice shared by all Gaussians with $\tau_i=1$, all seven sites are active from the first iteration, and sites and temperatures follow the delayed cosine schedule in \cref{tab:learning_rates}.
The reference stores the full color per site, whereas our base color takes the role of their shared component.
With softmax weights $w_i(\mathbf{d})$ that sum to one, both are equivalent:
\begin{equation}
\sum_i w_i(\mathbf{d})\,(\mathbf{c}_0 + \mathbf{a}_i) = \mathbf{c}_0 + \sum_i w_i(\mathbf{d})\,\mathbf{a}_i = \mathbf{c}_0 + \mathcal{A}_{\psi}(\mathbf{d}).
\end{equation}
Accordingly, the base color uses the site color learning rate and is initialized from the SfM color, with all site colors initialized to zero, reproducing the reference initialization exactly.
We follow the reference implementation based on 3DGS-MCMC, which uses Gaussian kernels and propagates the color gradient into the Gaussian mean, whereas their beta-kernel implementation does not.
Unlike the reference, we do not freeze scales, rotations, and opacities during the last 3000 iterations.

\subsection{NASG and NASGabor}
We use the kernel definition, parameter activations, and numerical safeguards of Miazga~\etal~\cite{miazga2026nasgabor} unchanged.
Raw frame angles are mapped by $\tanh$ to the cosines of the lobe orientation, sharpness and anisotropy by a clamped $\exp$, and the Gabor frequency by $20\,(\tanh(\cdot)+1)$, while the lobe amplitudes pass through the $\tanh$ residual activation of \cref{eq:nasg}.
As for SV, we fuse all activations into the kernels, whereas the reference applies the $\tanh$ of the orientation angles and amplitudes in PyTorch and only the exponentials in-kernel.
Initialization also follows the reference, including raw amplitudes of 0.5, so that a lobe contributes a nonzero residual of $\tanh(0.5)$ once it is activated at iteration 1000.
The lobe learning rate is scaled by the reference's scene-adaptive factor derived from the training view spacing (\cf \cref{tab:learning_rates}).
The reference implementation differs from ours in three respects: it renders beta-kernel primitives~\cite{liu2025betasplatting}, composites unclamped and possibly negative colors instead of applying a color activation, and does not propagate the color gradient into the Gaussian mean.
We use Gaussian kernels for all models and prevent negative colors through the color activation, since PPISP~\cite{deutsch2026ppisp} requires non-negative per-pixel radiance and we observed a small improvement from doing so even without PPISP.
Moreover, we derive and implement the missing gradient with respect to the Gaussian mean for both lobe models, so that all models drive densification through the same appearance-dependent gradients.

\subsection{Neural}
The decoder input is assembled as in \cref{eq:mlp_input}: the $F=8$ features are frequency-encoded with $k=1$ into 16 values, and the view direction is encoded by the SH basis of degrees zero to three into 16 values.
The constant degree-zero term acts as a bias input, since the network itself has no biases.
Directional degrees are activated one by one every 1000 iterations following the SH schedule and zeroed while inactive, and the residual is disabled entirely until the first degree activates at iteration 1000.
Hidden layers use the default initialization of tiny-cuda-nn~\cite{tiny-cuda-nn}, whereas the output layer is initialized to zero, so that the residual is exactly zero when it activates.
The features are initialized to zero as well.
The base color and features follow the cosine schedule of the lobe models, whereas the shared weights use the warm-up schedule in \cref{tab:learning_rates}.
For the weights, we further set the second moment coefficient of Adam~\cite{kingma2014adam} to $\beta_2=0.99$ instead of 0.999, since they receive dense gradients from every visible Gaussian in every iteration.

\section{Implementation Details}
\label{sec:suppl_implementation_details}
This section describes the shared optimization protocol (\cref{sec:suppl_optimization}), the differentiable CUDA rasterizer with a focus on the fused evaluation of the neural model (\cref{sec:suppl_cuda}), and the WebGL viewer including its per-model payloads and the frame time measurement (\cref{sec:suppl_webgl}).

\subsection{Optimization}
\label{sec:suppl_optimization}
All appearance models are optimized for 30k iterations with the protocol of Faster-GS~\cite{hahlbohm2026fastergs} using MCMC densification~\cite{kheradmand243dgsmcmc}, and PPISP~\cite{deutsch2026ppisp} is enabled on real scenes.
Following 3DGS, the schedule activates one additional appearance degree, \ie, one SH degree, one lobe, or one directional encoding degree, every 1000 iterations, whereas SV has all sites active from the first iteration as in its reference.
Since sites and temperatures are frozen for the first 1000 iterations, this initially only trains the site colors while the Adam moments of the other parameters warm up.
For all models, the color gradient is propagated through the view direction into the Gaussian mean (\cf \cref{sec:cuda_implementation}).
The model without view-dependent appearance disables this path.
The only change to the fused Adam implementation of Hahlbohm~\etal~\cite{hahlbohm2026fastergs} is that it accepts one learning rate per column of a parameter tensor.
This applies the reference learning rates of SV to the seven parameters of each site without splitting them into separate tensors.

\subsection{Differentiable CUDA Rasterizer}
\label{sec:suppl_cuda}
Each appearance model contributes a pair of device functions, the residual and its backward, against the common interface described in \cref{sec:cuda_implementation}.
The rasterizer combines the selected pair with its preprocess kernels and JIT-compiles the result with the model configuration, \eg, the number of sites or lobes and the activations, as compile-time constants.
Adding a model therefore requires no changes to the rasterization pipeline, and since the compiled kernels are fully specialized, this configurability costs no performance.
The explicit models are evaluated per primitive with scalar arithmetic in single precision and require no further measures.
The neural model is evaluated as a tiny-cuda-nn fully fused MLP~\cite{tiny-cuda-nn} in half precision on Tensor Cores, one warp per 32 Gaussians, which requires all lanes of a warp to participate even if some primitives were culled.
Following M\"uller~\etal~\cite{mueller2022instant}, the weights are stored and updated by Adam in single precision, whereas the forward and backward pass operate on a half-precision copy created before each iteration.
The weight gradient is accumulated in half precision using a loss scale of 128.
As a power of two, the scale only shifts the floating-point exponent and is removed without rounding error when the gradient is converted back to single precision.
The gradients with respect to the features and the base color are computed in single precision.
The forward pass stores the layer activations of each Gaussian for the backward pass.
We validate all fused implementations against the reference path that evaluates the appearance model in a separate pass.
Images and gradients agree to within floating-point accumulation error, and for the neural model, the last two rows of \cref{tab:neural_architecture} show that both paths yield the same quality.

\subsection{WebGL Viewer}
\label{sec:suppl_webgl}
Each frame, the prepass of our Three.js-based viewer evaluates the color of every Gaussian in a fragment shader and writes it to a texture with one 8-bit RGBA texel per Gaussian.
The subsequent splatting pass reads this color and alpha-composites the Gaussians in the depth order computed asynchronously on the CPU, following SuperSplat~\cite{supersplat2024}.
Geometry is packed into 16 bytes per Gaussian following Spark~\cite{spark2025}: a half-precision mean, 8-bit logarithmic scale codes, a 24-bit octahedral-folded quaternion, and the 8-bit base color and opacity.
The per-model payloads in \cref{tab:webgl_payloads} therefore hold only the view-dependent parameters, stored in 32-bit integer textures that pack eight half-precision values or bit-packed coefficient codes per texel.
All view-independent computations, \ie, parameter activations and normalization constants, are precomputed when a model is exported for the viewer, so the per-frame work of every model is only its direction-dependent evaluation.
The residual is composed with the base color through the same activation $\phi$ as in CUDA.
Notably, the neural model is the only one whose residual is evaluated at the same precision as in CUDA.
Its features and weights are stored in half precision in both implementations, and the shader arithmetic on mobile GPUs is half precision as well.
The explicit models are evaluated in single precision in CUDA but stored in half precision for the viewer, except for SH, whose coefficients use an elaborate quantization scheme (see below).
We leave proposing and evaluating similarly elaborate quantization schemes for the other models to future work.

\paragraph{SH}
The coefficients are quantized to the packed layout of Spark~\cite{spark2025}.
Degree one stores its nine coefficients at 7 bits in a two-channel texel, and degrees two and three store their 15 and 21 coefficients at 8 and 6 bits in one four-channel texel each.
Quantization is symmetric with one shared scale per degree, the maximum coefficient magnitude across the scene, which is stored once per scene and used for dequantization in the shader.

\paragraph{SV}
Sites are stored normalized with activated temperatures and colors, and the shader evaluates the softmax in a single pass with a running maximum as in CUDA.

\paragraph{NASG and NASGabor}
Lobes are stored fully activated as the two frame axes, the shape parameters, the precomputed normalization constant, and the amplitude.
The per-frame evaluation of a lobe thus reduces to two dot products, one power, and one exponential, plus one cosine for NASGabor.

\paragraph{Neural}
The network is evaluated as in SMERF~\cite{duckworth2023smerf}: weights are stored as half-precision texels holding four consecutive input weights of one neuron, and each layer accumulates one 4\texttimes4 matrix product per block of four outputs.
The weights reside in a uniform array when the device's uniform limit permits and otherwise in a half-precision texture.
Since the network is shared by all Gaussians, its weights stay cache-resident throughout the pass.
We propose an additional optimization that exploits the view-independence of the 16 encoded features and the constant degree-zero input.
Their 17 columns of the first weight matrix are folded into a per-Gaussian vector of 16 half-precision pre-activations, one per neuron, which is stored in place of the features at the same size.
Only the 15 columns of the direction-dependent degrees are kept, padded to 16.
The shader thus evaluates only the direction-dependent part of the first layer, reducing the per-Gaussian cost from 816 to 560 multiply-accumulate operations, \ie, by 31\%.
This precomputation only pays off while its result is no larger than the features it replaces.
With 32 neurons, it would store 32 values per Gaussian, twice the 16 encoded feature values, so we store the features and the full first layer instead.
This is why the 32-neuron variant in \cref{tab:neural_architecture} lists the full count.

\paragraph{Frame Time Measurement}
For \cref{tab:inference}, the viewer renders the test views of each scene at 1280\texttimes720 with the native camera intrinsics.
After a warm-up pass over all views, it waits for the asynchronous depth sort of each view to complete and renders once untimed before timing repeated renders, so that sorting and the upload of the sort order are excluded and the measurements reflect the steady-state cost at a novel view.
The reported frame time is the mean over all test views and all repeated renders of each view.
On the iPad and iPhone, sustained rendering triggers an unavoidable slowdown after a few dozen seconds, which is why we render each view only 10 times on these devices, whereas all other frame time and FPS measurements use 100 renders per view.
\begin{table}[t]
\caption{%
Per-Gaussian appearance payload of the WebGL viewer for the default configuration of each model and the per-frame work of the shader. Byte counts are padded to whole 16-byte texels, the base color is packed with the geometry, and the contents of each payload are described in the text.
}\label{tab:webgl_payloads}
\centering
\scriptsize
\setlength\tabcolsep{4.5pt}
\begin{tabular}{llcl}
\toprule
Model & Payload & Bytes$\slash$G & Per-frame work \\
\midrule
SH       & 45 coefficients at 7$\slash$8$\slash$6 bit & 40  & SH basis, weighted sum  \\
SV       & 7 sites \texttimes\ 7 fp16                 & 112 & site distances, softmax \\
NASG     & 1 lobe \texttimes\ 12 fp16                 & 32  & lobe response          \\
NASGabor & 1 lobe \texttimes\ 13 fp16                 & 32  & lobe response, carrier \\
Neural   & 16 fp16 pre-activations                    & 32  & MLP                    \\
\bottomrule
\end{tabular}
\end{table}

\section{Additional Experiments}
\begin{table*}[t]
\caption{%
Quality metrics across Mip-NeRF~360 scenes~\cite{barron2022mipnerf360} with different densification strategies and PPISP~\cite{deutsch2026ppisp}. \#Gs refers to the number of potentially visible Gaussians in the final model, \ie, those with opacity $\ge\nicefrac{1}{255}$ after training.
}\label{tab:densification}
\centering
\setlength\tabcolsep{7.4pt}
\scriptsize
\begin{tabular}{lcccccccccccc}
\toprule
 & \multicolumn{4}{c}{SH} & \multicolumn{4}{c}{NASGabor} & \multicolumn{4}{c}{Neural} \\
\cmidrule(lr){2-5} \cmidrule(lr){6-9} \cmidrule(lr){10-13}
Method & PSNR$^\uparrow$ & SSIM$^\uparrow$ & LPIPS$^\downarrow$ & \#Gs$^\downarrow$ & PSNR$^\uparrow$ & SSIM$^\uparrow$ & LPIPS$^\downarrow$ & \#Gs$^\downarrow$ & PSNR$^\uparrow$ & SSIM$^\uparrow$ & LPIPS$^\downarrow$ & \#Gs$^\downarrow$ \\
\midrule
MCMC \cite{kheradmand243dgsmcmc}                     & 28.35 & \cellcolor{1st}0.836 & \cellcolor{1st}0.222 & 2.91M & \cellcolor{2nd}28.51 & \cellcolor{1st}0.836 & \cellcolor{2nd}0.223 & 2.90M & \cellcolor{1st}28.53 & \cellcolor{2nd}0.831 & 0.227 & 2.96M \\
~$\hookrightarrow$ with 0.1$\times$\#Gs              & \cellcolor{2nd}27.30 & \cellcolor{1st}0.798 & \cellcolor{1st}0.309 & 0.32M & \cellcolor{1st}27.32 & \cellcolor{2nd}0.796 & \cellcolor{2nd}0.312 & 0.32M & \cellcolor{2nd}27.30 & 0.793 & 0.316 & 0.32M \\
ADC \cite{kerbl3Dgaussians}                          & 27.88 & \cellcolor{1st}0.818 & \cellcolor{1st}0.250 & 2.79M & \cellcolor{1st}28.14 & \cellcolor{1st}0.818 & \cellcolor{2nd}0.252 & 2.80M & \cellcolor{2nd}28.04 & \cellcolor{2nd}0.812 & 0.258 & 2.88M \\
~$\hookrightarrow$ with \cite{hanson2025speedysplat} & 27.23 & \cellcolor{2nd}0.777 & \cellcolor{2nd}0.346 & 0.33M & \cellcolor{1st}27.29 & \cellcolor{1st}0.779 & \cellcolor{1st}0.343 & 0.33M & \cellcolor{2nd}27.24 & \cellcolor{2nd}0.777 & 0.349 & 0.34M \\
\bottomrule
\end{tabular}
\end{table*}

This section extends the evaluation of the main paper with an ablation of the densification strategy (\cref{sec:suppl_densification}), an ablation of the neural model architecture (\cref{sec:suppl_architecture}), qualitative examples of the decomposition into base color and residual (\cref{sec:suppl_decomposition}), and per-scene results with additional metrics (\cref{sec:suppl_per_scene}).
We first provide additional details on how the numbers reported in the main paper and in this section are obtained.
Unless stated otherwise, all experiments follow the setup of \cref{sec:evaluation}, \ie, MCMC densification~\cite{kheradmand243dgsmcmc} with scene-specific primitive budgets~\cite{liu2025betasplatting} and PPISP~\cite{deutsch2026ppisp} on real scenes.
Since image metrics vary between training runs, \cref{tab:training} and the per-scene results in \cref{tab:per_scene_m360,tab:per_scene_tandt_db,tab:per_scene_synthetic} report averages over five runs, whereas all other tables report a single run.
We measure optimization time, peak memory, and frame rates only during the first run, which we execute for all methods on the same workstation with an RTX~4090.
The reported optimization times exclude the PPISP controller distillation, which runs after optimization and is identical for all models.
The remaining runs are distributed across shared compute servers, where concurrent jobs compete for CPU resources and render timings unreliable.
As timings and memory consumption vary far less between runs than image metrics, a single measurement is sufficient for these quantities.

\subsection{Densification Strategies}
\label{sec:suppl_densification}
To verify that our findings are not tied to the densification strategy, \cref{tab:densification} repeats the Mip-NeRF~360 comparison for SH, NASGabor, and the neural model with the adaptive density control (ADC) of 3DGS~\cite{kerbl3Dgaussians}.
We further include two low-budget variants: MCMC with a tenth of the primitive budget and ADC combined with the pruning of Speedy-Splat~\cite{hanson2025speedysplat}.
Since the two high-budget and the two low-budget configurations arrive at almost identical numbers of visible Gaussians, this also compares the densification strategies at equal model size.
The relative ranking of the appearance models is stable across densification strategies: NASGabor and the neural model match or exceed SH in PSNR, whereas SH and NASGabor share the best SSIM and LPIPS, mirroring \cref{tab:training}.
The differences between appearance models shrink with the primitive budget, as reconstruction quality becomes limited by geometric coverage, which no amount of angular expressivity can compensate.
Conversely, the advantage of the expressive models is largest with ADC.
A plausible explanation is that ADC decides where to clone and split primitives based on positional gradients, which include the directional gradients propagated through the appearance model (\cf \cref{sec:cuda_implementation}), whereas MCMC relocates primitives based on opacity alone.
Finally, MCMC outperforms ADC at equal model size for all appearance models.
Its advantage in SSIM and LPIPS persists at the low budget, as opacity-driven relocation distributes a limited budget more evenly than a pruned ADC model.

\begin{table*}[t]
\caption{%
Comparison of different model configurations for the neural appearance model across Mip-NeRF~360 scenes~\cite{barron2022mipnerf360}. Our default configuration uses two hidden layers with 16 neurons and $F\!=\!8$ frequency-encoded input features. MAC$\slash$G is the number of multiply-accumulate operations needed to evaluate the model for a single Gaussian in our WebGL implementation.
}\label{tab:neural_architecture}
\centering
\setlength\tabcolsep{3.9pt}
\scriptsize
\begin{tabular}{lcccccccccccccc}
\toprule
 & \multicolumn{3}{c}{PSNR$^\uparrow$} & \multicolumn{3}{c}{SSIM$^\uparrow$} & \multicolumn{3}{c}{LPIPS$^\downarrow$} & \multirow{2}{*}{Train$^\downarrow$\vspace{-4pt}} & \multirow{2}{*}{VRAM$^\downarrow$\vspace{-4pt}} & \multirow{2}{*}{FPS$^\uparrow$\vspace{-4pt}} & \multirow{2}{*}{Bytes$\slash$G\vspace{-4pt}} & \multirow{2}{*}{MAC$\slash$G\vspace{-4pt}} \\
\cmidrule(lr){2-4} \cmidrule(lr){5-7} \cmidrule(lr){8-10}
Configuration & Outdoor & Indoor & All & Outdoor & Indoor & All & Outdoor & Indoor & All & & & & & \\
\midrule
1 hidden layer                           & \cellcolor{2nd}25.37 & 32.15 & 28.39 & \cellcolor{1st}0.753 & \cellcolor{2nd}0.933 & \cellcolor{1st}0.833 & \cellcolor{1st}0.232 & \cellcolor{3rd}0.219 & \cellcolor{1st}0.226 & \cellcolor{1st}4m14s  & \cellcolor{1st}5.2GiB & 432.8 & \cellcolor{1st}28 & \cellcolor{1st}304  \\
3 hidden layers                          & 25.25 & \cellcolor{1st}32.71 & \cellcolor{2nd}28.57 & 0.745 & \cellcolor{1st}0.935 & 0.829 & 0.238 & \cellcolor{1st}0.217 & \cellcolor{3rd}0.229 & \cellcolor{3rd}4m24s  & \cellcolor{3rd}5.4GiB & 435.7 & \cellcolor{1st}28 & \cellcolor{3rd}816  \\
32 neurons$\slash$layer                  & \cellcolor{1st}25.38 & \cellcolor{2nd}32.67 & \cellcolor{1st}28.62 & \cellcolor{2nd}0.749 & \cellcolor{1st}0.935 & \cellcolor{2nd}0.832 & \cellcolor{2nd}0.234 & \cellcolor{1st}0.217 & \cellcolor{1st}0.226 & 4m29s  & 5.5GiB & \cellcolor{1st}449.0 & \cellcolor{1st}28 & 2144 \\
$F\!=\!16$ w$\slash$o frequency encoding & 25.31 & \cellcolor{3rd}32.63 & \cellcolor{3rd}28.56 & \cellcolor{3rd}0.748 & \cellcolor{1st}0.935 & \cellcolor{3rd}0.831 & 0.236 & \cellcolor{1st}0.217 & \cellcolor{2nd}0.227 & 4m42s  & 5.7GiB & \cellcolor{3rd}438.5 & \cellcolor{2nd}44 & \cellcolor{2nd}560  \\
Default w$\slash$o fusion                & 25.34 & 32.58 & \cellcolor{3rd}28.56 & \cellcolor{3rd}0.748 & \cellcolor{1st}0.935 & \cellcolor{3rd}0.831 & \cellcolor{2nd}0.234 & \cellcolor{2nd}0.218 & \cellcolor{2nd}0.227 & 16m08s & 5.7GiB & 112.1 & \cellcolor{1st}28 & \cellcolor{2nd}560  \\
Default                                  & \cellcolor{3rd}25.35 & 32.51 & 28.53 & \cellcolor{3rd}0.748 & \cellcolor{1st}0.935 & \cellcolor{3rd}0.831 & \cellcolor{3rd}0.235 & \cellcolor{2nd}0.218 & \cellcolor{2nd}0.227 & \cellcolor{2nd}4m17s  & \cellcolor{2nd}5.3GiB & \cellcolor{2nd}443.6 & \cellcolor{1st}28 & \cellcolor{2nd}560  \\
\bottomrule
\end{tabular}
\end{table*}

\subsection{Neural Model Architecture}
\label{sec:suppl_architecture}
\cref{tab:neural_architecture} varies the architecture of the neural appearance model.
Starting from the default decoder (\cf \cref{sec:neural_appearance}), we vary depth, width, and input encoding, where $F=16$ raw features without frequency encoding yield the same 16 inputs, and additionally evaluate the default configuration through the unfused reference path.
Besides image metrics, optimization time, peak memory, and CUDA frame rate on the RTX~4090, we report the appearance footprint per Gaussian and the number of multiply-accumulate operations (MAC) needed to decode a single Gaussian.
The latter is the relevant cost for the WebGL viewer since it cannot use Tensor Cores.

\paragraph{Capacity}
All fused configurations achieve similar quality, so the decoder is far from being the limiting factor.
The differences that do exist are concentrated in the indoor scenes, which contain the glossy surfaces for which angular resolution matters (\cf \cref{fig:bonsai_block}).
Outdoor scenes are dominated by diffuse content, and their quality is limited by geometry and observation inconsistency rather than by appearance capacity (\cf \cref{sec:discussion}).
Accordingly, a single hidden layer loses quality indoors, while wider or deeper decoders gain.
Doubling the latent code without frequency encoding performs on par with a deeper decoder but is the only configuration that increases the per-Gaussian footprint.
Additional capacity in the shared decoder is free in terms of storage and memory traffic, whereas additional per-primitive features are paid for by every Gaussian.
We therefore consider wider or deeper decoders the preferable way of scaling the model.

\paragraph{Cost}
Optimization time and peak memory vary only marginally across the fused configurations, as the decoder accounts for a small fraction of the per-iteration cost.
CUDA frame rates behave counterintuitively: the widest decoder is the fastest to render despite requiring almost four times as many MACs as the default.
In the fused CUDA renderer, the decoder is evaluated once per visible Gaussian on Tensor Cores in the preprocess kernel, whose cost is small compared to the subsequent blending, which scales with the number of pixel--Gaussian pairs.
We hypothesize that a more expressive decoder relieves the geometry from compensating for missing angular capacity, resulting in more compact primitives and fewer blending operations, consistent with the geometric differences observed in \cref{fig:bonsai_block}.
In WebGL, in contrast, the MAC count translates directly into the cost of the appearance prepass, which matters most on the mobile GPUs in \cref{tab:inference}.
The default configuration balances both deployment targets, while a single hidden layer offers an even cheaper option for mobile GPUs at a small loss in indoor scenes.

\paragraph{Fusion}
The unfused reference path computes the appearance in a separate pass and exchanges inputs and outputs through global memory.
Evaluating the default architecture through it yields identical quality within run-to-run variation, confirming the correctness of the fused forward and backward passes.
However, it is roughly four times slower to optimize and render, and requires more memory.
This gap is what makes the neural model practical, and it explains the common assumption that neural decoding is too expensive for real-time rendering.

\subsection{Appearance Decomposition}
\label{sec:suppl_decomposition}
\newlength{\dcgap}       \setlength{\dcgap}{1pt}    %
\newlength{\dcblockgap}  \setlength{\dcblockgap}{3pt}  %
\newlength{\dcsep}       \setlength{\dcsep}{1pt}    %
\newlength{\dclabelw}    \setlength{\dclabelw}{6.5pt}
\newlength{\dctmp}       \setlength{\dctmp}{\dimexpr\textwidth-\dclabelw-\dcsep-6\dcgap\relax}
\newlength{\dcpanelw}    \setlength{\dcpanelw}{0.142857\dctmp}
\newcommand{\dcimg}[1]{\includegraphics[width=\dcpanelw]{resources/decomposition_examples/#1.jpg}}
\newlength{\dcpanelh}    \settoheight{\dcpanelh}{\dcimg{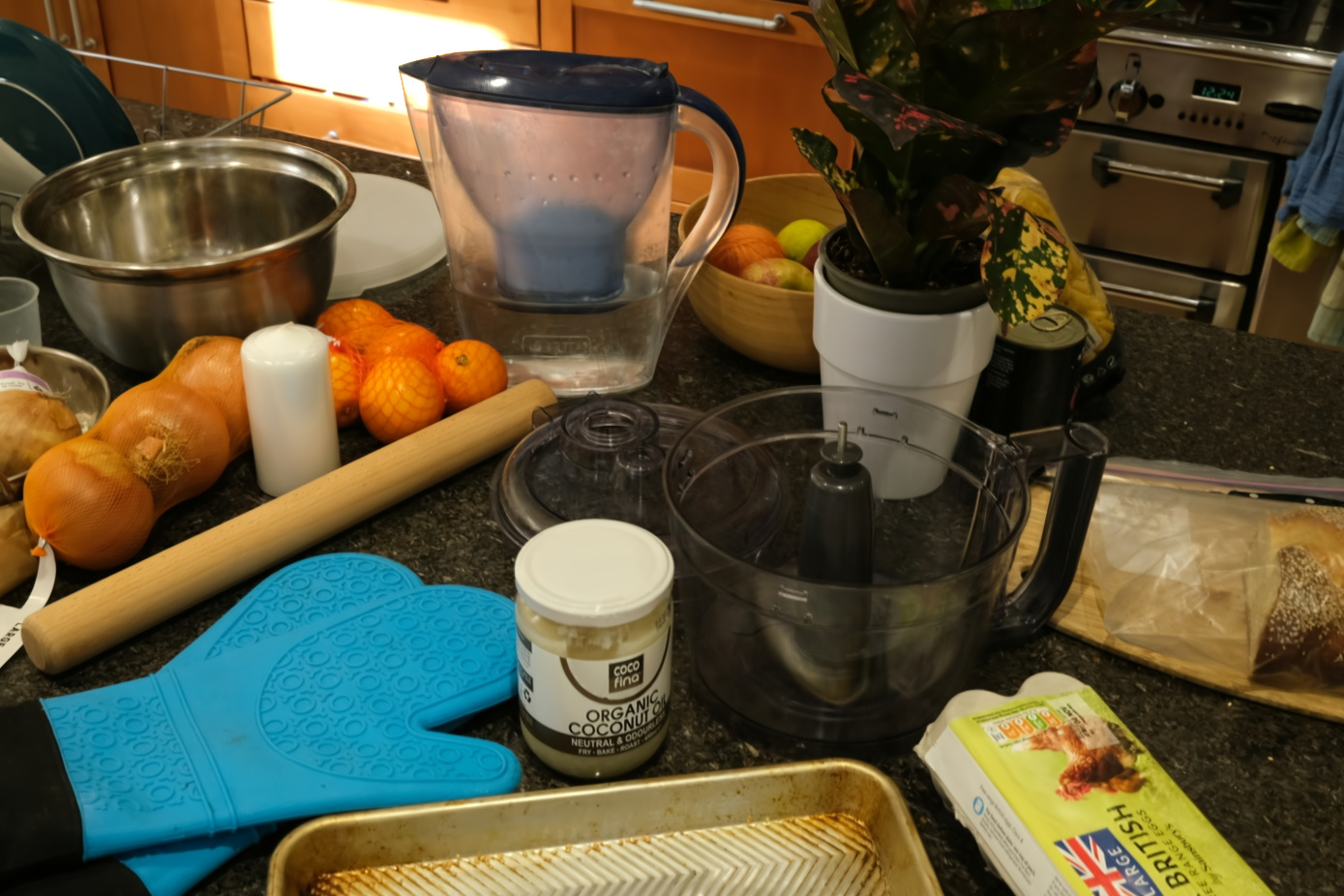}}
\newlength{\dcrefh}      \setlength{\dcrefh}{\dimexpr3\dcpanelh+2\dcgap\relax}
\newcommand{\dcref}[1]{\includegraphics[height=\dcrefh]{resources/decomposition_examples/#1}}
\newcommand{\dcheader}[2]{\node[anchor=south, inner sep=2pt, font=\scriptsize] at (#1.north) {#2};}
\newcommand{\dcrowlabel}[2]{%
\node[rotate=90, anchor=south, inner sep=1pt, font=\tiny, text height=3.5pt, text depth=1pt] at (#1.west) {#2};}
\newcommand{\dcscenelabel}[2]{\node[ovlabel, anchor=south east, font=\tiny] at (#1.south east) {#2};}

\begin{figure*}[p]
\centering
\noindent\makebox[\textwidth][c]{%
\begin{tikzpicture}
\node[ovpanel, anchor=north west] (b0r0c0) at (\dclabelw,0) {\dcimg{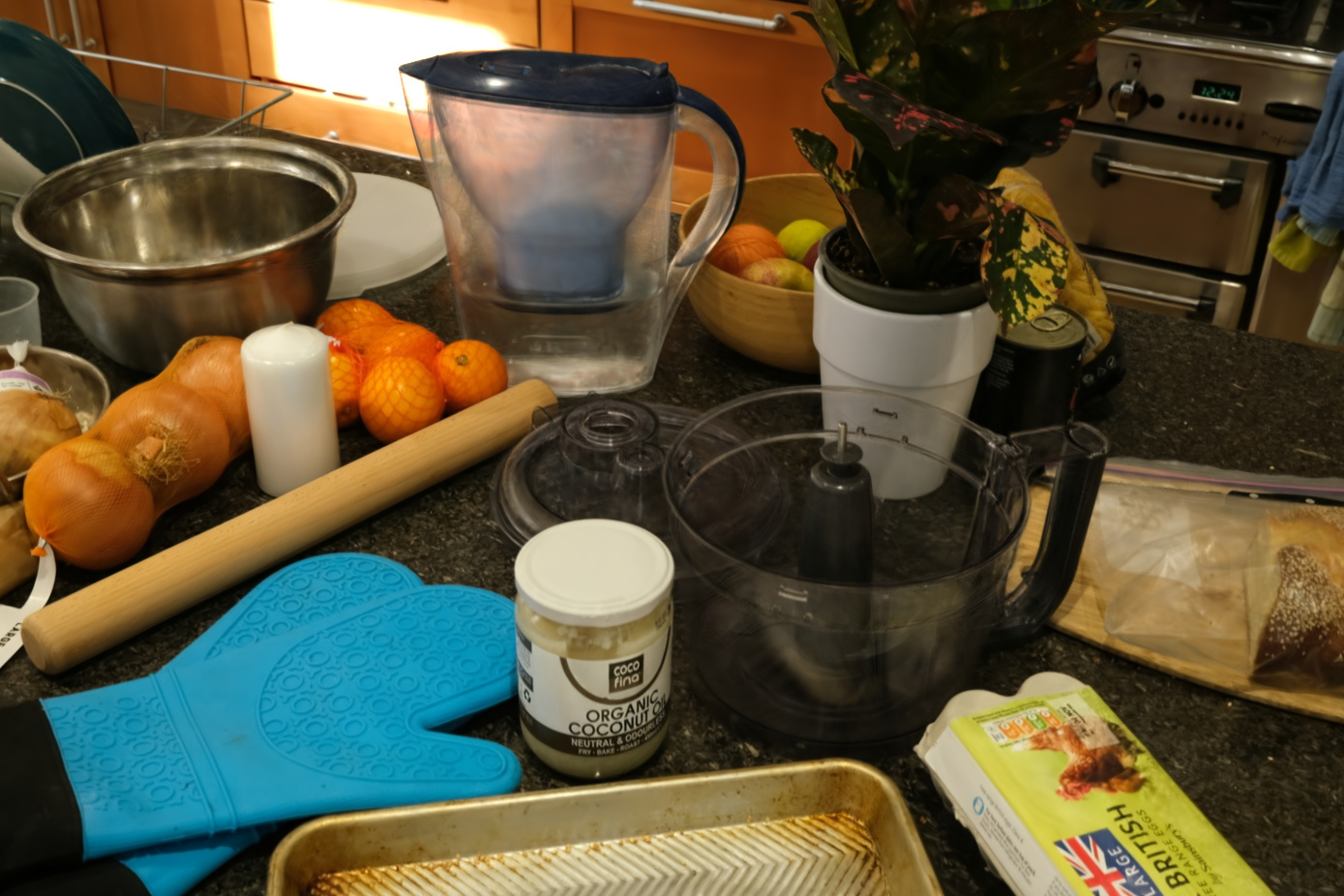}};
\node[ovpanel, anchor=north west, xshift=\dcgap] (b0r0c1) at (b0r0c0.north east) {\dcimg{counter_sh_full}};
\node[ovpanel, anchor=north west, xshift=\dcgap] (b0r0c2) at (b0r0c1.north east) {\dcimg{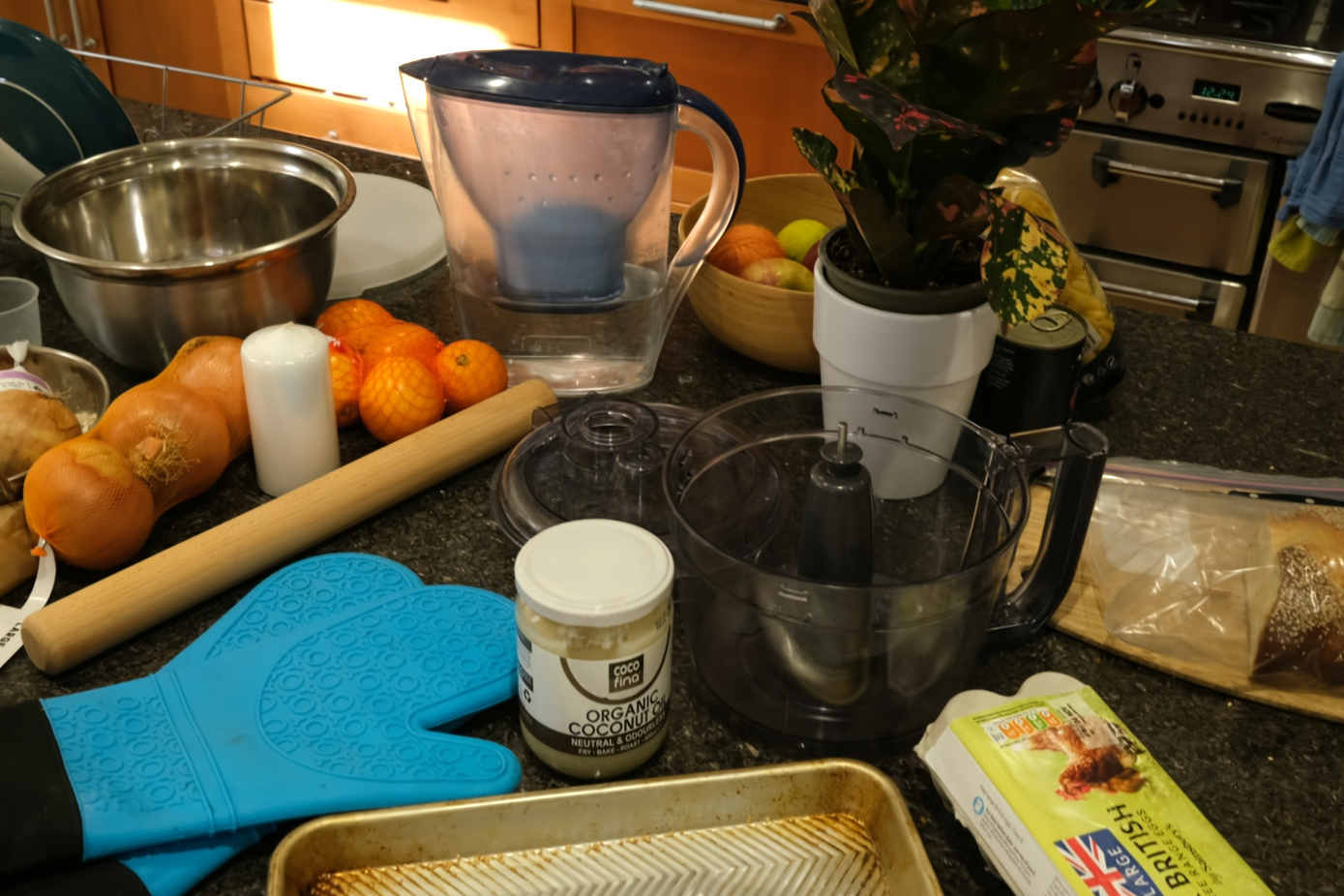}};
\node[ovpanel, anchor=north west, xshift=\dcgap] (b0r0c3) at (b0r0c2.north east) {\dcimg{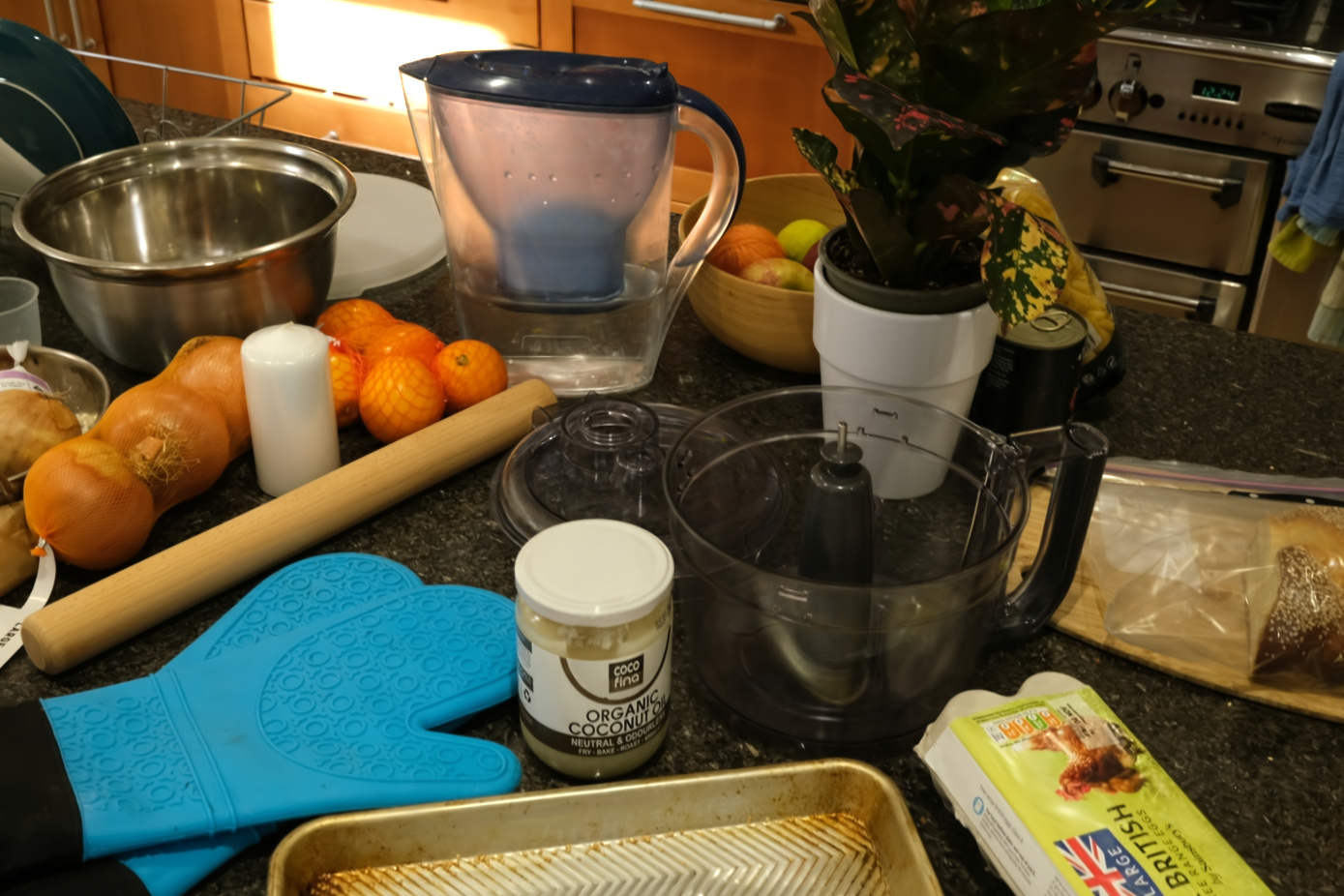}};
\node[ovpanel, anchor=north west, xshift=\dcgap] (b0r0c4) at (b0r0c3.north east) {\dcimg{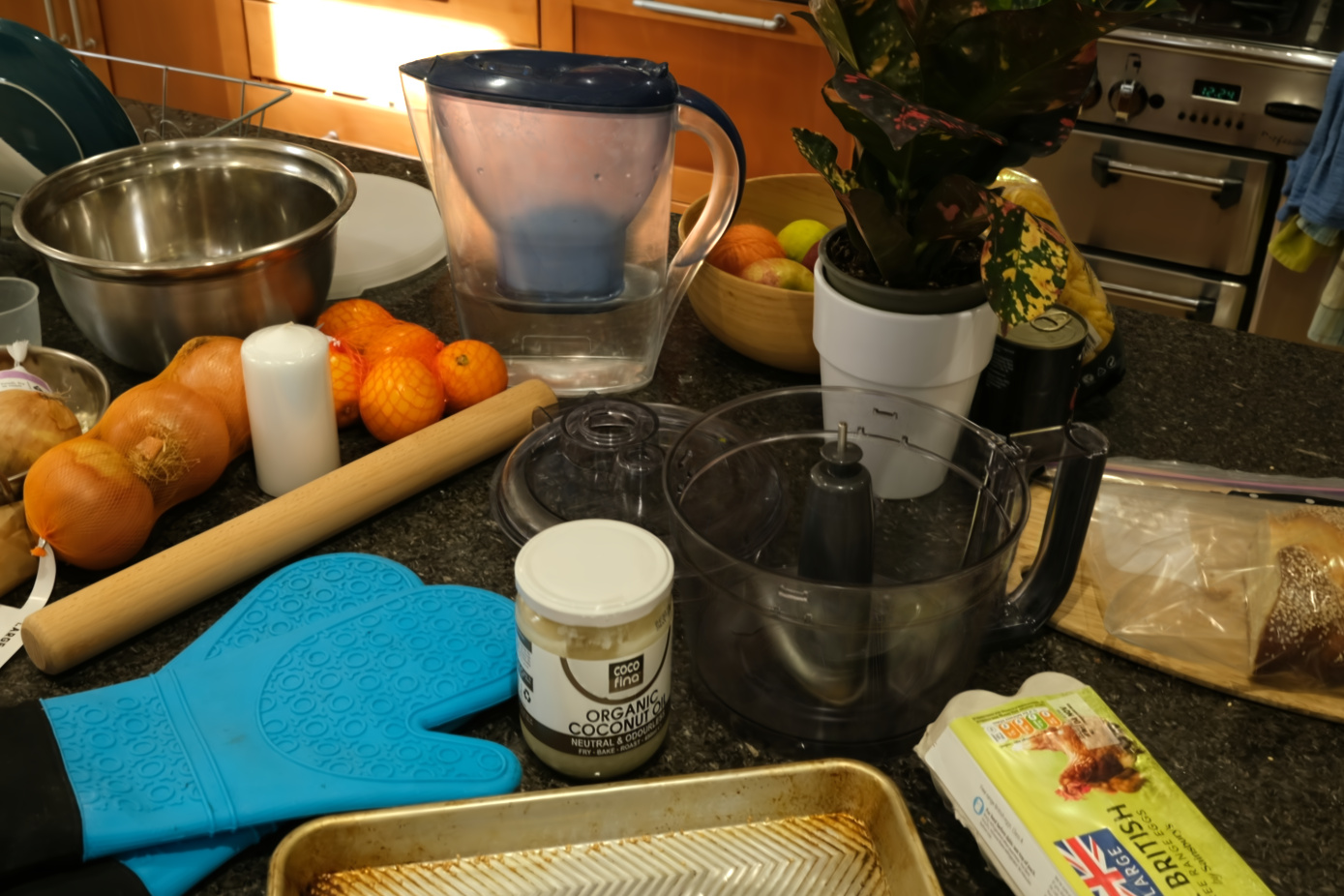}};
\node[ovpanel, anchor=north west, xshift=\dcgap] (b0r0c5) at (b0r0c4.north east) {\dcimg{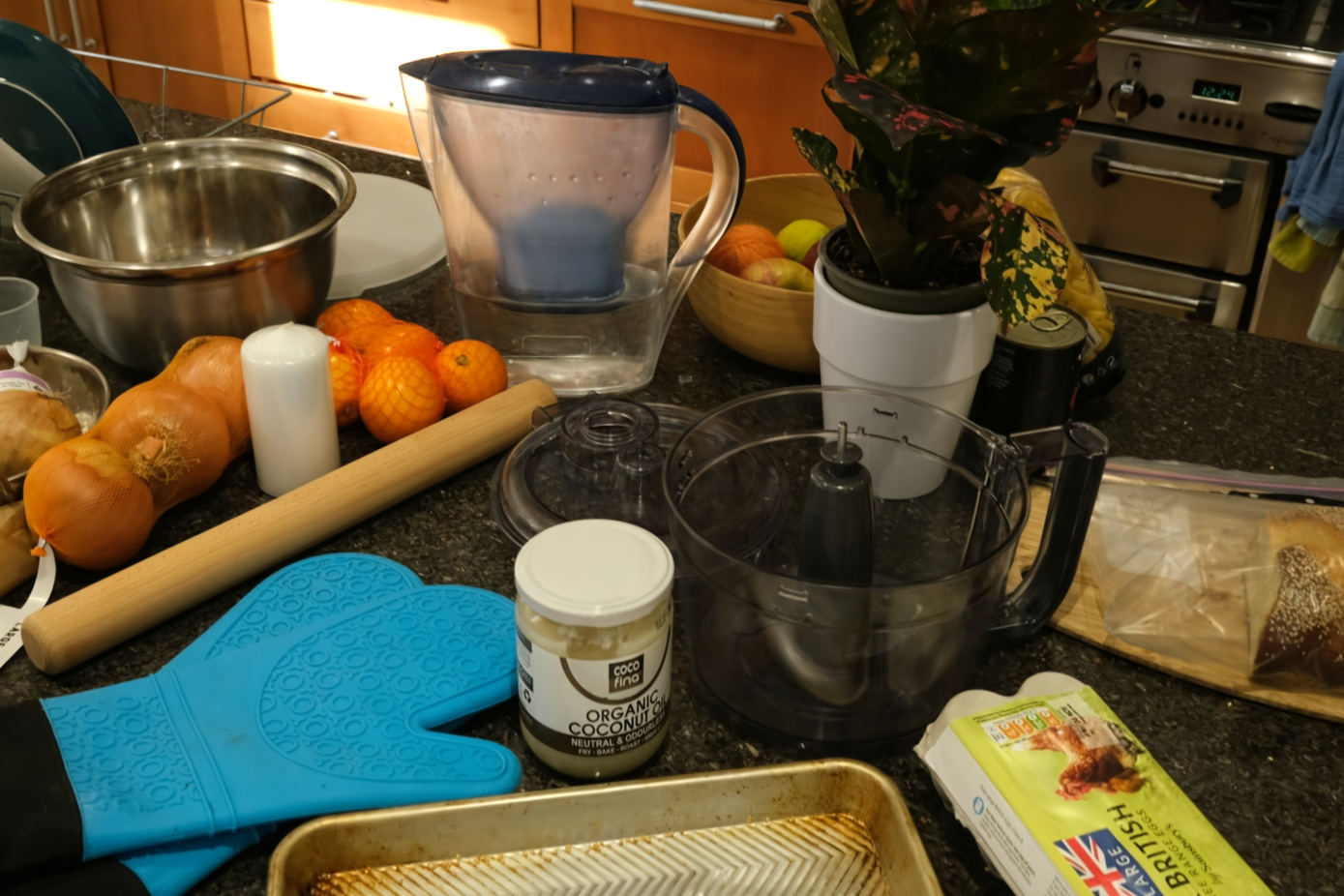}};
\node[ovpanel, anchor=north west, yshift=-\dcgap] (b0r1c0) at (b0r0c0.south west) {\dcimg{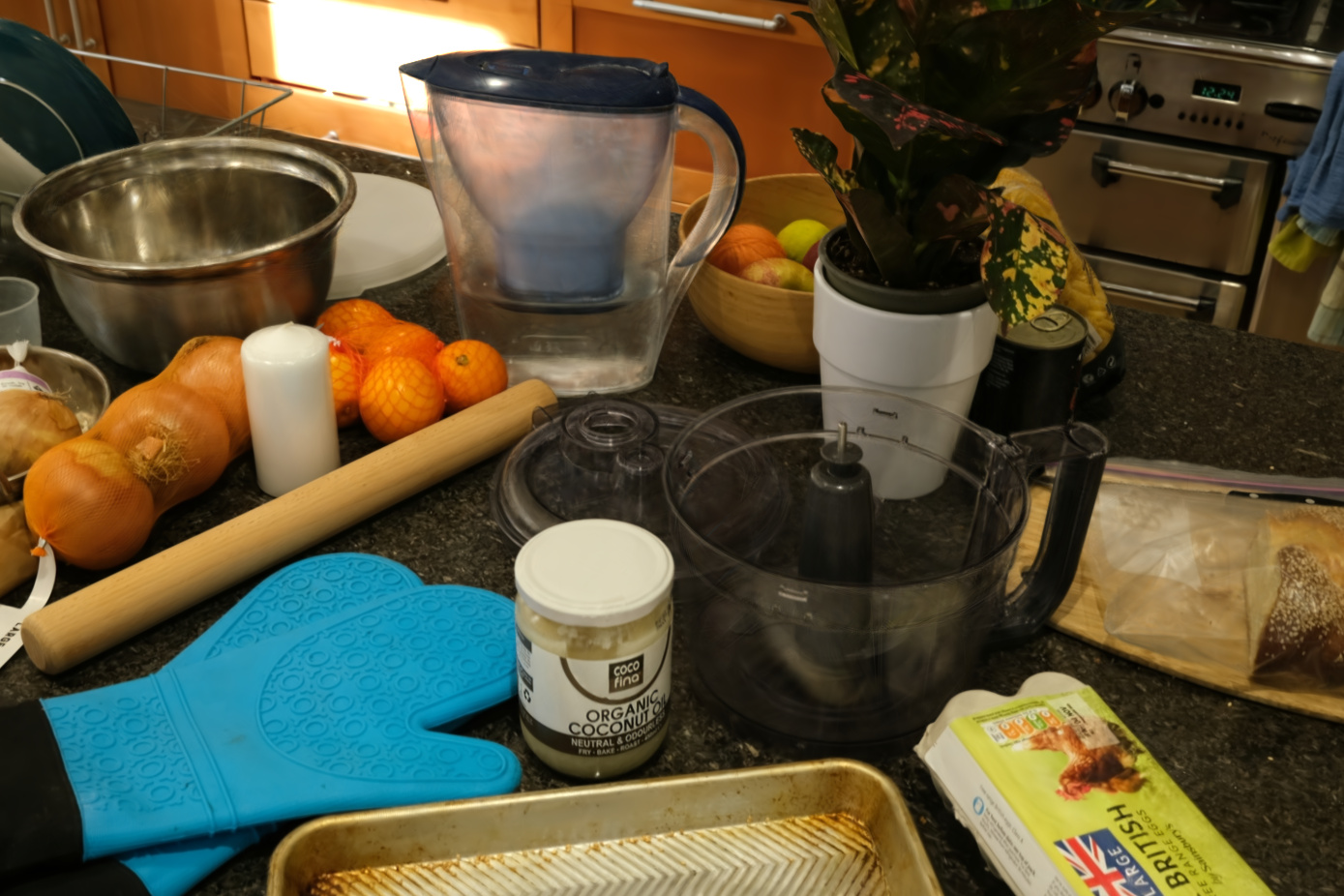}};
\node[ovpanel, anchor=north west, xshift=\dcgap] (b0r1c1) at (b0r1c0.north east) {\dcimg{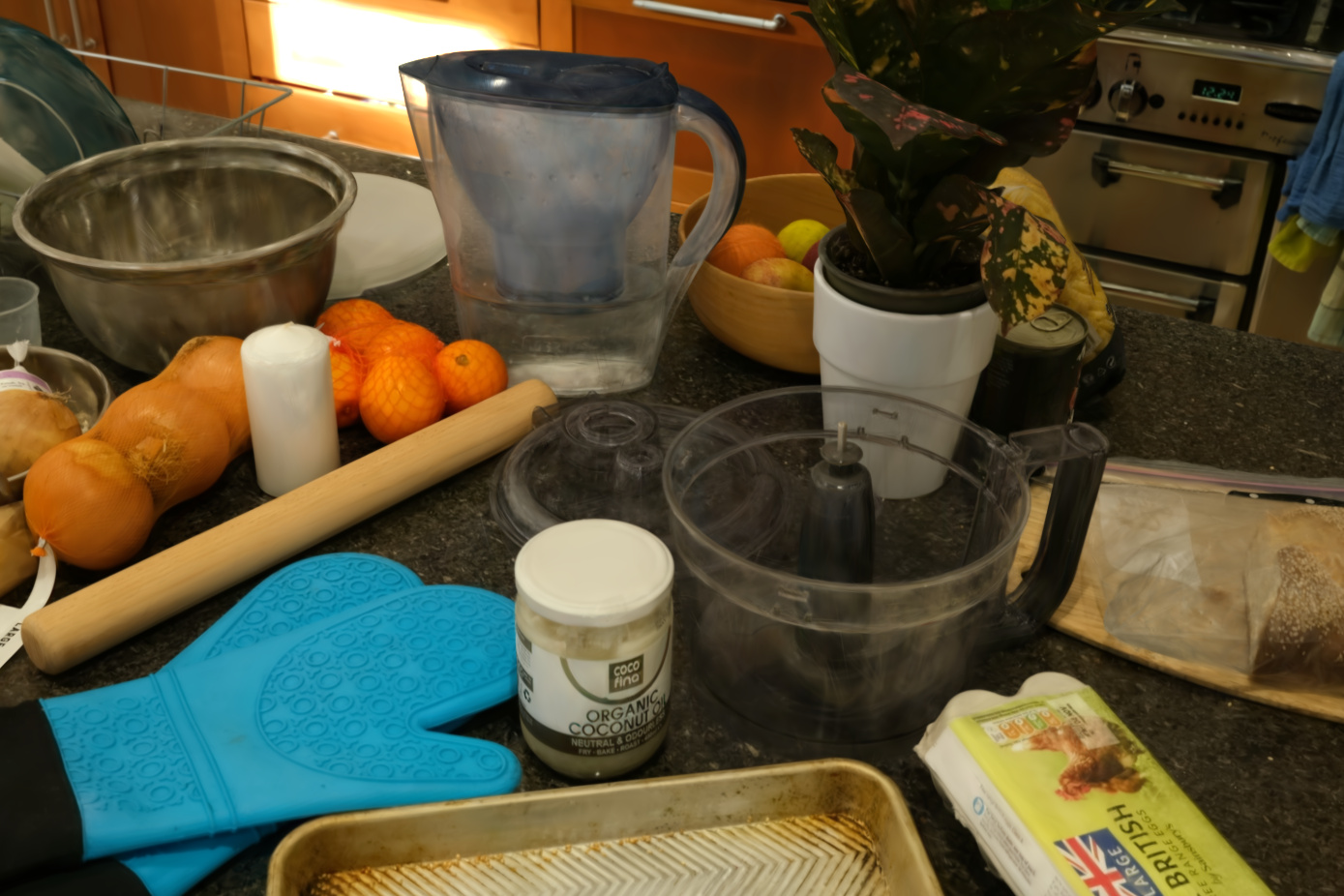}};
\node[ovpanel, anchor=north west, xshift=\dcgap] (b0r1c2) at (b0r1c1.north east) {\dcimg{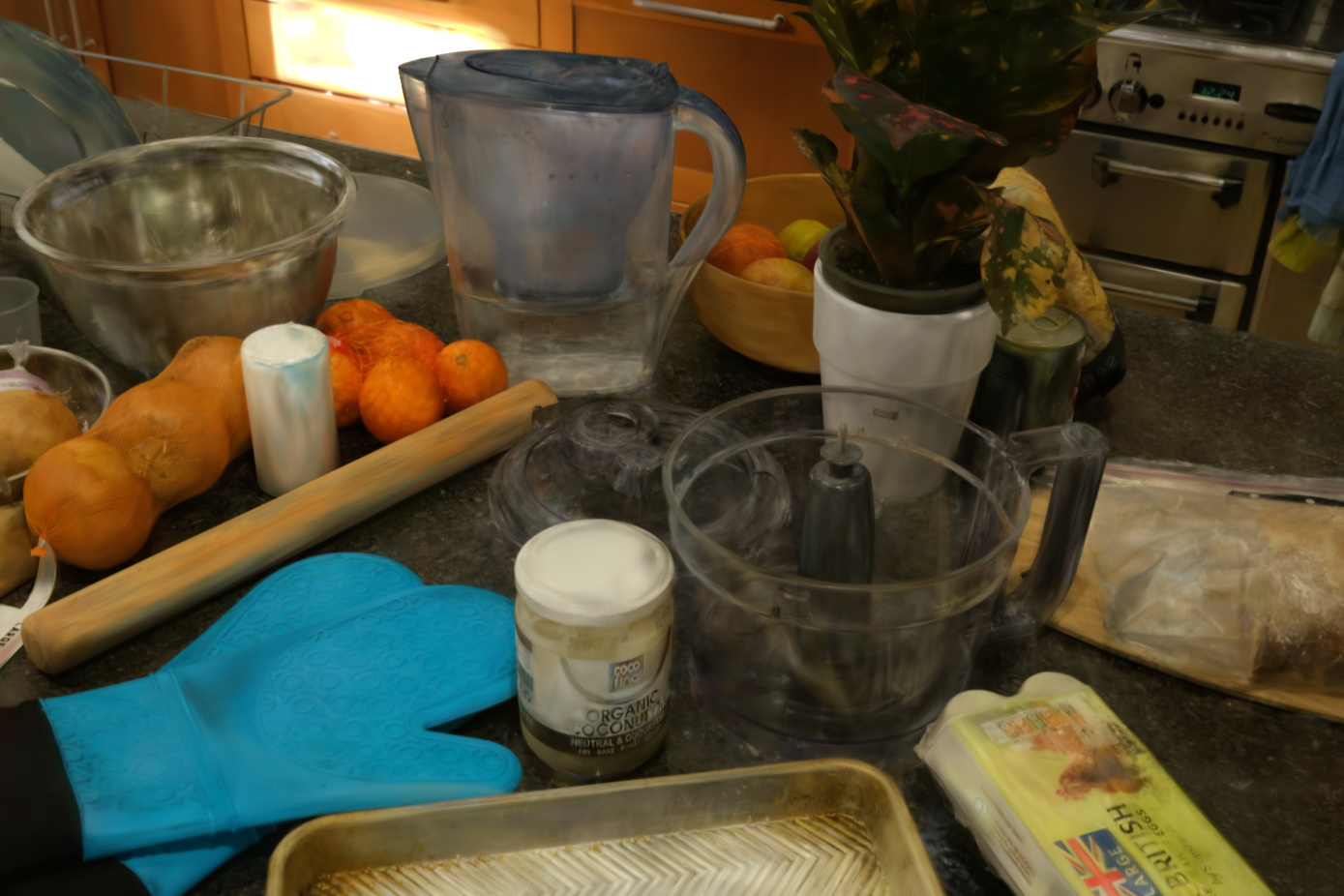}};
\node[ovpanel, anchor=north west, xshift=\dcgap] (b0r1c3) at (b0r1c2.north east) {\dcimg{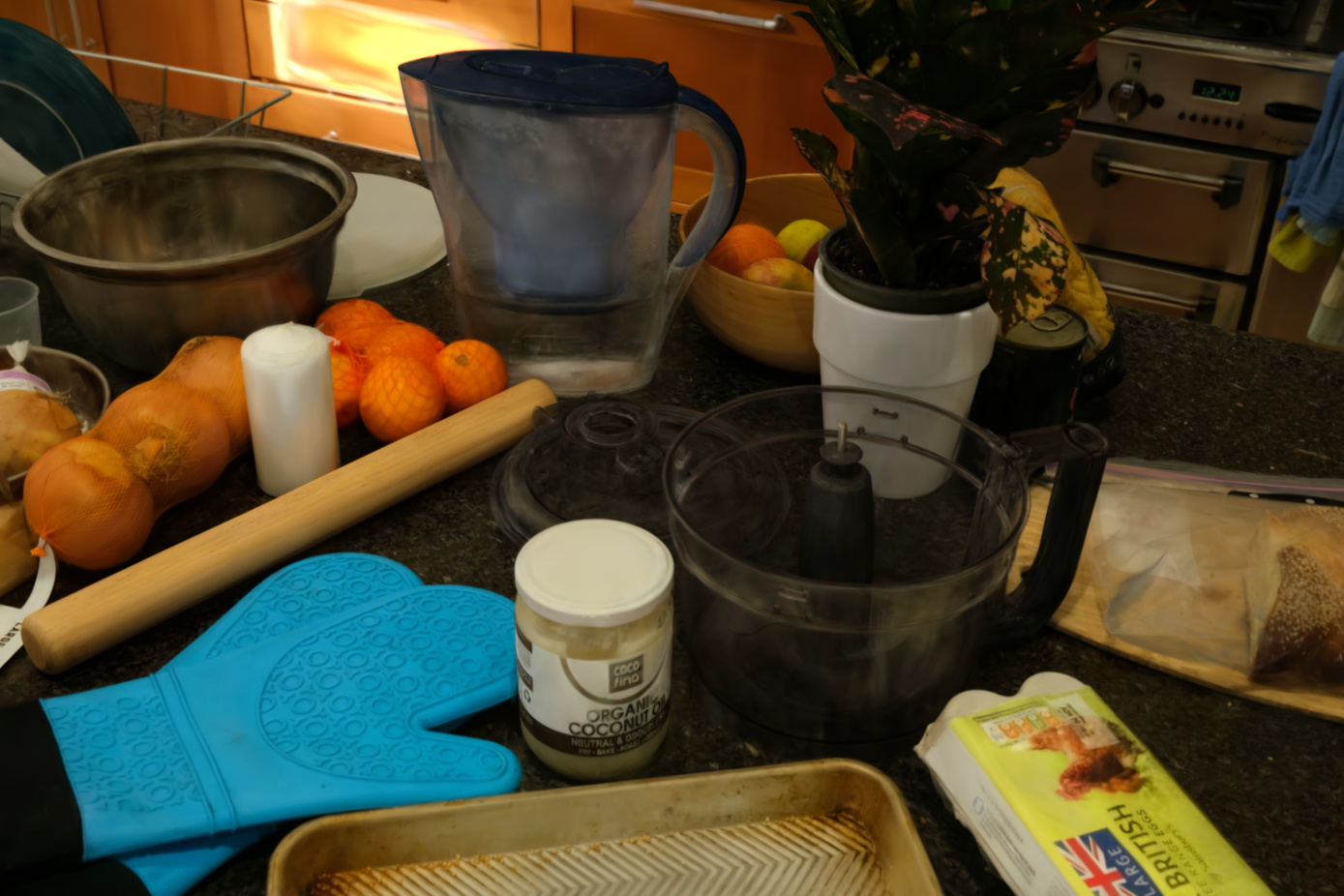}};
\node[ovpanel, anchor=north west, xshift=\dcgap] (b0r1c4) at (b0r1c3.north east) {\dcimg{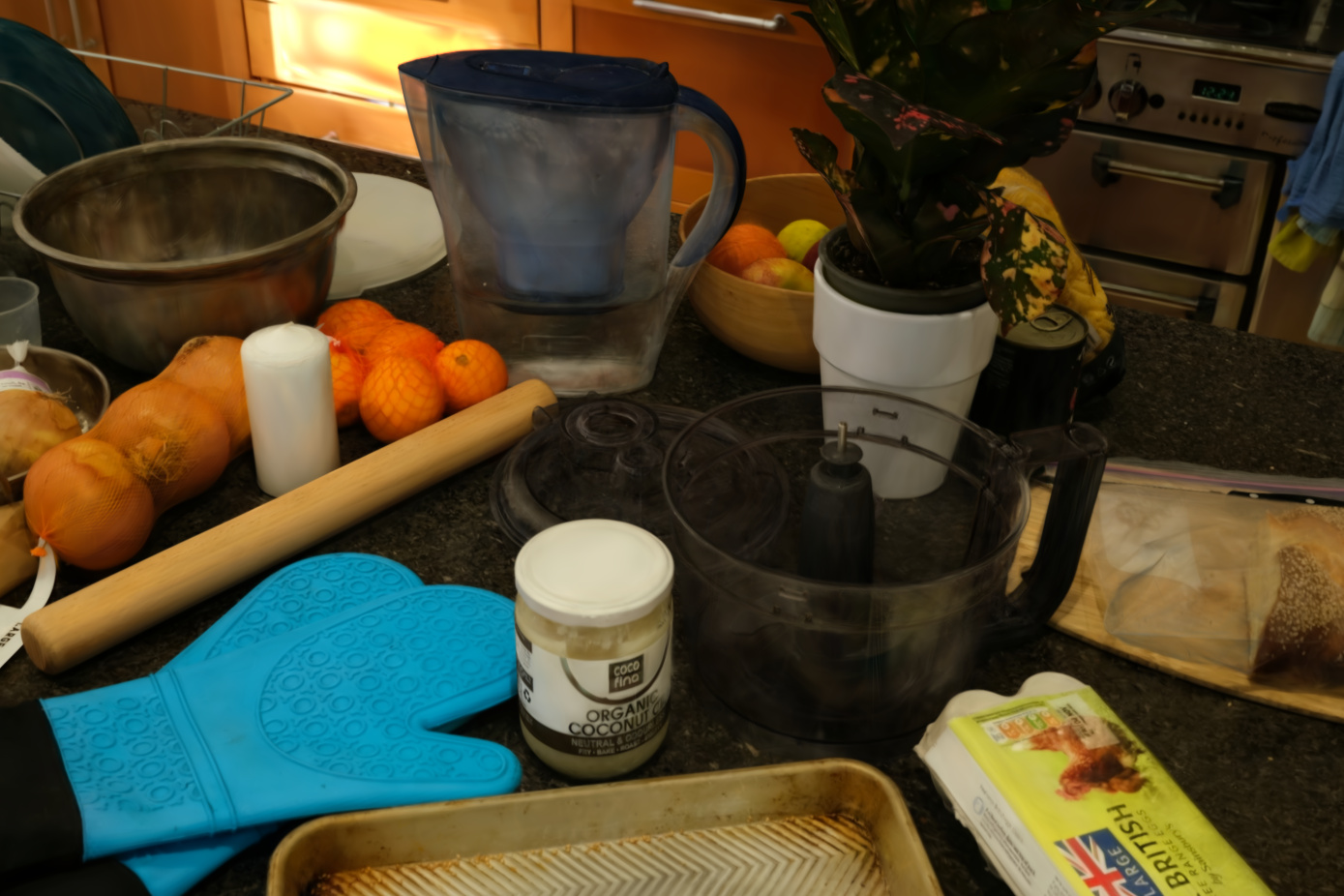}};
\node[ovpanel, anchor=north west, xshift=\dcgap] (b0r1c5) at (b0r1c4.north east) {\dcimg{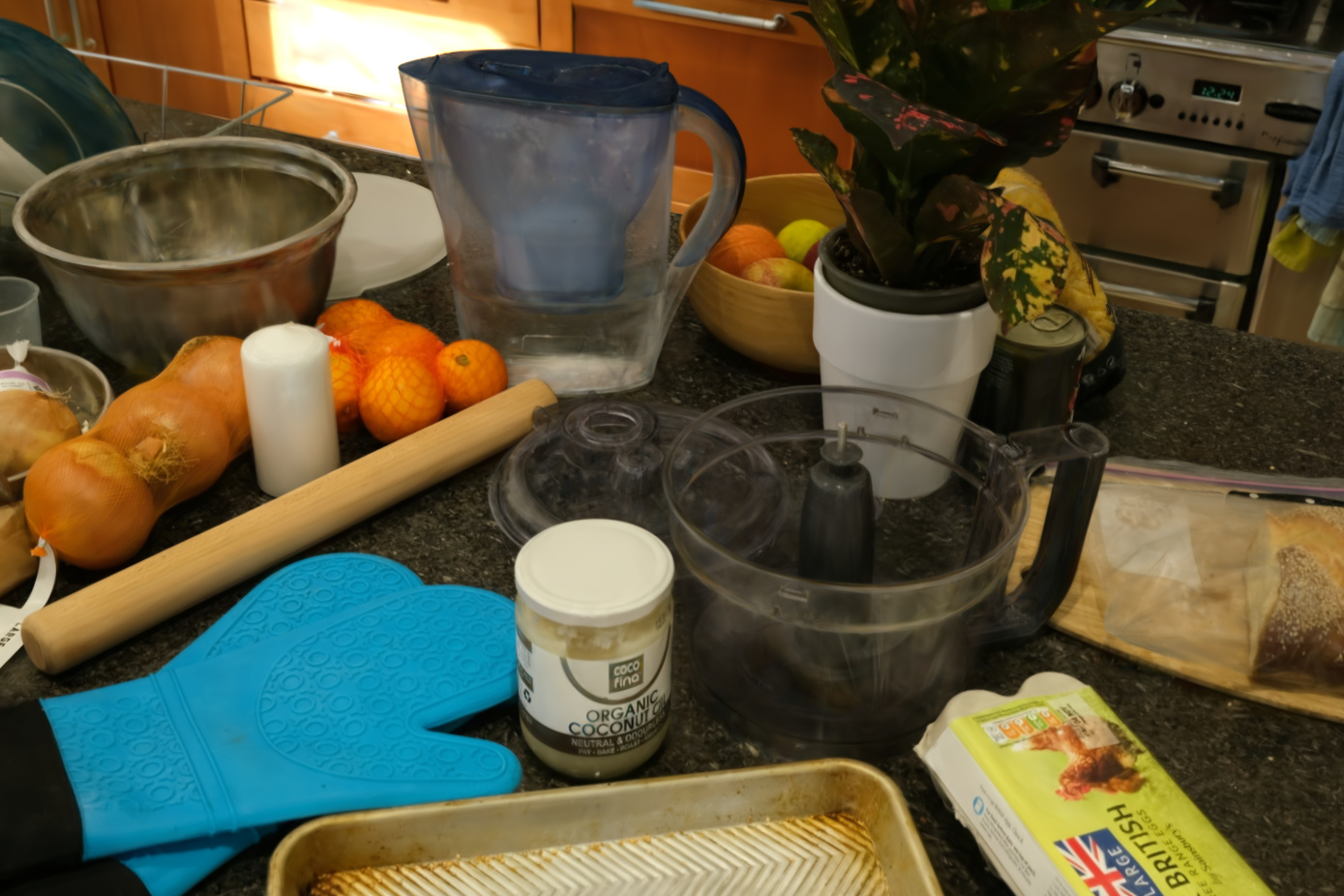}};
\node[ovpanel, anchor=north west, yshift=-\dcgap] (b0r2c0) at (b0r1c0.south west) {\dcimg{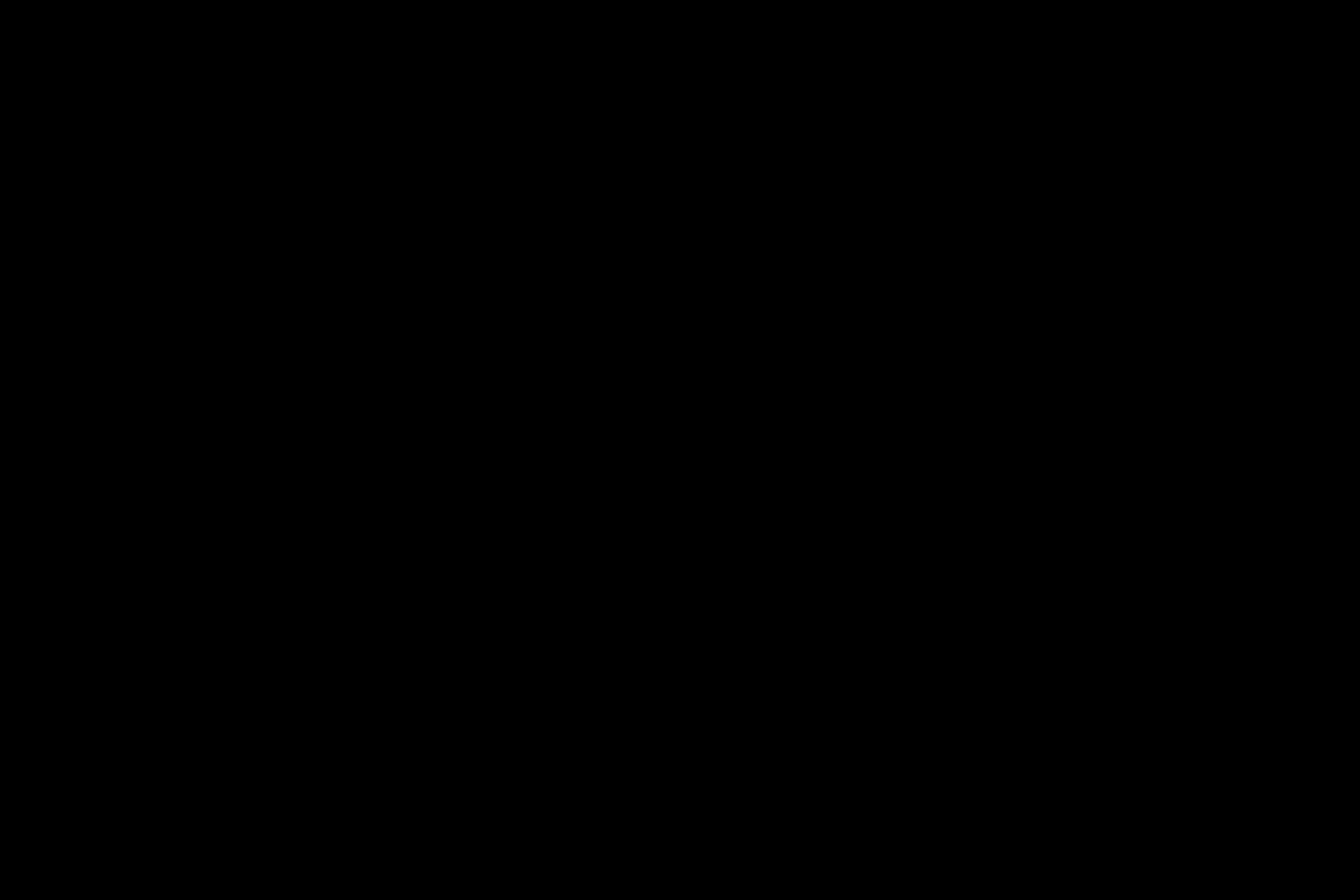}};
\node[ovpanel, anchor=north west, xshift=\dcgap] (b0r2c1) at (b0r2c0.north east) {\dcimg{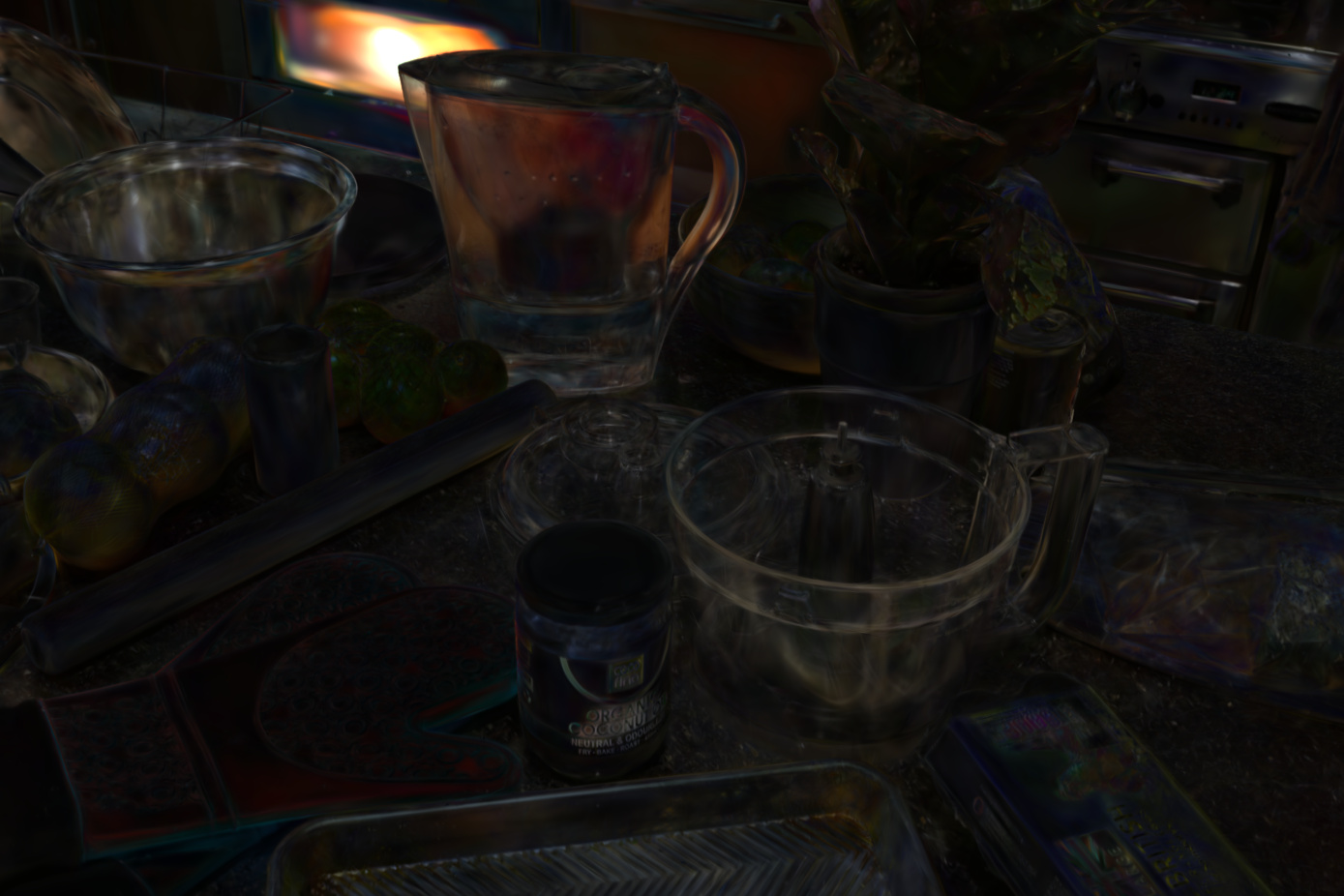}};
\node[ovpanel, anchor=north west, xshift=\dcgap] (b0r2c2) at (b0r2c1.north east) {\dcimg{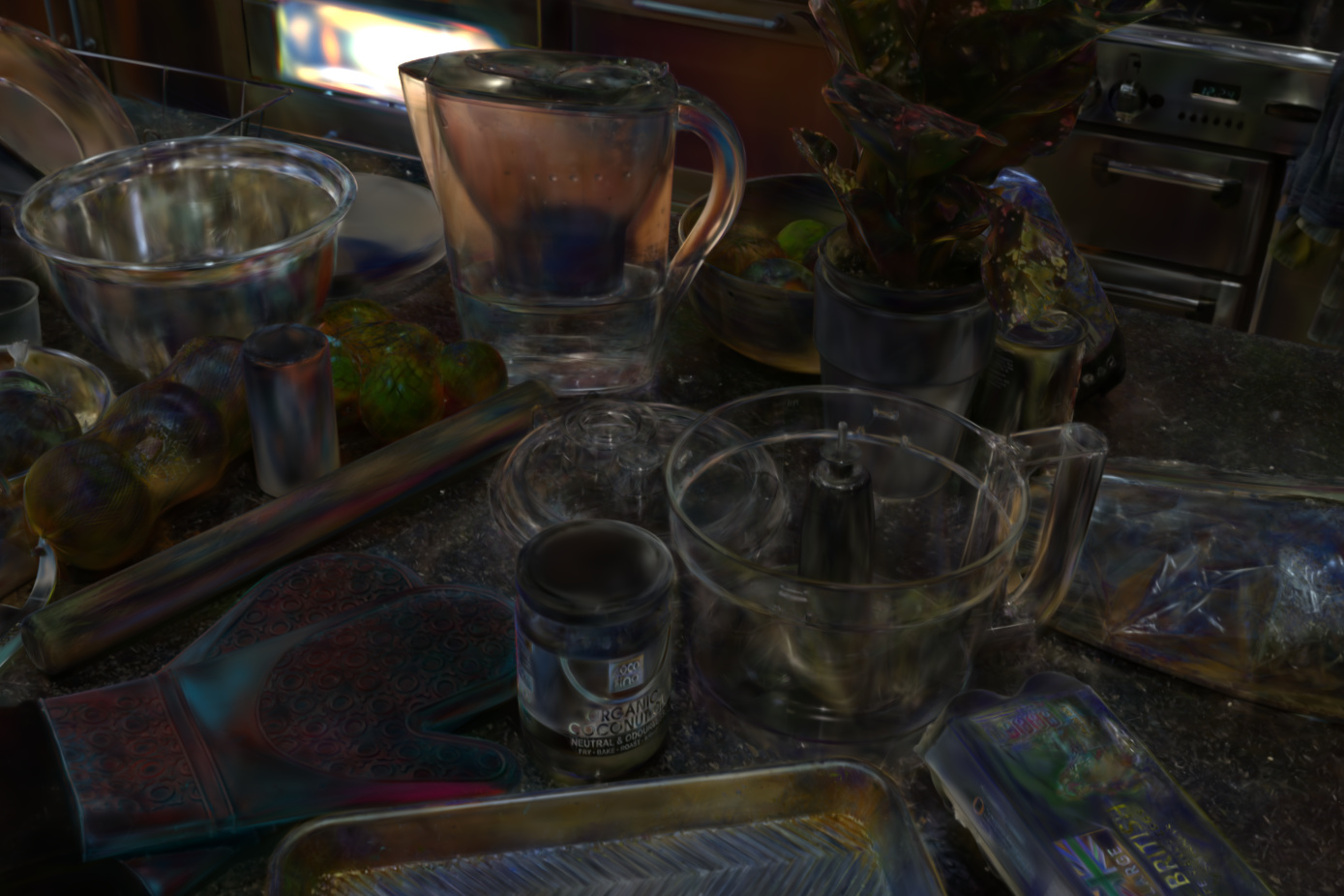}};
\node[ovpanel, anchor=north west, xshift=\dcgap] (b0r2c3) at (b0r2c2.north east) {\dcimg{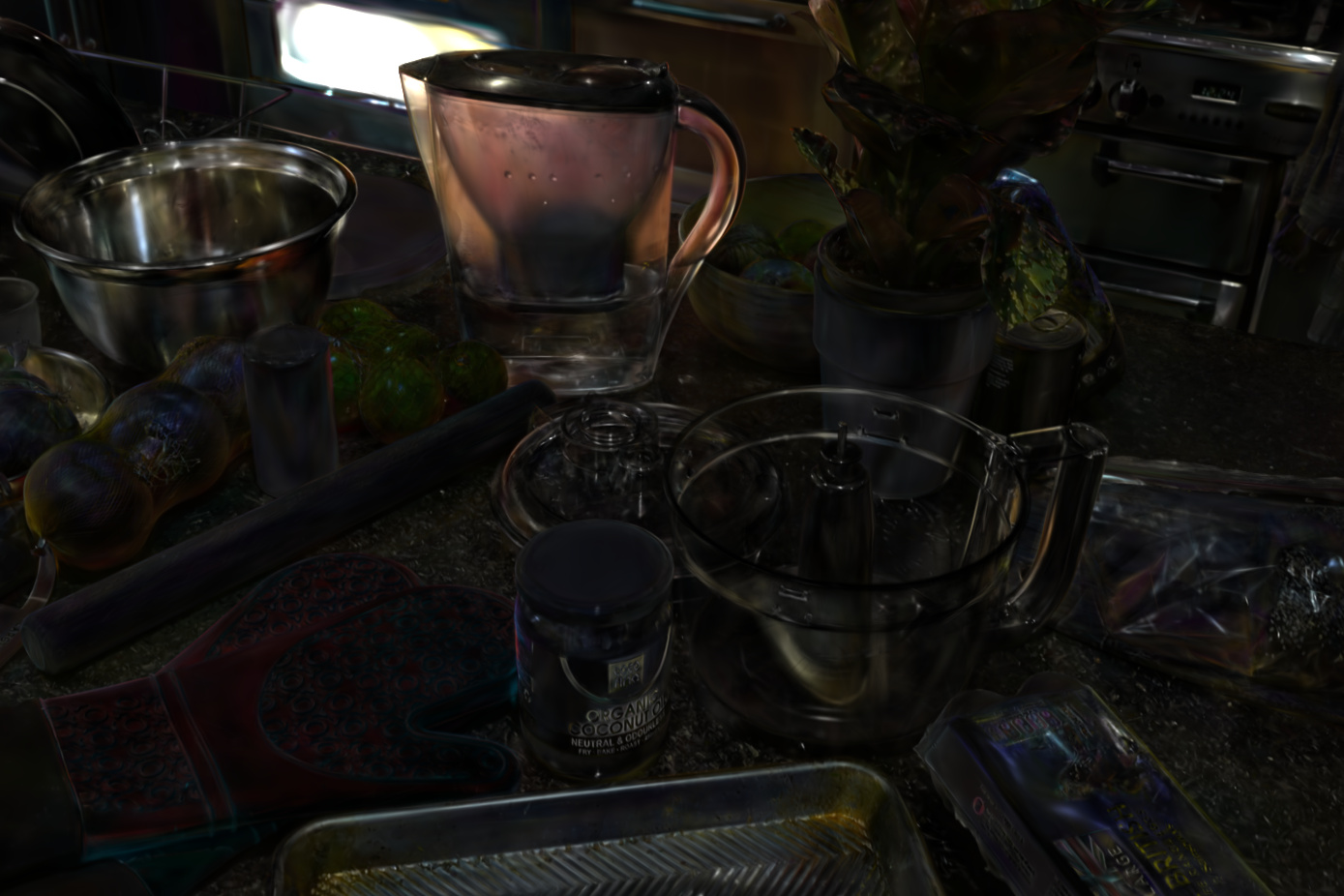}};
\node[ovpanel, anchor=north west, xshift=\dcgap] (b0r2c4) at (b0r2c3.north east) {\dcimg{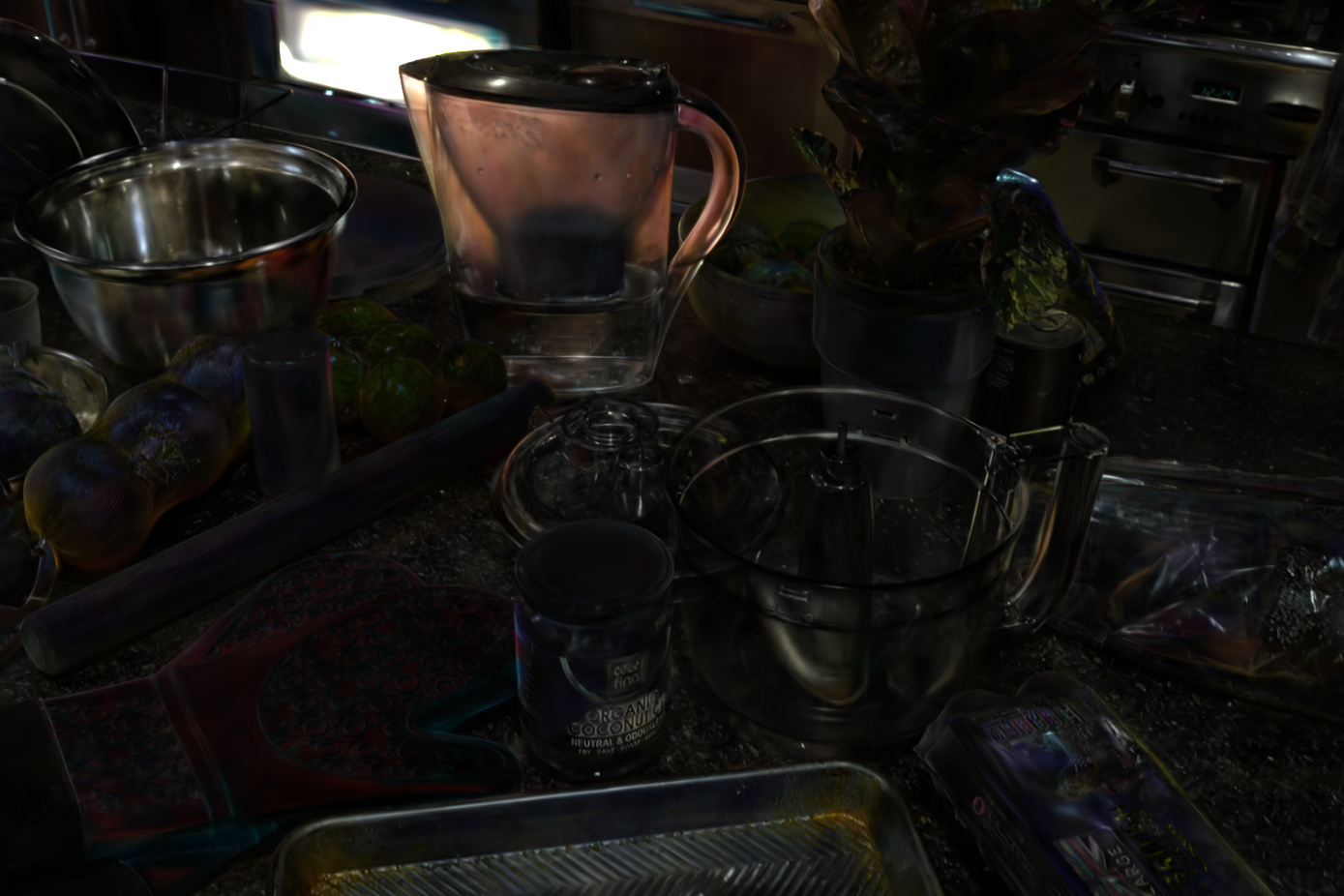}};
\node[ovpanel, anchor=north west, xshift=\dcgap] (b0r2c5) at (b0r2c4.north east) {\dcimg{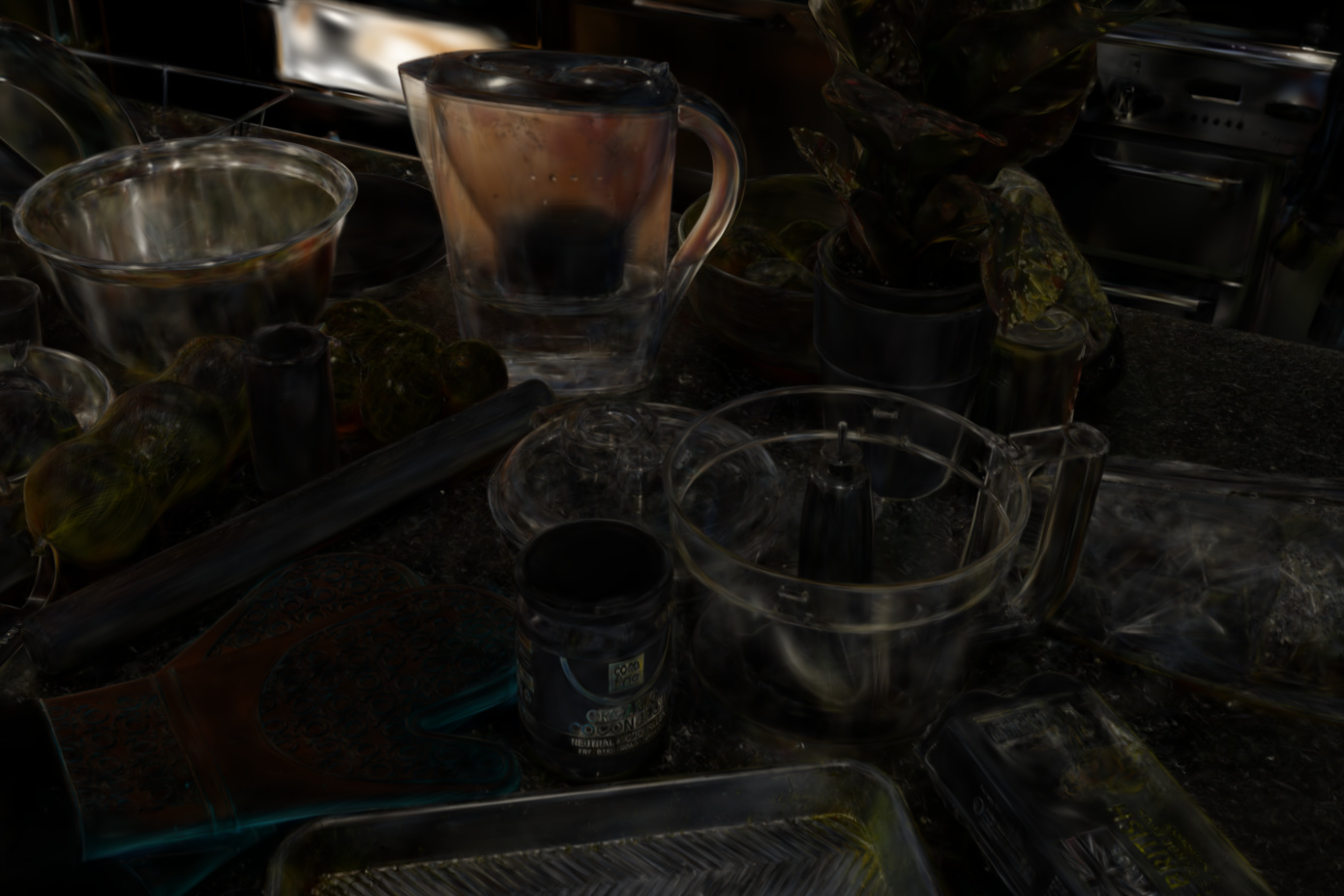}};
\node[ovpanel, anchor=north west, xshift=\dcgap] (ref0) at (b0r0c5.north east) {\dcref{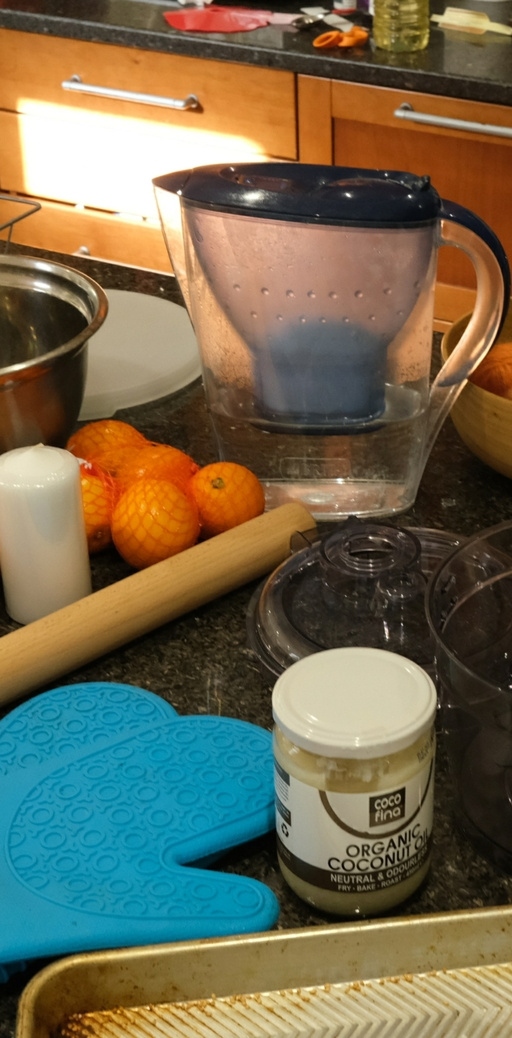}};
\node[ovpanel, anchor=north west, yshift=-\dcblockgap] (b1r0c0) at (b0r2c0.south west) {\dcimg{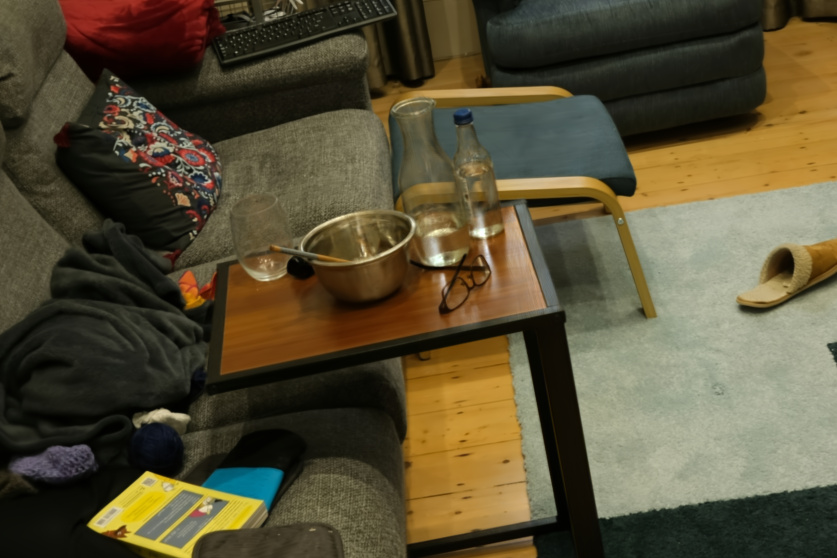}};
\node[ovpanel, anchor=north west, xshift=\dcgap] (b1r0c1) at (b1r0c0.north east) {\dcimg{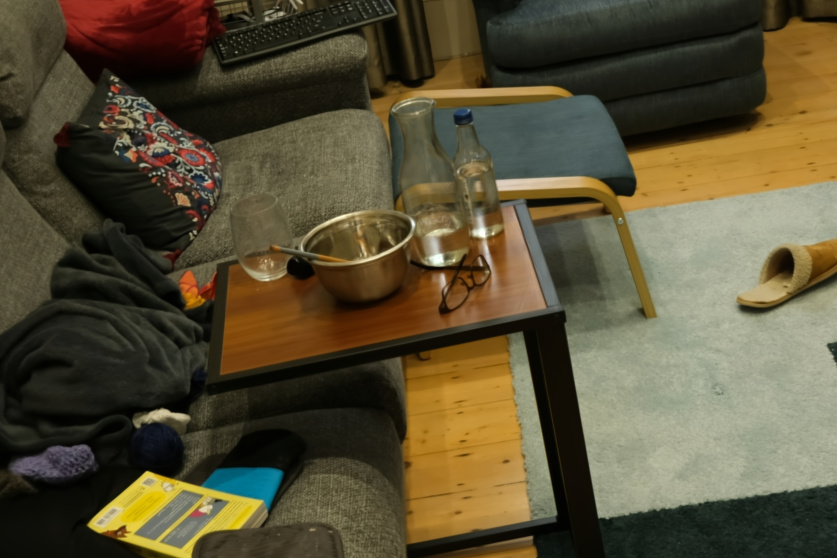}};
\node[ovpanel, anchor=north west, xshift=\dcgap] (b1r0c2) at (b1r0c1.north east) {\dcimg{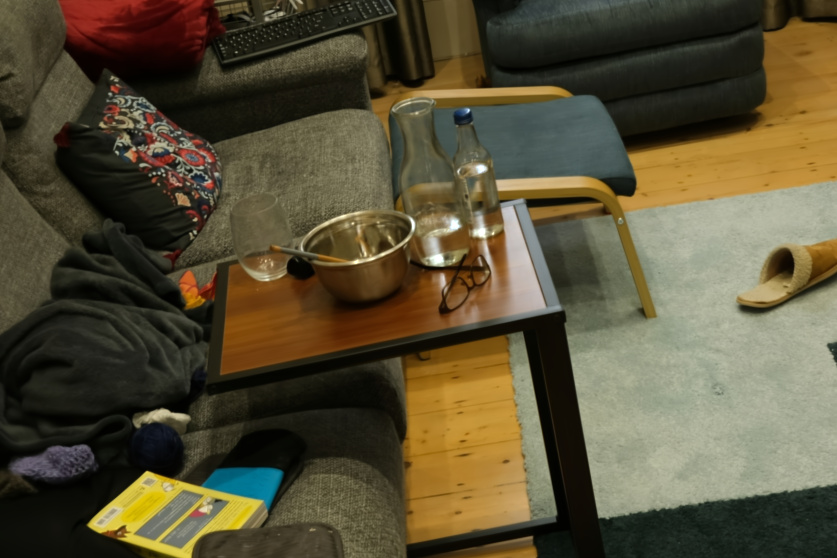}};
\node[ovpanel, anchor=north west, xshift=\dcgap] (b1r0c3) at (b1r0c2.north east) {\dcimg{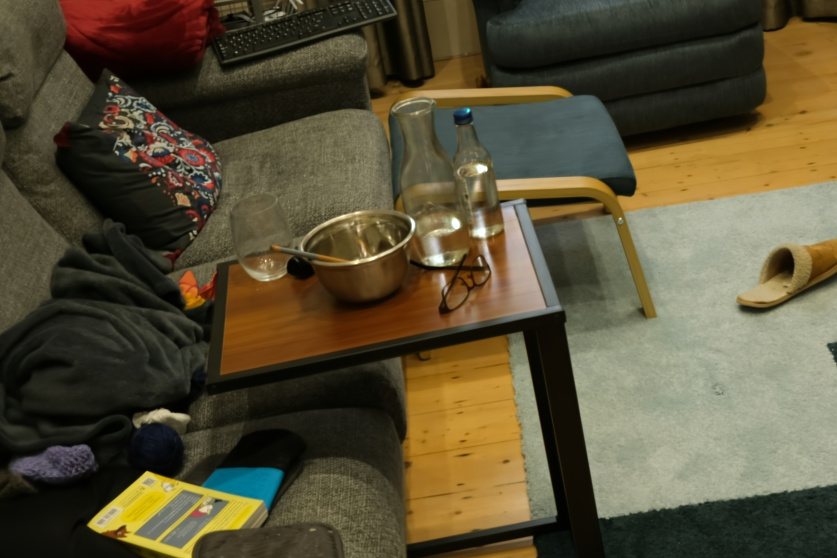}};
\node[ovpanel, anchor=north west, xshift=\dcgap] (b1r0c4) at (b1r0c3.north east) {\dcimg{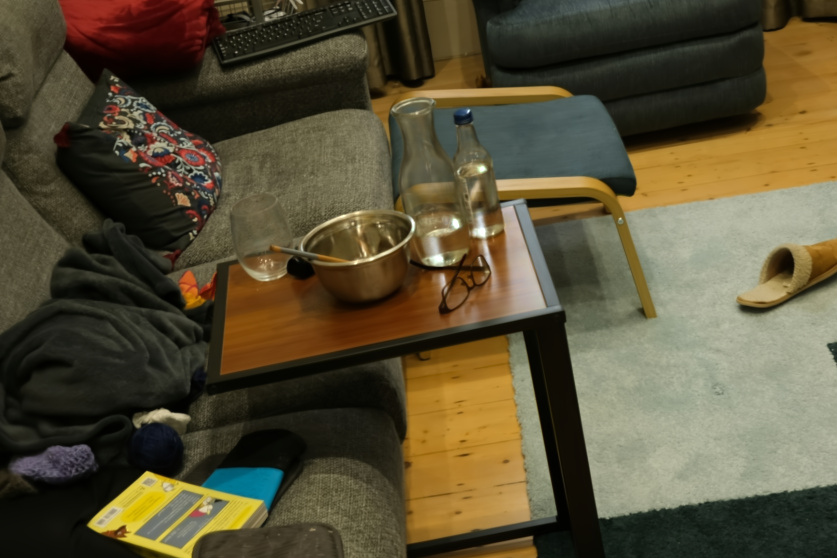}};
\node[ovpanel, anchor=north west, xshift=\dcgap] (b1r0c5) at (b1r0c4.north east) {\dcimg{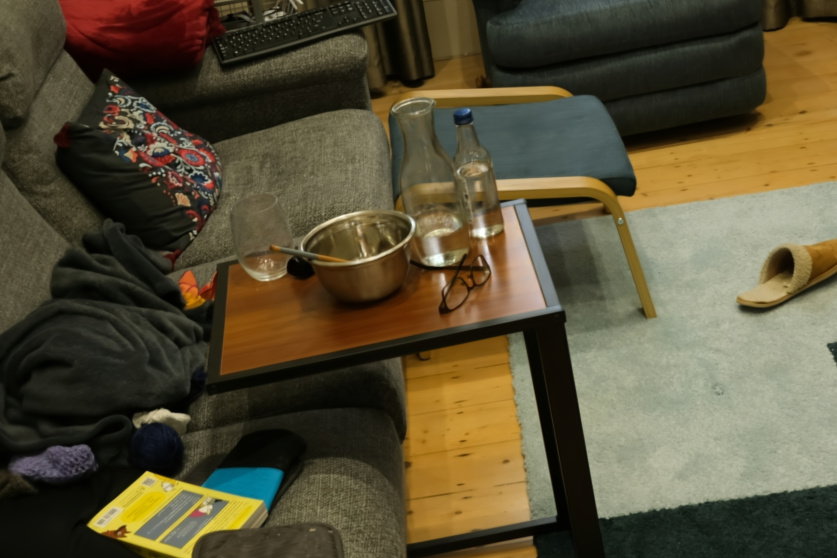}};
\node[ovpanel, anchor=north west, yshift=-\dcgap] (b1r1c0) at (b1r0c0.south west) {\dcimg{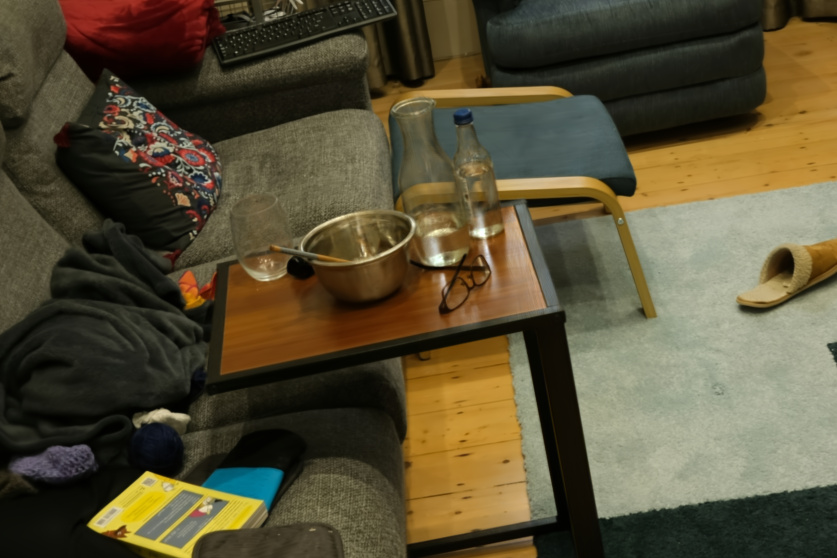}};
\node[ovpanel, anchor=north west, xshift=\dcgap] (b1r1c1) at (b1r1c0.north east) {\dcimg{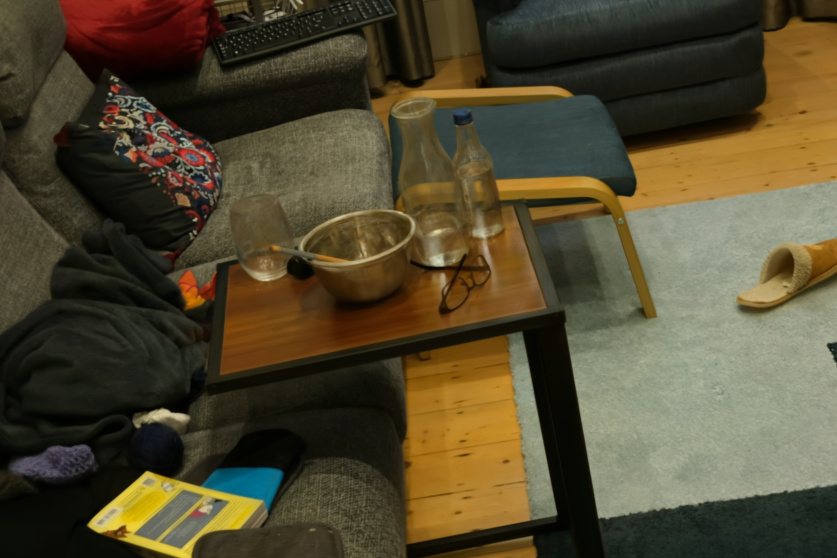}};
\node[ovpanel, anchor=north west, xshift=\dcgap] (b1r1c2) at (b1r1c1.north east) {\dcimg{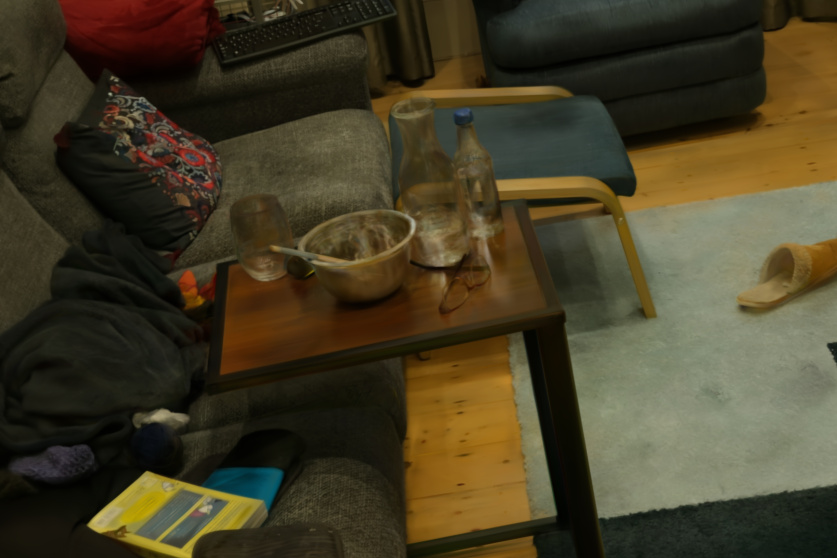}};
\node[ovpanel, anchor=north west, xshift=\dcgap] (b1r1c3) at (b1r1c2.north east) {\dcimg{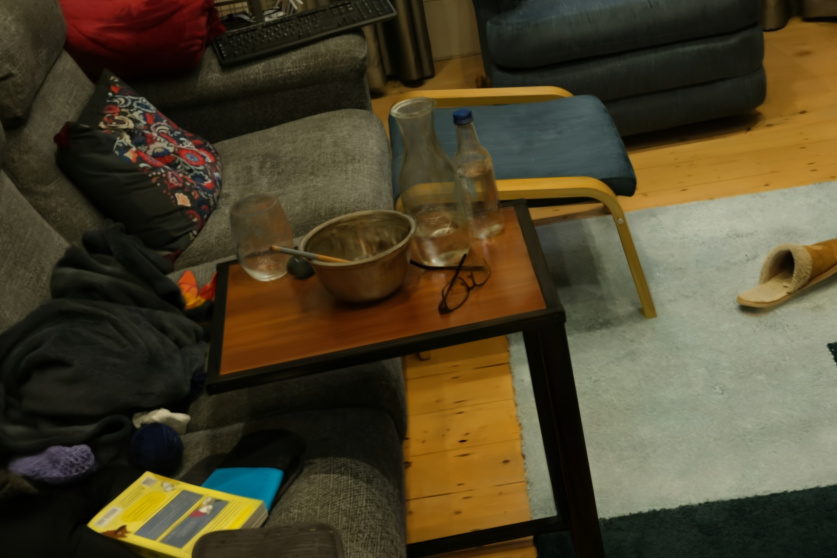}};
\node[ovpanel, anchor=north west, xshift=\dcgap] (b1r1c4) at (b1r1c3.north east) {\dcimg{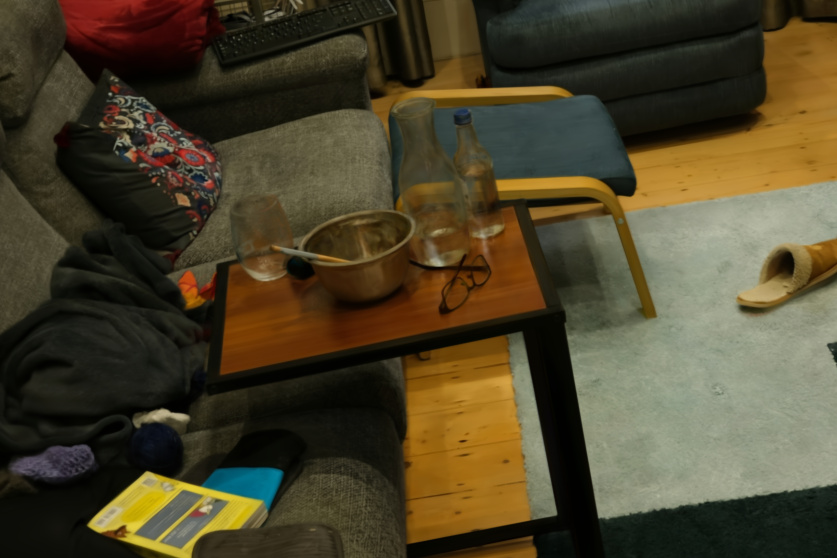}};
\node[ovpanel, anchor=north west, xshift=\dcgap] (b1r1c5) at (b1r1c4.north east) {\dcimg{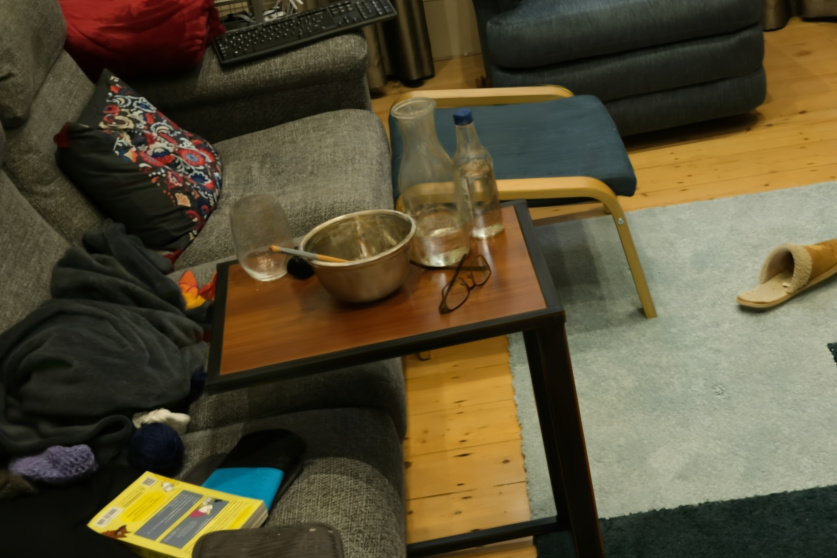}};
\node[ovpanel, anchor=north west, yshift=-\dcgap] (b1r2c0) at (b1r1c0.south west) {\dcimg{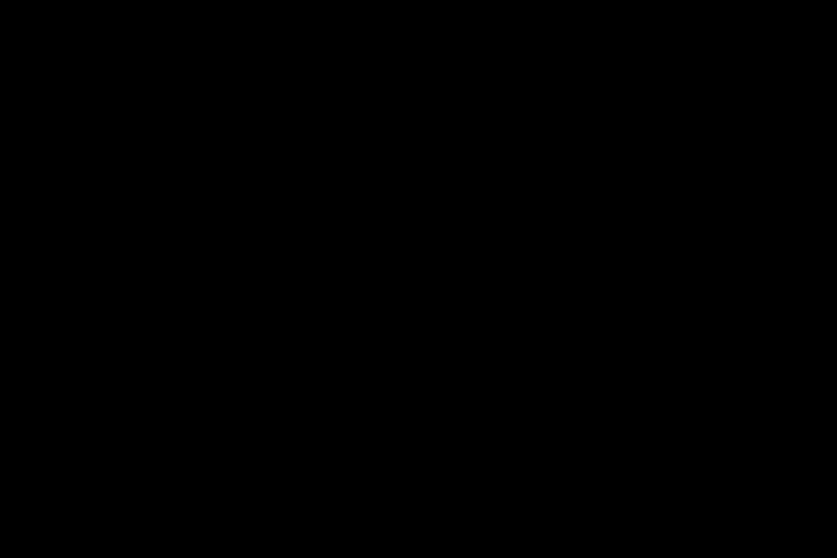}};
\node[ovpanel, anchor=north west, xshift=\dcgap] (b1r2c1) at (b1r2c0.north east) {\dcimg{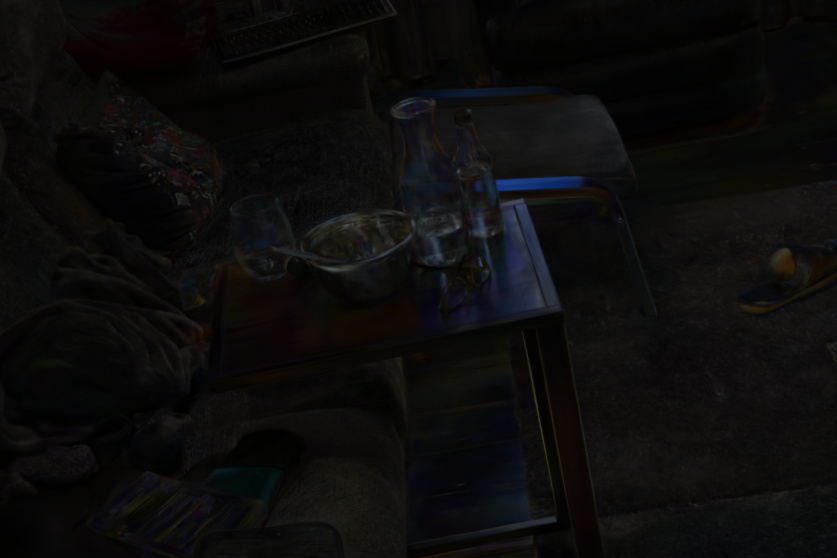}};
\node[ovpanel, anchor=north west, xshift=\dcgap] (b1r2c2) at (b1r2c1.north east) {\dcimg{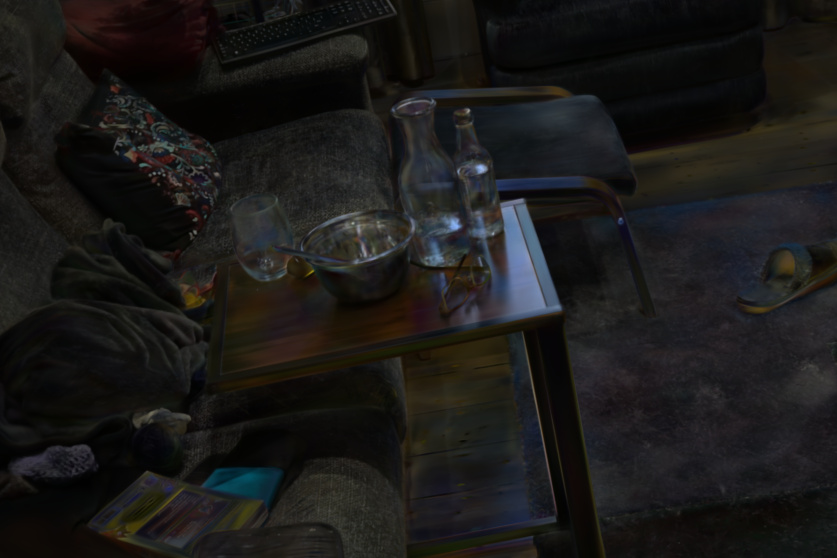}};
\node[ovpanel, anchor=north west, xshift=\dcgap] (b1r2c3) at (b1r2c2.north east) {\dcimg{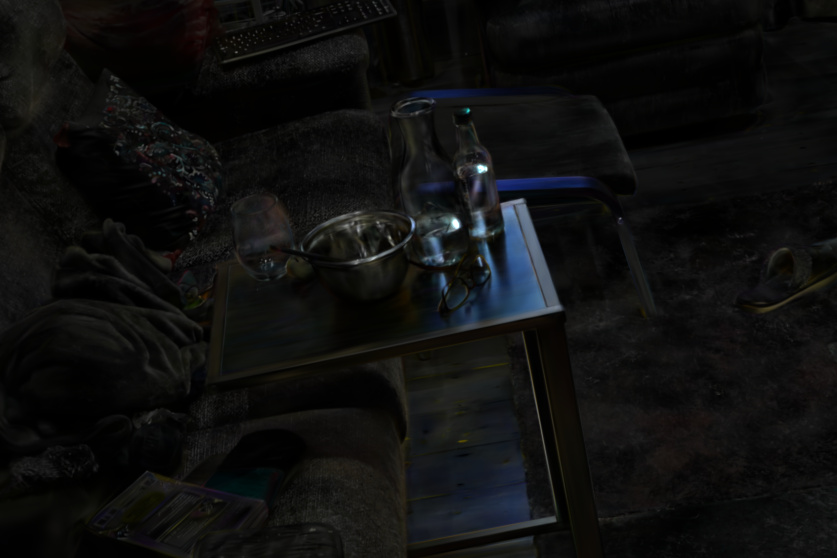}};
\node[ovpanel, anchor=north west, xshift=\dcgap] (b1r2c4) at (b1r2c3.north east) {\dcimg{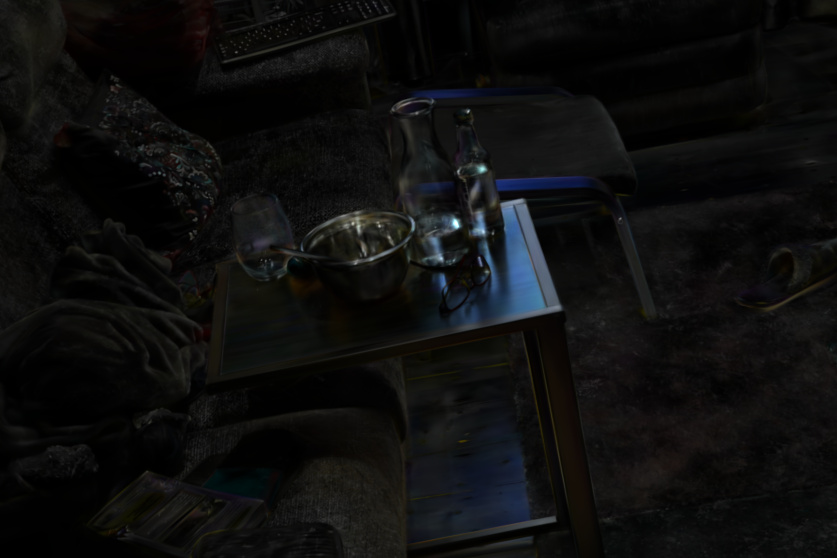}};
\node[ovpanel, anchor=north west, xshift=\dcgap] (b1r2c5) at (b1r2c4.north east) {\dcimg{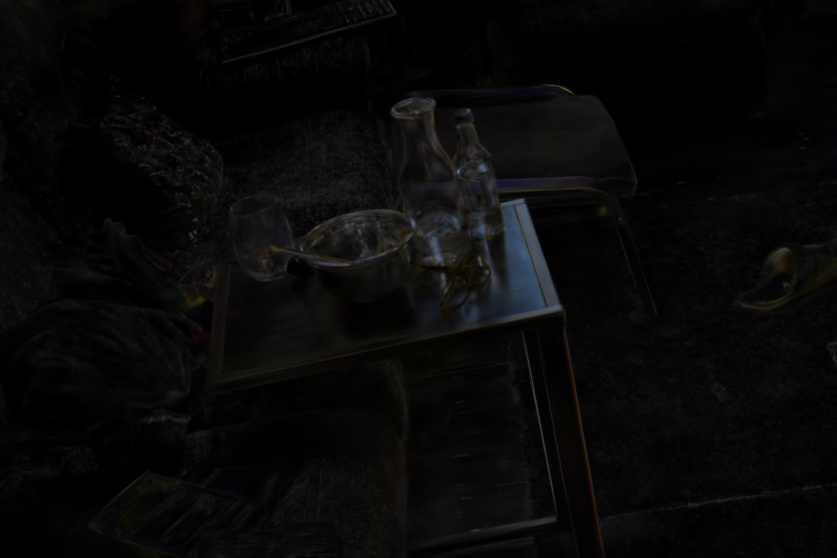}};
\node[ovpanel, anchor=north west, xshift=\dcgap] (ref1) at (b1r0c5.north east) {\dcref{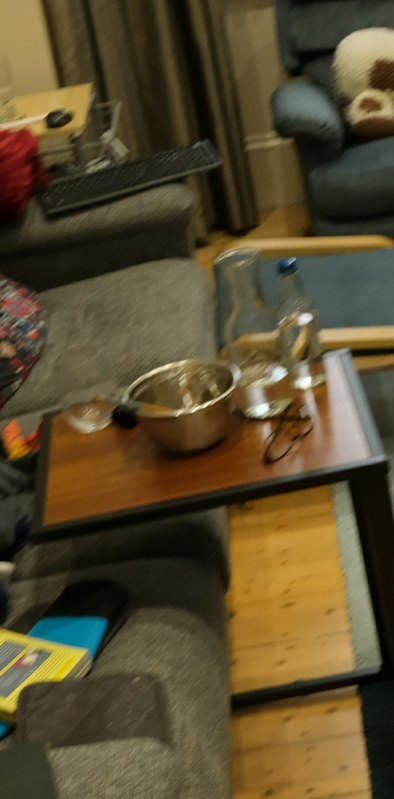}};
\node[ovpanel, anchor=north west, yshift=-\dcblockgap] (b2r0c0) at (b1r2c0.south west) {\dcimg{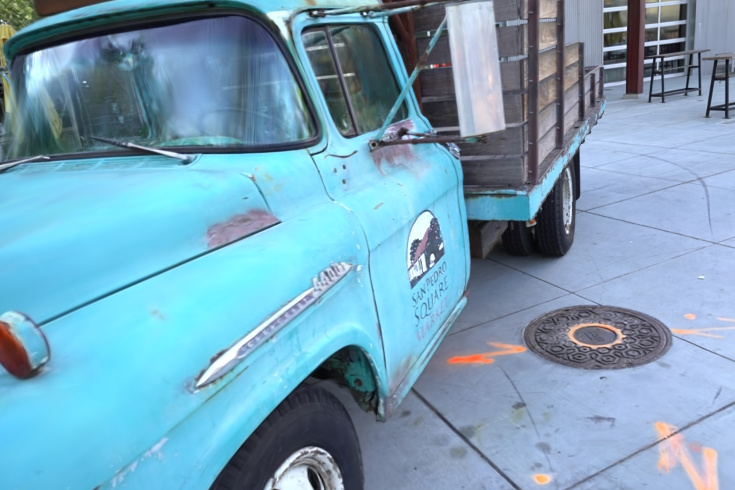}};
\node[ovpanel, anchor=north west, xshift=\dcgap] (b2r0c1) at (b2r0c0.north east) {\dcimg{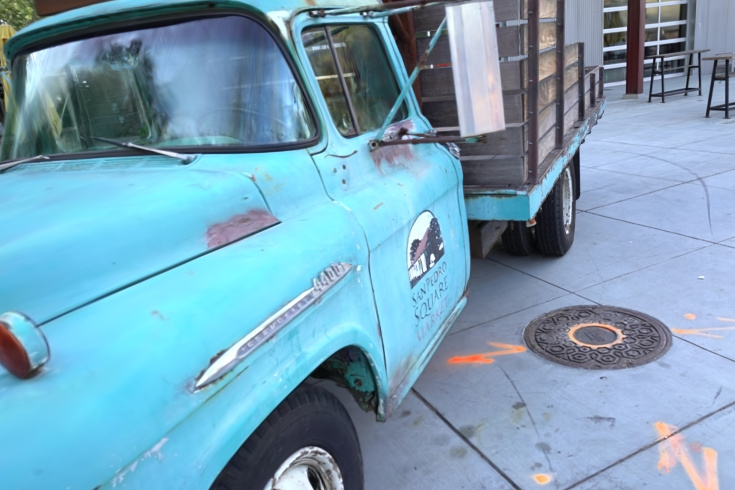}};
\node[ovpanel, anchor=north west, xshift=\dcgap] (b2r0c2) at (b2r0c1.north east) {\dcimg{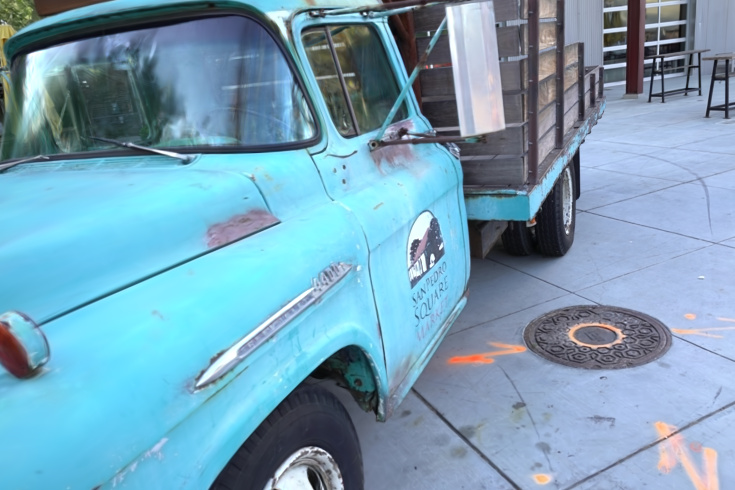}};
\node[ovpanel, anchor=north west, xshift=\dcgap] (b2r0c3) at (b2r0c2.north east) {\dcimg{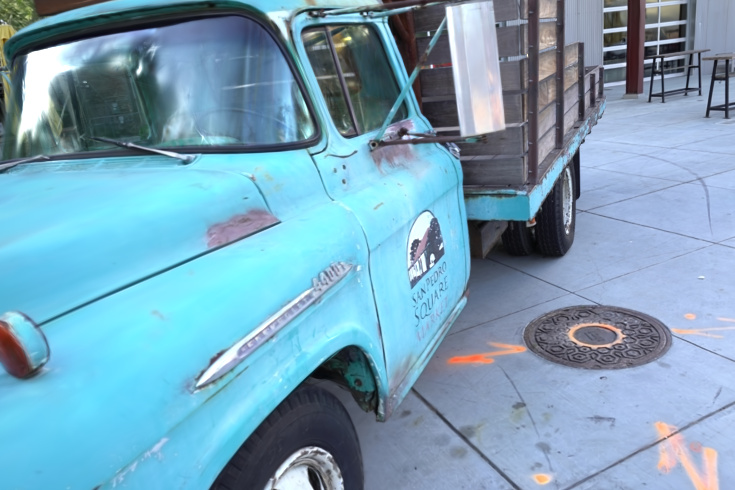}};
\node[ovpanel, anchor=north west, xshift=\dcgap] (b2r0c4) at (b2r0c3.north east) {\dcimg{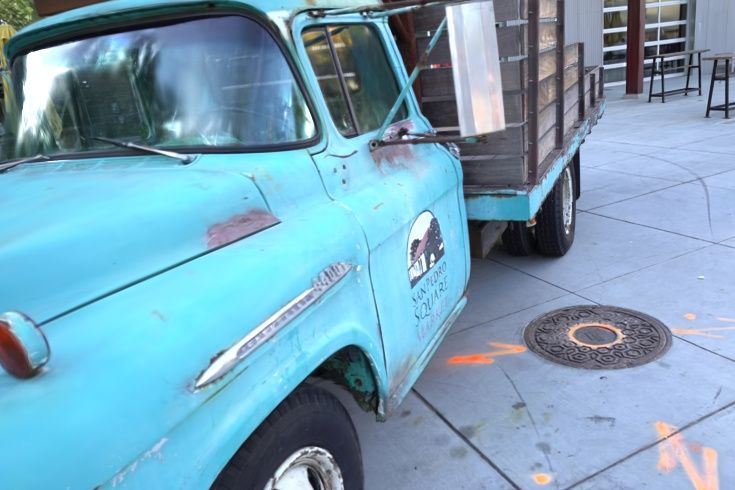}};
\node[ovpanel, anchor=north west, xshift=\dcgap] (b2r0c5) at (b2r0c4.north east) {\dcimg{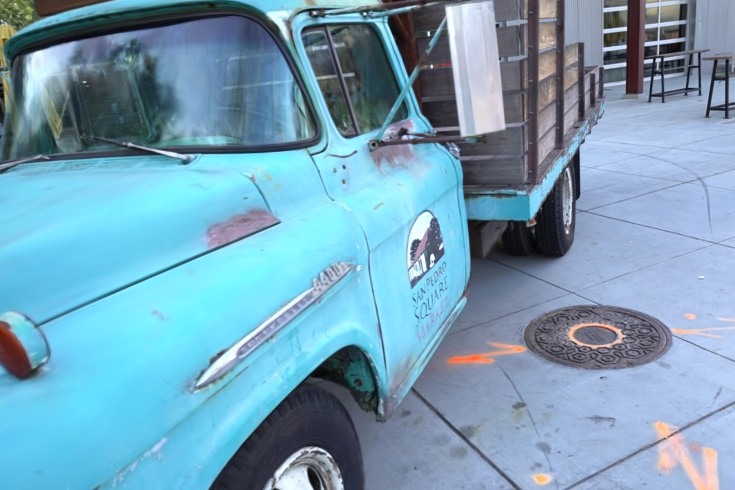}};
\node[ovpanel, anchor=north west, yshift=-\dcgap] (b2r1c0) at (b2r0c0.south west) {\dcimg{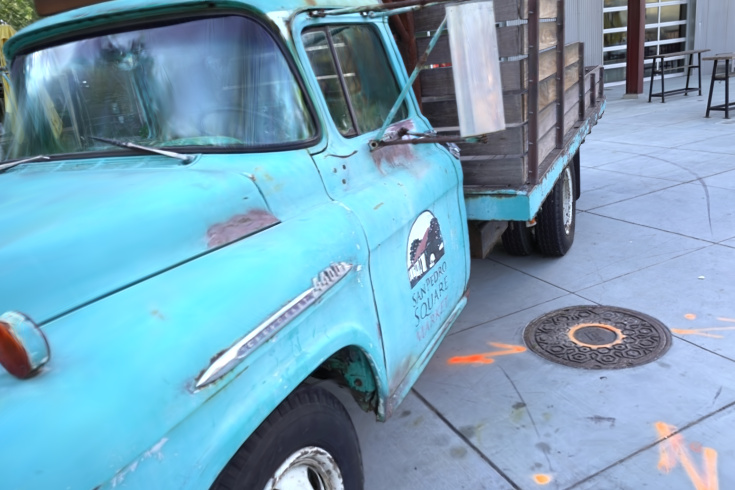}};
\node[ovpanel, anchor=north west, xshift=\dcgap] (b2r1c1) at (b2r1c0.north east) {\dcimg{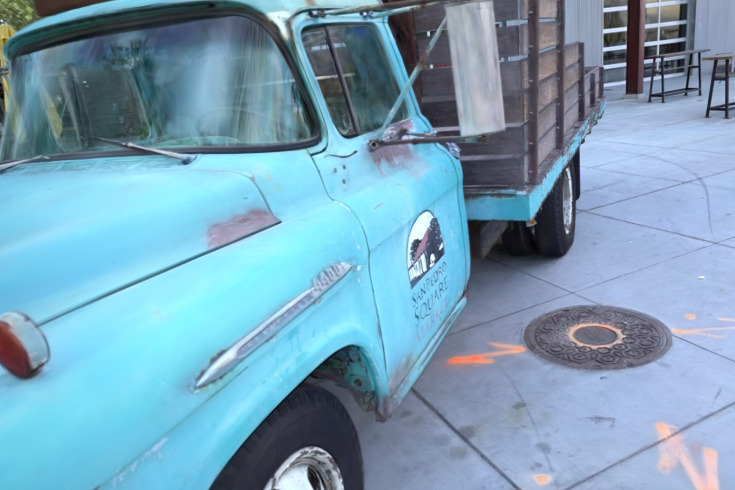}};
\node[ovpanel, anchor=north west, xshift=\dcgap] (b2r1c2) at (b2r1c1.north east) {\dcimg{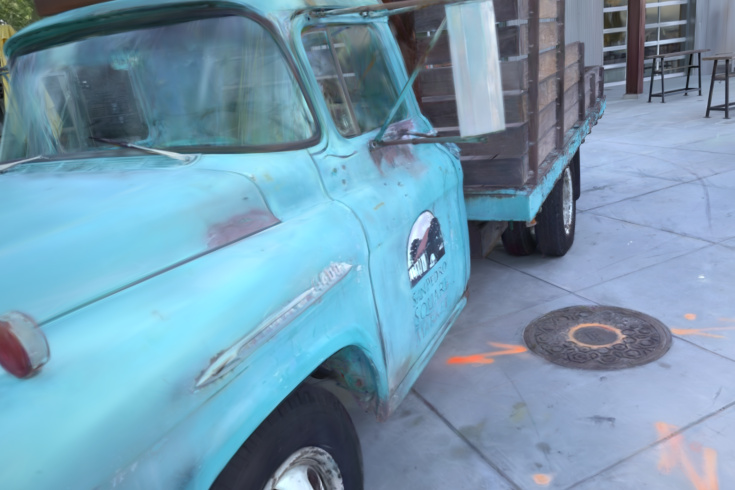}};
\node[ovpanel, anchor=north west, xshift=\dcgap] (b2r1c3) at (b2r1c2.north east) {\dcimg{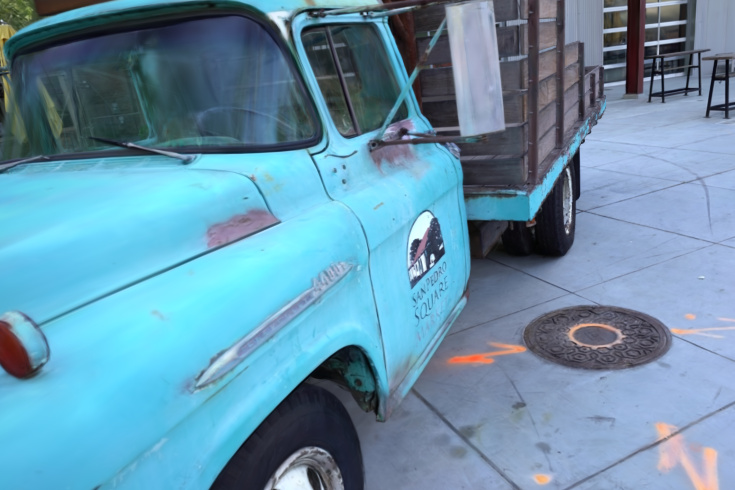}};
\node[ovpanel, anchor=north west, xshift=\dcgap] (b2r1c4) at (b2r1c3.north east) {\dcimg{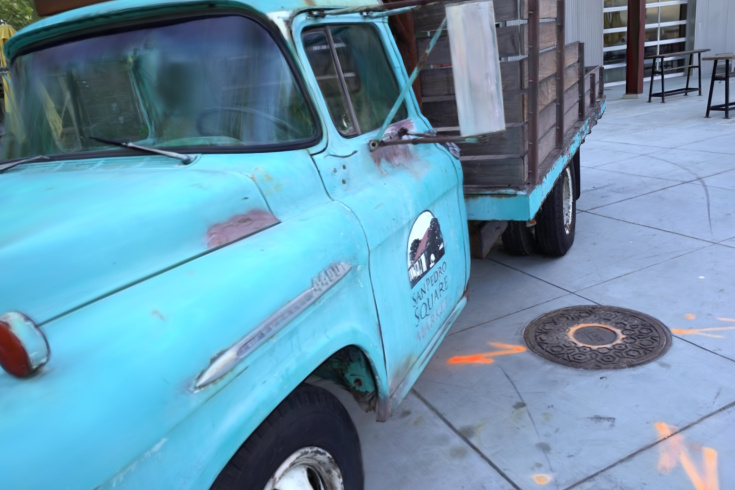}};
\node[ovpanel, anchor=north west, xshift=\dcgap] (b2r1c5) at (b2r1c4.north east) {\dcimg{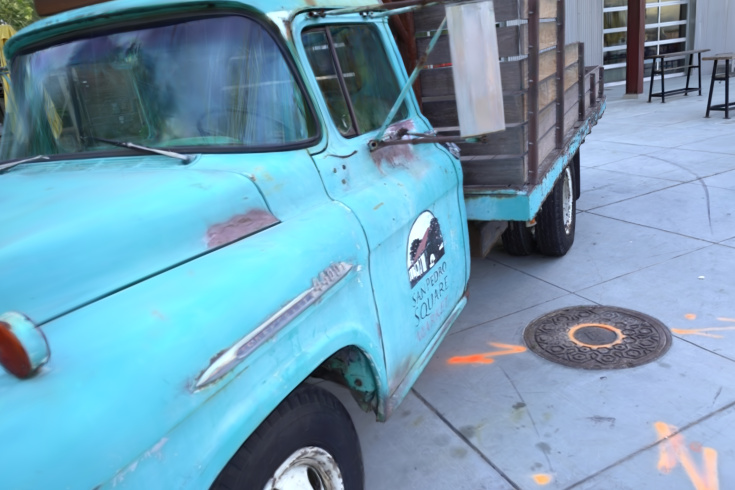}};
\node[ovpanel, anchor=north west, yshift=-\dcgap] (b2r2c0) at (b2r1c0.south west) {\dcimg{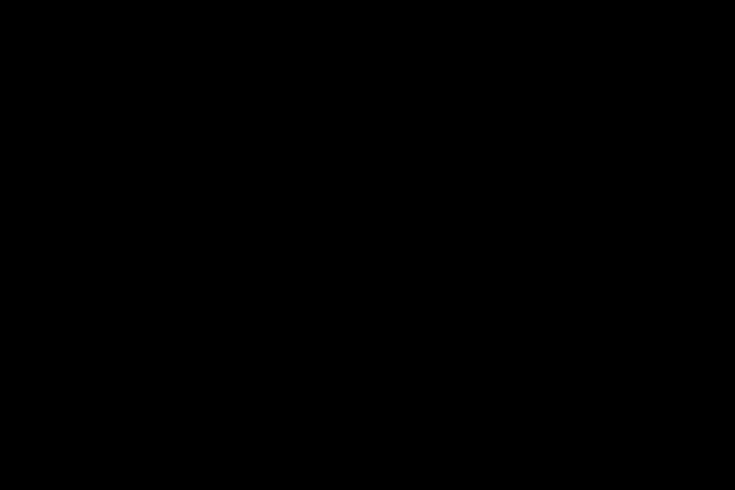}};
\node[ovpanel, anchor=north west, xshift=\dcgap] (b2r2c1) at (b2r2c0.north east) {\dcimg{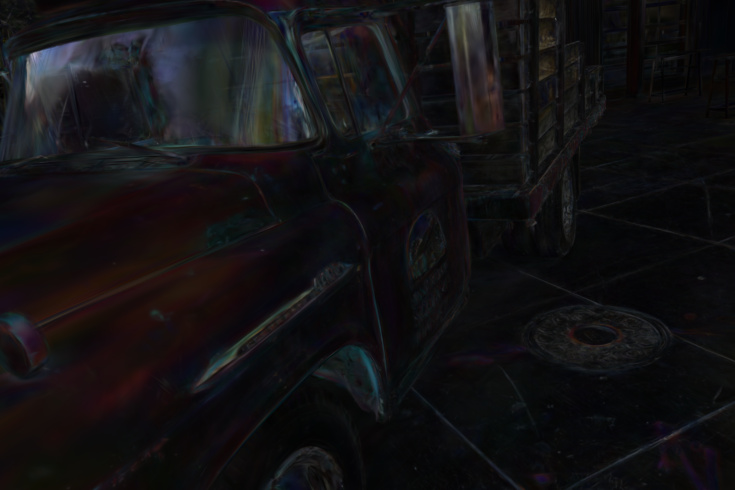}};
\node[ovpanel, anchor=north west, xshift=\dcgap] (b2r2c2) at (b2r2c1.north east) {\dcimg{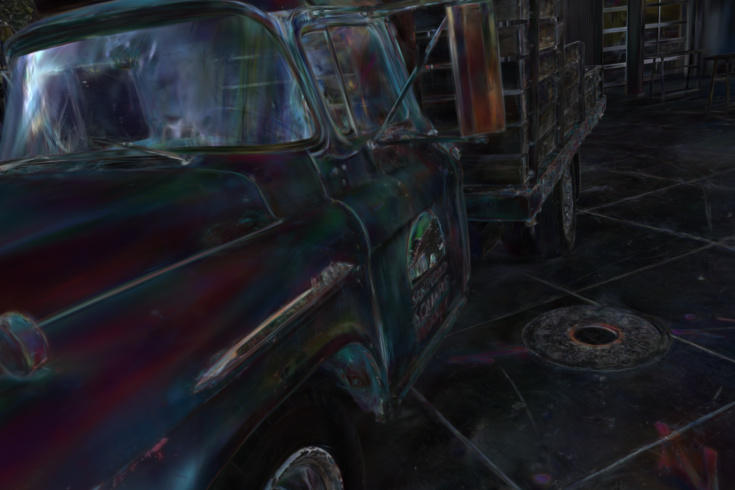}};
\node[ovpanel, anchor=north west, xshift=\dcgap] (b2r2c3) at (b2r2c2.north east) {\dcimg{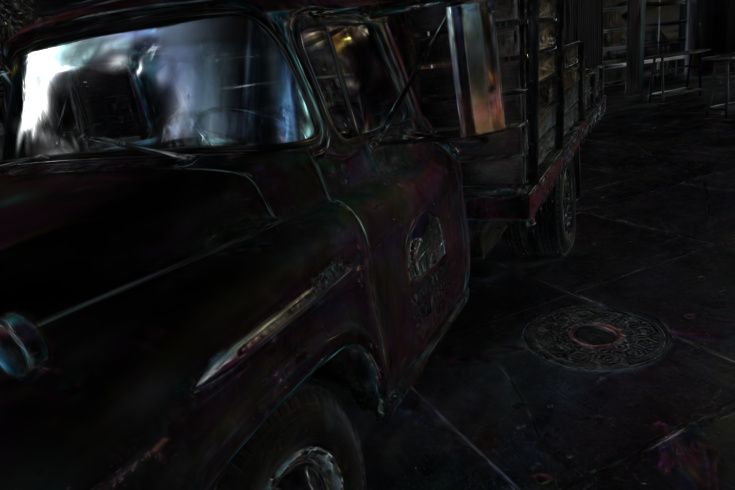}};
\node[ovpanel, anchor=north west, xshift=\dcgap] (b2r2c4) at (b2r2c3.north east) {\dcimg{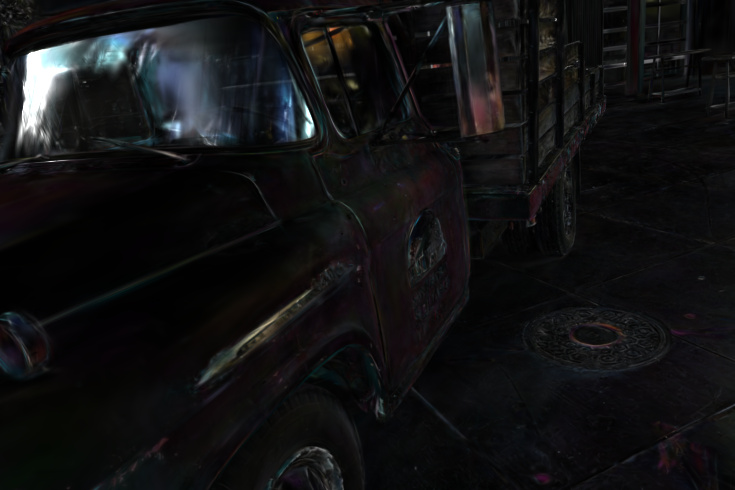}};
\node[ovpanel, anchor=north west, xshift=\dcgap] (b2r2c5) at (b2r2c4.north east) {\dcimg{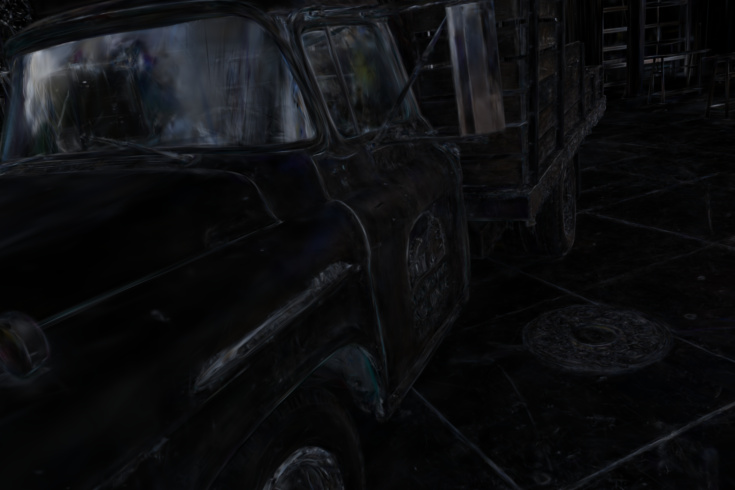}};
\node[ovpanel, anchor=north west, xshift=\dcgap] (ref2) at (b2r0c5.north east) {\dcref{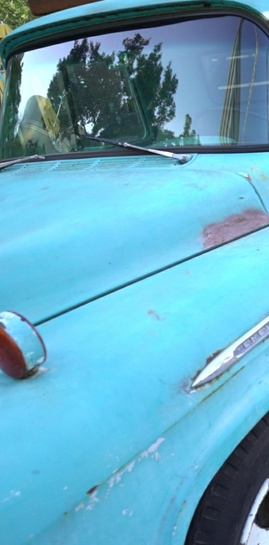}};
\node[ovpanel, anchor=north west, yshift=-\dcblockgap] (b3r0c0) at (b2r2c0.south west) {\dcimg{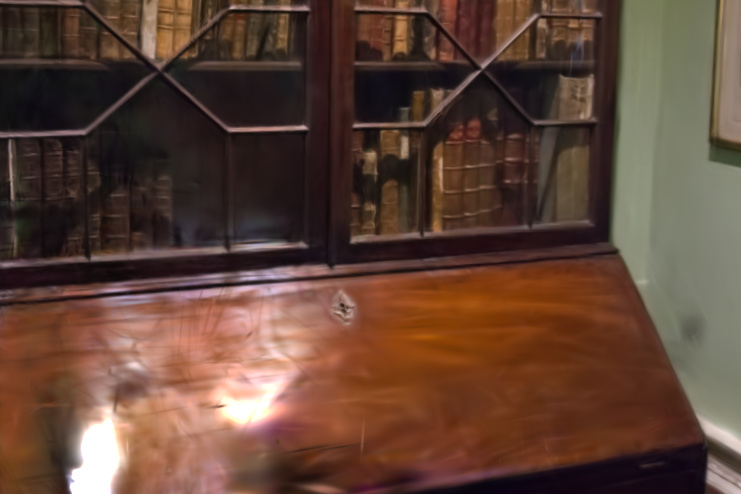}};
\node[ovpanel, anchor=north west, xshift=\dcgap] (b3r0c1) at (b3r0c0.north east) {\dcimg{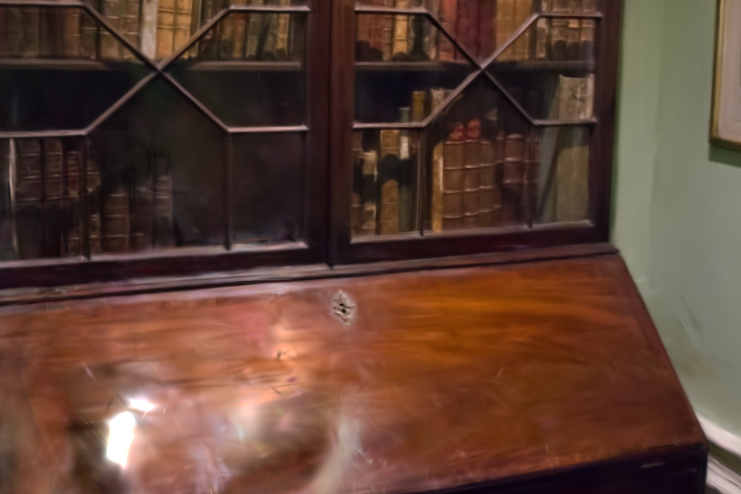}};
\node[ovpanel, anchor=north west, xshift=\dcgap] (b3r0c2) at (b3r0c1.north east) {\dcimg{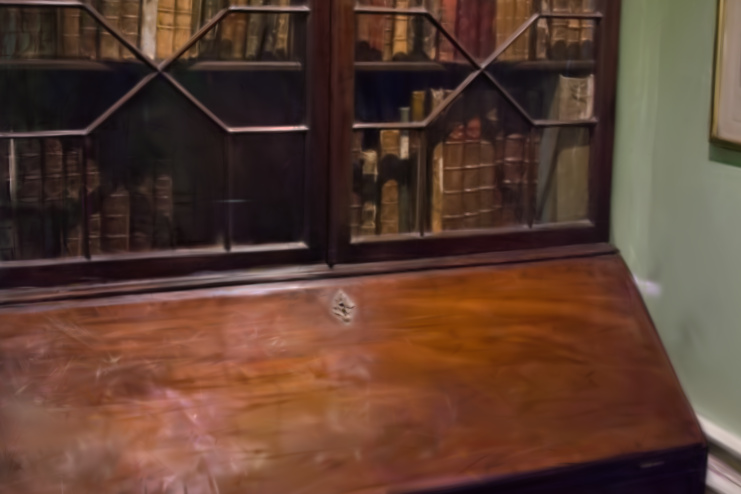}};
\node[ovpanel, anchor=north west, xshift=\dcgap] (b3r0c3) at (b3r0c2.north east) {\dcimg{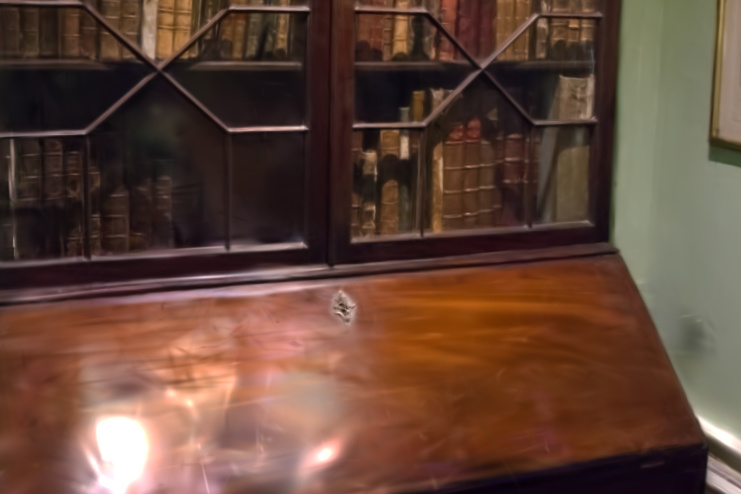}};
\node[ovpanel, anchor=north west, xshift=\dcgap] (b3r0c4) at (b3r0c3.north east) {\dcimg{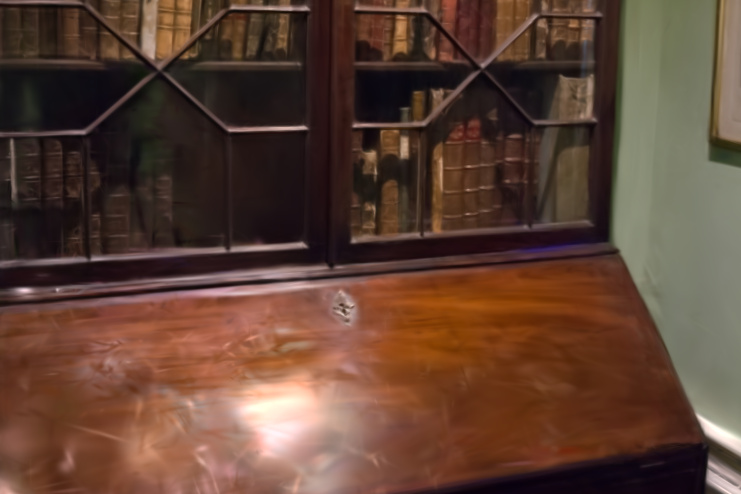}};
\node[ovpanel, anchor=north west, xshift=\dcgap] (b3r0c5) at (b3r0c4.north east) {\dcimg{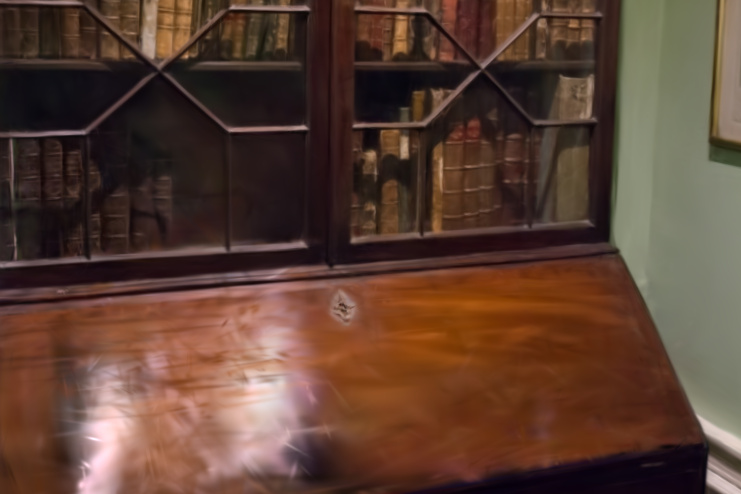}};
\node[ovpanel, anchor=north west, yshift=-\dcgap] (b3r1c0) at (b3r0c0.south west) {\dcimg{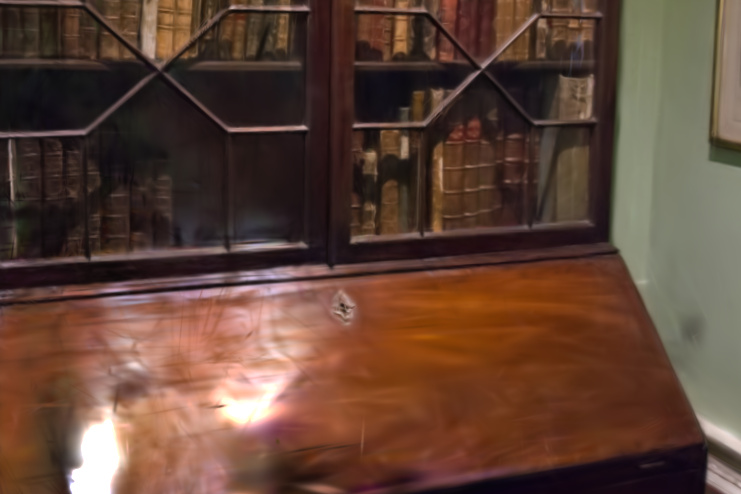}};
\node[ovpanel, anchor=north west, xshift=\dcgap] (b3r1c1) at (b3r1c0.north east) {\dcimg{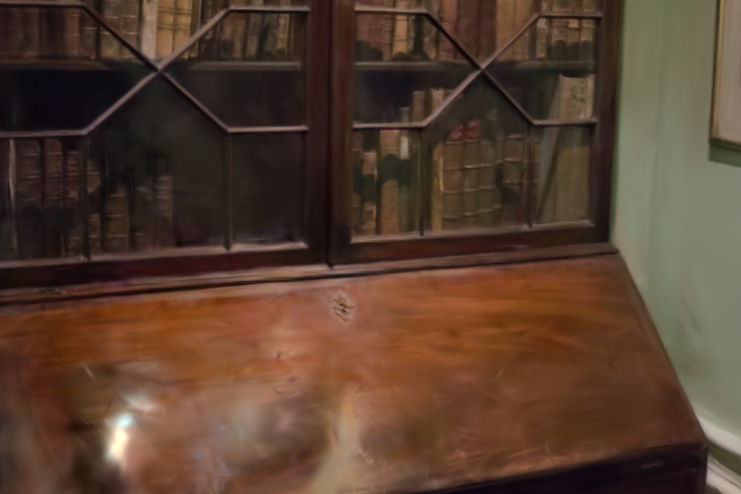}};
\node[ovpanel, anchor=north west, xshift=\dcgap] (b3r1c2) at (b3r1c1.north east) {\dcimg{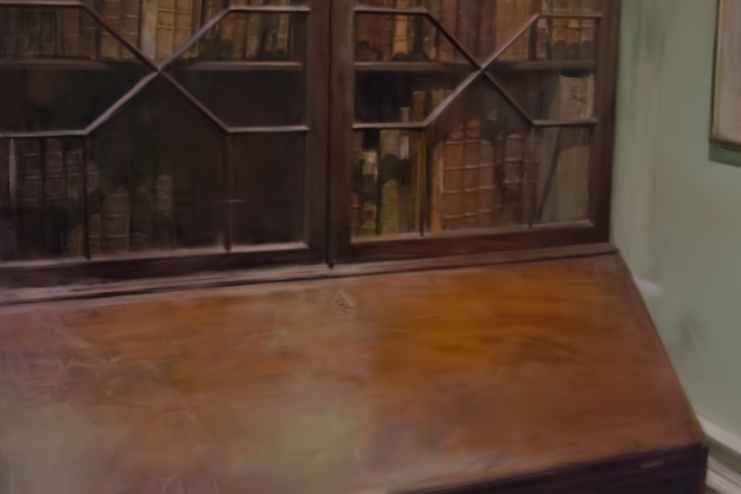}};
\node[ovpanel, anchor=north west, xshift=\dcgap] (b3r1c3) at (b3r1c2.north east) {\dcimg{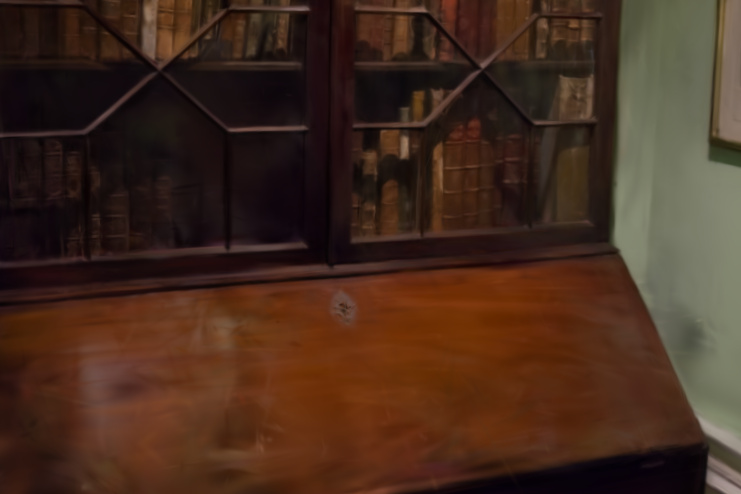}};
\node[ovpanel, anchor=north west, xshift=\dcgap] (b3r1c4) at (b3r1c3.north east) {\dcimg{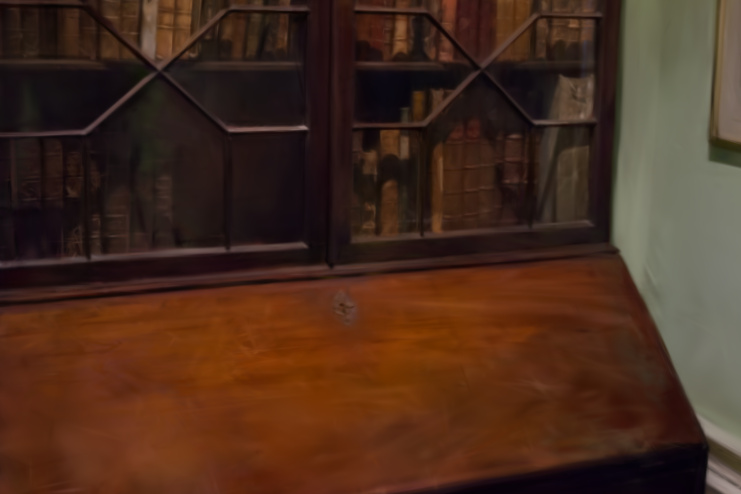}};
\node[ovpanel, anchor=north west, xshift=\dcgap] (b3r1c5) at (b3r1c4.north east) {\dcimg{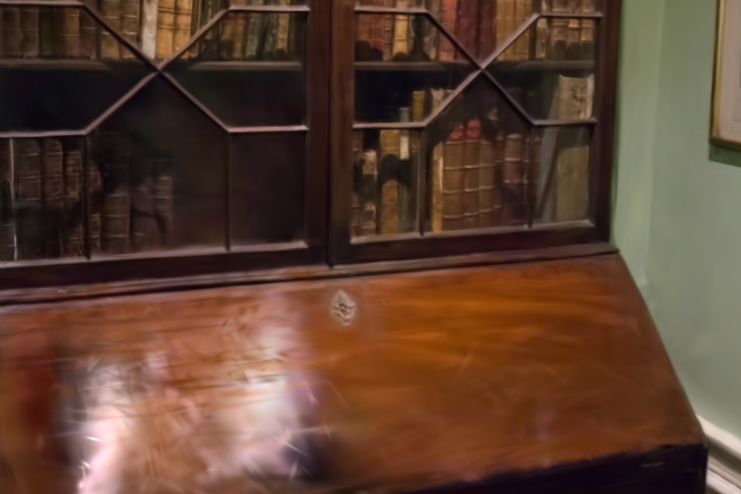}};
\node[ovpanel, anchor=north west, yshift=-\dcgap] (b3r2c0) at (b3r1c0.south west) {\dcimg{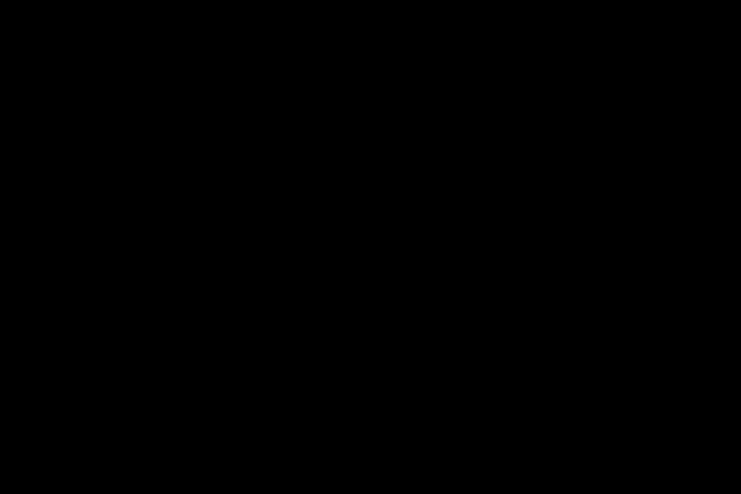}};
\node[ovpanel, anchor=north west, xshift=\dcgap] (b3r2c1) at (b3r2c0.north east) {\dcimg{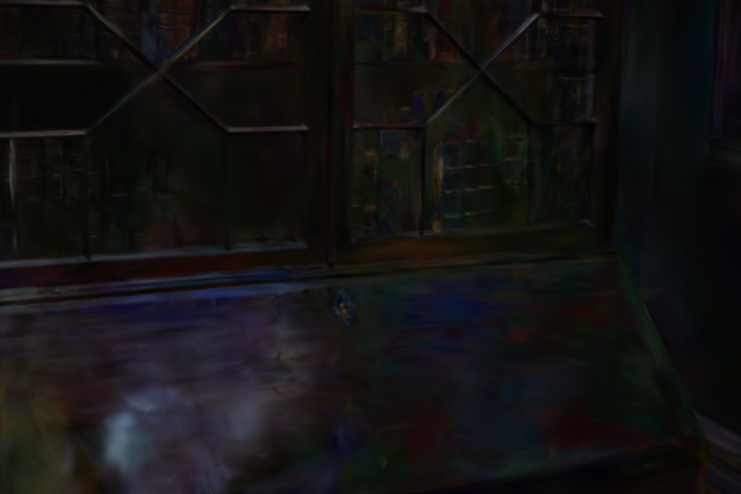}};
\node[ovpanel, anchor=north west, xshift=\dcgap] (b3r2c2) at (b3r2c1.north east) {\dcimg{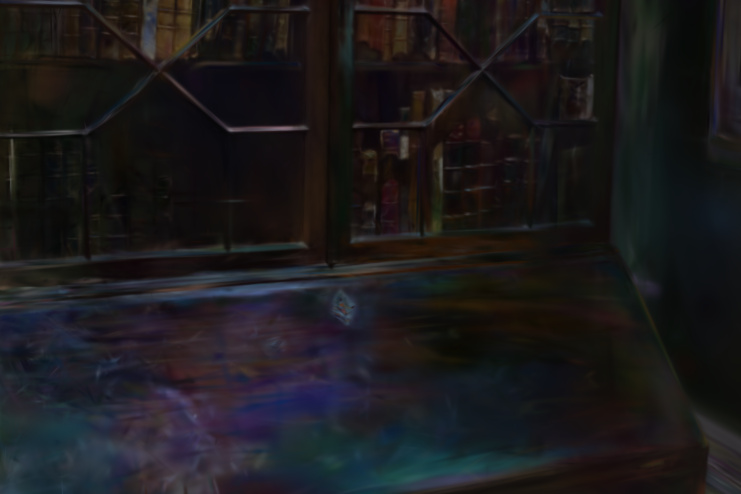}};
\node[ovpanel, anchor=north west, xshift=\dcgap] (b3r2c3) at (b3r2c2.north east) {\dcimg{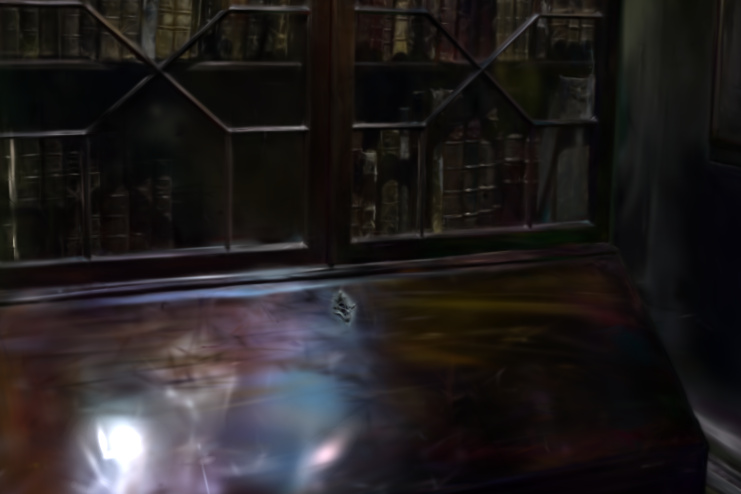}};
\node[ovpanel, anchor=north west, xshift=\dcgap] (b3r2c4) at (b3r2c3.north east) {\dcimg{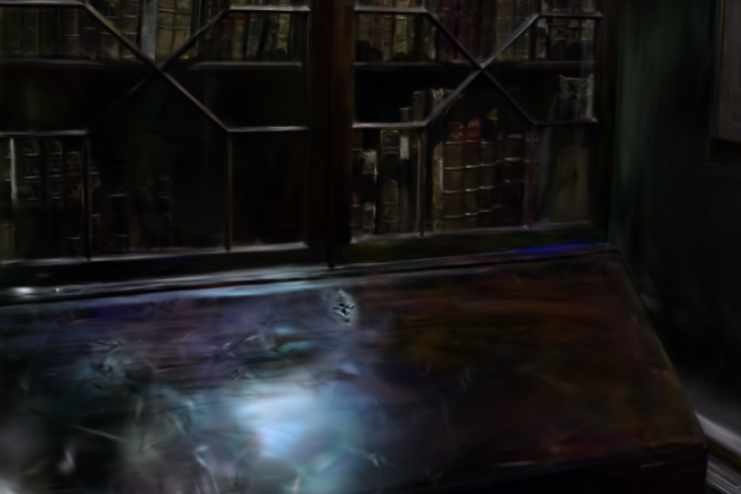}};
\node[ovpanel, anchor=north west, xshift=\dcgap] (b3r2c5) at (b3r2c4.north east) {\dcimg{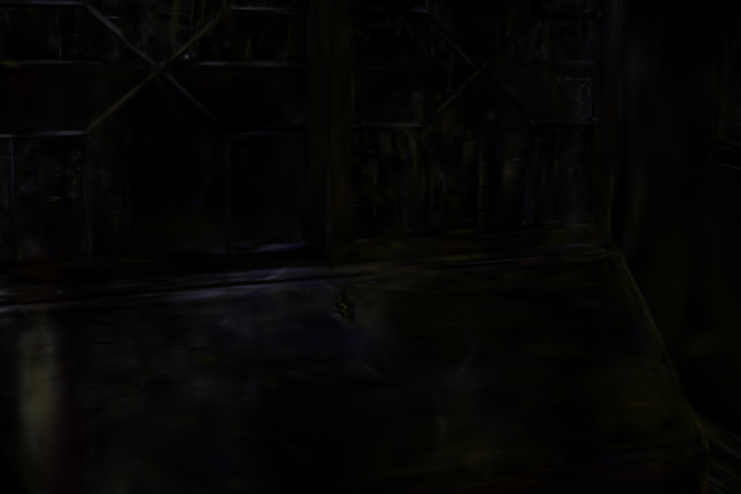}};
\node[ovpanel, anchor=north west, xshift=\dcgap] (ref3) at (b3r0c5.north east) {\dcref{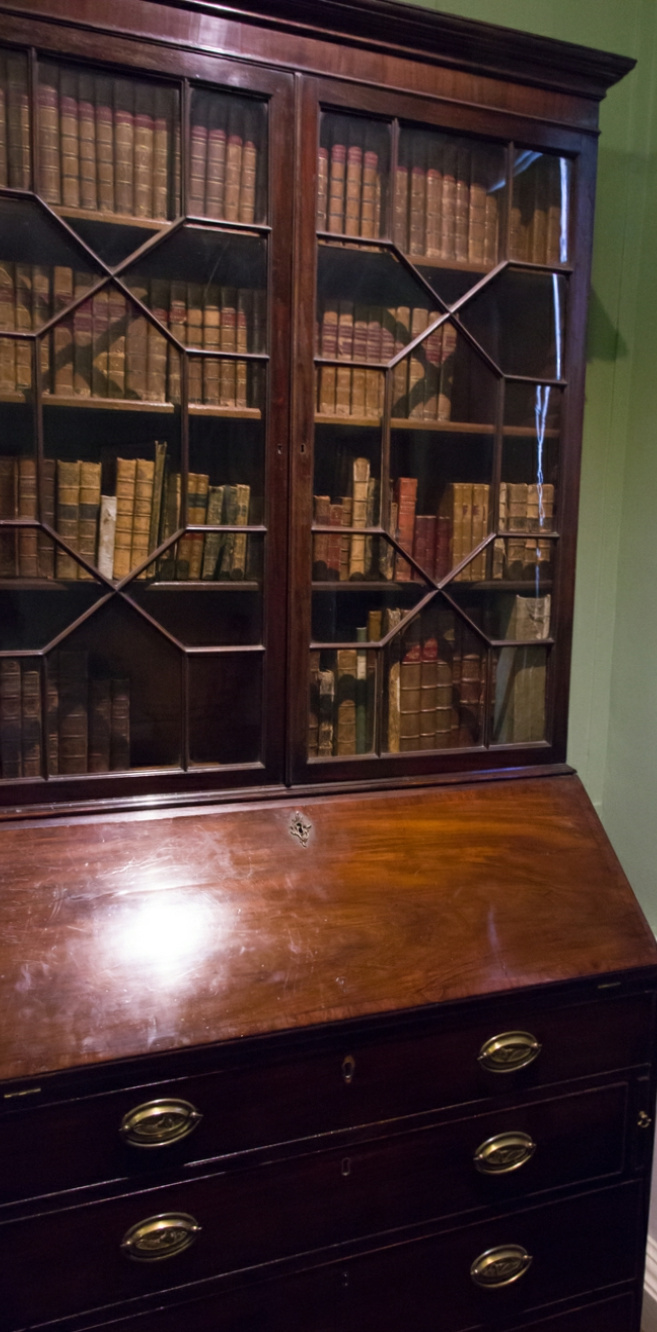}};
\dcheader{b0r0c0}{None}
\dcheader{b0r0c1}{SH}
\dcheader{b0r0c2}{SV}
\dcheader{b0r0c3}{NASG}
\dcheader{b0r0c4}{NASGabor}
\dcheader{b0r0c5}{Neural}
\dcheader{ref0}{Reference}
\dcrowlabel{b0r0c0}{Full}
\dcrowlabel{b0r1c0}{Base}
\dcrowlabel{b0r2c0}{Residual}
\dcscenelabel{ref0}{Counter}
\dcrowlabel{b1r0c0}{Full}
\dcrowlabel{b1r1c0}{Base}
\dcrowlabel{b1r2c0}{Residual}
\dcscenelabel{ref1}{Room}
\dcrowlabel{b2r0c0}{Full}
\dcrowlabel{b2r1c0}{Base}
\dcrowlabel{b2r2c0}{Residual}
\dcscenelabel{ref2}{Truck}
\dcrowlabel{b3r0c0}{Full}
\dcrowlabel{b3r1c0}{Base}
\dcrowlabel{b3r2c0}{Residual}
\dcscenelabel{ref3}{Dr Johnson}
\end{tikzpicture}}%
\caption{%
Base color and view-dependent residual, visualized as in \cref{fig:bonsai_block}, for a test view of four scenes. NASGabor achieves the best decomposition, with the residual limited to reflections such as those on the desk in \emph{Dr Johnson}. The SV residual often also models diffuse appearance. The neural model incorrectly represents the highlight on the desk in \emph{Dr Johnson} with the base color instead of the residual.
}\label{fig:decomposition_examples}
\end{figure*}

\cref{fig:decomposition_examples} shows the base color and residual of all models for a test view of four scenes, extending \cref{fig:bonsai_block}.
Consistent with the results in the main paper, NASGabor achieves the best decomposition: its residual only contains reflections, and its base color produces a reasonable diffuse rendering.
For SV, the residual often does too much work and absorbs part of the diffuse color, which could probably be partially resolved by using fewer sites.
The compact neural representation usually ranks somewhere between SH and NASGabor.
However, in the \emph{Dr Johnson} scene it fails to model the highlight on the desk through the residual, \ie, as a view-dependent effect.

\subsection{Per-Scene Results}
\label{sec:suppl_per_scene}
\begin{table*}[p]
\centering
\setlength\tabcolsep{4.7pt}
{\scriptsize
\begin{tabular}{lccccc|cccc|c}
\toprule
\multicolumn{11}{c}{PSNR$\uparrow$ on Mip-NeRF~360~\cite{barron2022mipnerf360}\vspace{2pt}}\\
Method & \textit{Bicycle} & \textit{Flowers} & \textit{Garden} & \textit{Stump} & \textit{Treehill} & \textit{Bonsai} & \textit{Counter} & \textit{Kitchen} & \textit{Room} & \textit{Average} \\
\midrule
None & 25.65\,{\tiny$\pm$0.0507} & 21.66\,{\tiny$\pm$0.0276} & 27.50\,{\tiny$\pm$0.0226} & 27.14\,{\tiny$\pm$0.0364} & 23.29\,{\tiny$\pm$0.0578} & 31.89\,{\tiny$\pm$0.0919} & 28.39\,{\tiny$\pm$0.0394} & 31.47\,{\tiny$\pm$0.0310} & 32.58\,{\tiny$\pm$0.0373} & 27.73\,{\tiny$\pm$0.0230} \\
SH & \cellcolor{1st}26.19\,{\tiny$\pm$0.0337} & \cellcolor{1st}22.15\,{\tiny$\pm$0.0238} & 27.96\,{\tiny$\pm$0.0167} & \cellcolor{1st}27.73\,{\tiny$\pm$0.0346} & \cellcolor{1st}23.40\,{\tiny$\pm$0.0275} & 32.97\,{\tiny$\pm$0.0251} & 29.49\,{\tiny$\pm$0.0348} & 32.35\,{\tiny$\pm$0.0354} & 32.98\,{\tiny$\pm$0.0550} & 28.36\,{\tiny$\pm$0.0094} \\
SV & \cellcolor{3rd}26.10\,{\tiny$\pm$0.0368} & \cellcolor{2nd}22.08\,{\tiny$\pm$0.0293} & \cellcolor{2nd}28.04\,{\tiny$\pm$0.0201} & 27.46\,{\tiny$\pm$0.0383} & 23.23\,{\tiny$\pm$0.0928} & \cellcolor{1st}34.11\,{\tiny$\pm$0.0272} & \cellcolor{1st}30.77\,{\tiny$\pm$0.0379} & \cellcolor{1st}32.81\,{\tiny$\pm$0.0617} & \cellcolor{3rd}33.16\,{\tiny$\pm$0.0591} & \cellcolor{1st}28.64\,{\tiny$\pm$0.0266} \\
NASG & \cellcolor{2nd}26.12\,{\tiny$\pm$0.0150} & \cellcolor{3rd}21.98\,{\tiny$\pm$0.0145} & 27.94\,{\tiny$\pm$0.0313} & \cellcolor{3rd}27.53\,{\tiny$\pm$0.0628} & \cellcolor{2nd}23.32\,{\tiny$\pm$0.0362} & 33.61\,{\tiny$\pm$0.0446} & 30.28\,{\tiny$\pm$0.0328} & 32.52\,{\tiny$\pm$0.0595} & \cellcolor{2nd}33.20\,{\tiny$\pm$0.0505} & 28.50\,{\tiny$\pm$0.0117} \\
NASGabor & \cellcolor{2nd}26.12\,{\tiny$\pm$0.0522} & 21.96\,{\tiny$\pm$0.0326} & \cellcolor{3rd}27.98\,{\tiny$\pm$0.0160} & \cellcolor{2nd}27.63\,{\tiny$\pm$0.0152} & \cellcolor{3rd}23.30\,{\tiny$\pm$0.0634} & \cellcolor{3rd}33.66\,{\tiny$\pm$0.0807} & \cellcolor{2nd}30.43\,{\tiny$\pm$0.0274} & \cellcolor{3rd}32.60\,{\tiny$\pm$0.0829} & \cellcolor{2nd}33.20\,{\tiny$\pm$0.0814} & \cellcolor{3rd}28.54\,{\tiny$\pm$0.0255} \\
Neural & 25.93\,{\tiny$\pm$0.0211} & 21.87\,{\tiny$\pm$0.0252} & \cellcolor{1st}28.06\,{\tiny$\pm$0.0344} & \cellcolor{3rd}27.53\,{\tiny$\pm$0.0418} & 23.28\,{\tiny$\pm$0.0190} & \cellcolor{2nd}33.94\,{\tiny$\pm$0.0662} & \cellcolor{3rd}30.38\,{\tiny$\pm$0.0872} & \cellcolor{2nd}32.69\,{\tiny$\pm$0.0108} & \cellcolor{1st}33.23\,{\tiny$\pm$0.0616} & \cellcolor{2nd}28.55\,{\tiny$\pm$0.0210} \\
\bottomrule
\end{tabular}
\begin{tabular}{lccccc|cccc|c}
\toprule
\multicolumn{11}{c}{SSIM$\uparrow$ on Mip-NeRF~360~\cite{barron2022mipnerf360}\vspace{2pt}}\\
Method & \textit{Bicycle} & \textit{Flowers} & \textit{Garden} & \textit{Stump} & \textit{Treehill} & \textit{Bonsai} & \textit{Counter} & \textit{Kitchen} & \textit{Room} & \textit{Average} \\
\midrule
None & 0.791\,{\tiny$\pm$0.0004} & 0.620\,{\tiny$\pm$0.0003} & 0.865\,{\tiny$\pm$0.0002} & 0.796\,{\tiny$\pm$0.0003} & \cellcolor{3rd}0.653\,{\tiny$\pm$0.0008} & 0.941\,{\tiny$\pm$0.0001} & 0.905\,{\tiny$\pm$0.0002} & 0.925\,{\tiny$\pm$0.0002} & 0.928\,{\tiny$\pm$0.0001} & 0.825\,{\tiny$\pm$0.0001} \\
SH & \cellcolor{2nd}0.802\,{\tiny$\pm$0.0002} & \cellcolor{1st}0.643\,{\tiny$\pm$0.0003} & \cellcolor{1st}0.874\,{\tiny$\pm$0.0004} & \cellcolor{3rd}0.809\,{\tiny$\pm$0.0004} & \cellcolor{2nd}0.659\,{\tiny$\pm$0.0003} & \cellcolor{3rd}0.949\,{\tiny$\pm$0.0001} & 0.919\,{\tiny$\pm$0.0001} & \cellcolor{2nd}0.933\,{\tiny$\pm$0.0001} & \cellcolor{2nd}0.932\,{\tiny$\pm$0.0001} & \cellcolor{1st}0.836\,{\tiny$\pm$0.0001} \\
SV & \cellcolor{3rd}0.797\,{\tiny$\pm$0.0004} & \cellcolor{3rd}0.638\,{\tiny$\pm$0.0007} & \cellcolor{2nd}0.873\,{\tiny$\pm$0.0002} & 0.802\,{\tiny$\pm$0.0003} & 0.652\,{\tiny$\pm$0.0004} & \cellcolor{1st}0.951\,{\tiny$\pm$0.0001} & \cellcolor{1st}0.926\,{\tiny$\pm$0.0001} & \cellcolor{1st}0.934\,{\tiny$\pm$0.0003} & \cellcolor{3rd}0.931\,{\tiny$\pm$0.0001} & \cellcolor{2nd}0.834\,{\tiny$\pm$0.0002} \\
NASG & \cellcolor{1st}0.804\,{\tiny$\pm$0.0003} & \cellcolor{2nd}0.639\,{\tiny$\pm$0.0005} & \cellcolor{2nd}0.873\,{\tiny$\pm$0.0004} & \cellcolor{2nd}0.812\,{\tiny$\pm$0.0005} & \cellcolor{1st}0.660\,{\tiny$\pm$0.0005} & \cellcolor{2nd}0.950\,{\tiny$\pm$0.0002} & \cellcolor{3rd}0.922\,{\tiny$\pm$0.0002} & \cellcolor{3rd}0.932\,{\tiny$\pm$0.0002} & \cellcolor{2nd}0.932\,{\tiny$\pm$0.0001} & \cellcolor{1st}0.836\,{\tiny$\pm$0.0001} \\
NASGabor & \cellcolor{1st}0.804\,{\tiny$\pm$0.0003} & \cellcolor{2nd}0.639\,{\tiny$\pm$0.0002} & \cellcolor{2nd}0.873\,{\tiny$\pm$0.0003} & \cellcolor{1st}0.813\,{\tiny$\pm$0.0001} & \cellcolor{2nd}0.659\,{\tiny$\pm$0.0003} & \cellcolor{1st}0.951\,{\tiny$\pm$0.0001} & \cellcolor{2nd}0.923\,{\tiny$\pm$0.0001} & \cellcolor{2nd}0.933\,{\tiny$\pm$0.0002} & \cellcolor{1st}0.933\,{\tiny$\pm$0.0003} & \cellcolor{1st}0.836\,{\tiny$\pm$0.0001} \\
Neural & 0.788\,{\tiny$\pm$0.0007} & 0.633\,{\tiny$\pm$0.0007} & \cellcolor{3rd}0.872\,{\tiny$\pm$0.0004} & 0.798\,{\tiny$\pm$0.0012} & 0.649\,{\tiny$\pm$0.0007} & \cellcolor{1st}0.951\,{\tiny$\pm$0.0002} & \cellcolor{3rd}0.922\,{\tiny$\pm$0.0004} & \cellcolor{2nd}0.933\,{\tiny$\pm$0.0002} & \cellcolor{1st}0.933\,{\tiny$\pm$0.0002} & \cellcolor{3rd}0.831\,{\tiny$\pm$0.0002} \\
\bottomrule
\end{tabular}
\begin{tabular}{lccccc|cccc|c}
\toprule
\multicolumn{11}{c}{LPIPS$\downarrow$ on Mip-NeRF~360~\cite{barron2022mipnerf360}\vspace{2pt}}\\
Method & \textit{Bicycle} & \textit{Flowers} & \textit{Garden} & \textit{Stump} & \textit{Treehill} & \textit{Bonsai} & \textit{Counter} & \textit{Kitchen} & \textit{Room} & \textit{Average} \\
\midrule
None & 0.196\,{\tiny$\pm$0.0003} & 0.333\,{\tiny$\pm$0.0007} & 0.124\,{\tiny$\pm$0.0003} & 0.216\,{\tiny$\pm$0.0008} & 0.325\,{\tiny$\pm$0.0008} & \cellcolor{3rd}0.242\,{\tiny$\pm$0.0005} & 0.249\,{\tiny$\pm$0.0002} & 0.160\,{\tiny$\pm$0.0003} & 0.264\,{\tiny$\pm$0.0002} & 0.234\,{\tiny$\pm$0.0001} \\
SH & \cellcolor{1st}0.184\,{\tiny$\pm$0.0004} & \cellcolor{1st}0.317\,{\tiny$\pm$0.0004} & \cellcolor{1st}0.112\,{\tiny$\pm$0.0002} & \cellcolor{2nd}0.201\,{\tiny$\pm$0.0007} & \cellcolor{1st}0.312\,{\tiny$\pm$0.0007} & \cellcolor{2nd}0.232\,{\tiny$\pm$0.0005} & 0.232\,{\tiny$\pm$0.0003} & \cellcolor{2nd}0.148\,{\tiny$\pm$0.0002} & \cellcolor{1st}0.258\,{\tiny$\pm$0.0002} & \cellcolor{1st}0.222\,{\tiny$\pm$0.0001} \\
SV & \cellcolor{3rd}0.191\,{\tiny$\pm$0.0003} & \cellcolor{3rd}0.322\,{\tiny$\pm$0.0003} & \cellcolor{2nd}0.114\,{\tiny$\pm$0.0002} & \cellcolor{3rd}0.207\,{\tiny$\pm$0.0004} & 0.322\,{\tiny$\pm$0.0009} & \cellcolor{1st}0.231\,{\tiny$\pm$0.0003} & \cellcolor{1st}0.226\,{\tiny$\pm$0.0004} & \cellcolor{1st}0.147\,{\tiny$\pm$0.0002} & \cellcolor{3rd}0.260\,{\tiny$\pm$0.0001} & \cellcolor{3rd}0.225\,{\tiny$\pm$0.0001} \\
NASG & \cellcolor{2nd}0.185\,{\tiny$\pm$0.0003} & \cellcolor{2nd}0.318\,{\tiny$\pm$0.0003} & 0.116\,{\tiny$\pm$0.0003} & \cellcolor{1st}0.200\,{\tiny$\pm$0.0005} & \cellcolor{2nd}0.318\,{\tiny$\pm$0.0006} & \cellcolor{1st}0.231\,{\tiny$\pm$0.0002} & \cellcolor{3rd}0.230\,{\tiny$\pm$0.0003} & 0.151\,{\tiny$\pm$0.0002} & \cellcolor{2nd}0.259\,{\tiny$\pm$0.0002} & \cellcolor{2nd}0.223\,{\tiny$\pm$0.0001} \\
NASGabor & \cellcolor{2nd}0.185\,{\tiny$\pm$0.0002} & \cellcolor{2nd}0.318\,{\tiny$\pm$0.0006} & \cellcolor{3rd}0.115\,{\tiny$\pm$0.0003} & \cellcolor{1st}0.200\,{\tiny$\pm$0.0003} & \cellcolor{3rd}0.319\,{\tiny$\pm$0.0007} & \cellcolor{1st}0.231\,{\tiny$\pm$0.0003} & \cellcolor{2nd}0.229\,{\tiny$\pm$0.0003} & 0.151\,{\tiny$\pm$0.0002} & \cellcolor{2nd}0.259\,{\tiny$\pm$0.0004} & \cellcolor{2nd}0.223\,{\tiny$\pm$0.0002} \\
Neural & 0.194\,{\tiny$\pm$0.0007} & 0.324\,{\tiny$\pm$0.0007} & 0.118\,{\tiny$\pm$0.0004} & 0.209\,{\tiny$\pm$0.0010} & 0.328\,{\tiny$\pm$0.0013} & \cellcolor{2nd}0.232\,{\tiny$\pm$0.0003} & 0.232\,{\tiny$\pm$0.0006} & \cellcolor{3rd}0.149\,{\tiny$\pm$0.0003} & \cellcolor{1st}0.258\,{\tiny$\pm$0.0004} & 0.227\,{\tiny$\pm$0.0001} \\
\bottomrule
\end{tabular}
\begin{tabular}{lccccc|cccc|c}
\toprule
\multicolumn{11}{c}{\FLIP$\downarrow$ on Mip-NeRF~360~\cite{barron2022mipnerf360}\vspace{2pt}}\\
Method & \textit{Bicycle} & \textit{Flowers} & \textit{Garden} & \textit{Stump} & \textit{Treehill} & \textit{Bonsai} & \textit{Counter} & \textit{Kitchen} & \textit{Room} & \textit{Average} \\
\midrule
None & 0.137\,{\tiny$\pm$0.0024} & 0.212\,{\tiny$\pm$0.0008} & 0.121\,{\tiny$\pm$0.0010} & 0.143\,{\tiny$\pm$0.0016} & 0.175\,{\tiny$\pm$0.0028} & 0.083\,{\tiny$\pm$0.0013} & 0.113\,{\tiny$\pm$0.0011} & 0.100\,{\tiny$\pm$0.0004} & 0.085\,{\tiny$\pm$0.0005} & \cellcolor{3rd}0.130\,{\tiny$\pm$0.0005} \\
SH & \cellcolor{2nd}0.129\,{\tiny$\pm$0.0012} & \cellcolor{1st}0.201\,{\tiny$\pm$0.0011} & \cellcolor{2nd}0.112\,{\tiny$\pm$0.0004} & \cellcolor{1st}0.133\,{\tiny$\pm$0.0015} & \cellcolor{1st}0.167\,{\tiny$\pm$0.0010} & \cellcolor{3rd}0.076\,{\tiny$\pm$0.0005} & 0.102\,{\tiny$\pm$0.0008} & \cellcolor{3rd}0.089\,{\tiny$\pm$0.0011} & \cellcolor{2nd}0.081\,{\tiny$\pm$0.0005} & \cellcolor{1st}0.121\,{\tiny$\pm$0.0002} \\
SV & \cellcolor{3rd}0.131\,{\tiny$\pm$0.0011} & \cellcolor{3rd}0.205\,{\tiny$\pm$0.0009} & \cellcolor{1st}0.110\,{\tiny$\pm$0.0004} & \cellcolor{2nd}0.135\,{\tiny$\pm$0.0013} & \cellcolor{3rd}0.172\,{\tiny$\pm$0.0067} & \cellcolor{1st}0.073\,{\tiny$\pm$0.0004} & \cellcolor{1st}0.093\,{\tiny$\pm$0.0004} & \cellcolor{1st}0.085\,{\tiny$\pm$0.0004} & \cellcolor{3rd}0.082\,{\tiny$\pm$0.0009} & \cellcolor{1st}0.121\,{\tiny$\pm$0.0009} \\
NASG & \cellcolor{1st}0.128\,{\tiny$\pm$0.0004} & \cellcolor{2nd}0.204\,{\tiny$\pm$0.0005} & 0.115\,{\tiny$\pm$0.0005} & 0.140\,{\tiny$\pm$0.0027} & \cellcolor{2nd}0.171\,{\tiny$\pm$0.0019} & \cellcolor{1st}0.073\,{\tiny$\pm$0.0005} & \cellcolor{3rd}0.096\,{\tiny$\pm$0.0004} & 0.091\,{\tiny$\pm$0.0009} & \cellcolor{1st}0.079\,{\tiny$\pm$0.0000} & \cellcolor{2nd}0.122\,{\tiny$\pm$0.0004} \\
NASGabor & \cellcolor{2nd}0.129\,{\tiny$\pm$0.0011} & \cellcolor{2nd}0.204\,{\tiny$\pm$0.0000} & 0.114\,{\tiny$\pm$0.0004} & \cellcolor{3rd}0.136\,{\tiny$\pm$0.0008} & \cellcolor{3rd}0.172\,{\tiny$\pm$0.0033} & \cellcolor{2nd}0.074\,{\tiny$\pm$0.0013} & \cellcolor{2nd}0.095\,{\tiny$\pm$0.0004} & 0.090\,{\tiny$\pm$0.0011} & \cellcolor{1st}0.079\,{\tiny$\pm$0.0011} & \cellcolor{1st}0.121\,{\tiny$\pm$0.0003} \\
Neural & 0.133\,{\tiny$\pm$0.0027} & 0.210\,{\tiny$\pm$0.0007} & \cellcolor{3rd}0.113\,{\tiny$\pm$0.0005} & \cellcolor{1st}0.133\,{\tiny$\pm$0.0009} & 0.173\,{\tiny$\pm$0.0016} & \cellcolor{1st}0.073\,{\tiny$\pm$0.0009} & \cellcolor{2nd}0.095\,{\tiny$\pm$0.0011} & \cellcolor{2nd}0.086\,{\tiny$\pm$0.0005} & \cellcolor{1st}0.079\,{\tiny$\pm$0.0007} & \cellcolor{2nd}0.122\,{\tiny$\pm$0.0005} \\
\bottomrule
\end{tabular}
\begin{tabular}{lccccc|cccc|c}
\toprule
\multicolumn{11}{c}{DISTS$\downarrow$ on Mip-NeRF~360~\cite{barron2022mipnerf360}\vspace{2pt}}\\
Method & \textit{Bicycle} & \textit{Flowers} & \textit{Garden} & \textit{Stump} & \textit{Treehill} & \textit{Bonsai} & \textit{Counter} & \textit{Kitchen} & \textit{Room} & \textit{Average} \\
\midrule
None & 0.054\,{\tiny$\pm$0.0000} & 0.113\,{\tiny$\pm$0.0004} & 0.037\,{\tiny$\pm$0.0004} & 0.074\,{\tiny$\pm$0.0008} & \cellcolor{2nd}0.115\,{\tiny$\pm$0.0007} & \cellcolor{3rd}0.077\,{\tiny$\pm$0.0004} & 0.058\,{\tiny$\pm$0.0000} & \cellcolor{2nd}0.052\,{\tiny$\pm$0.0004} & \cellcolor{1st}0.088\,{\tiny$\pm$0.0000} & 0.074\,{\tiny$\pm$0.0001} \\
SH & \cellcolor{1st}0.045\,{\tiny$\pm$0.0004} & \cellcolor{1st}0.108\,{\tiny$\pm$0.0004} & \cellcolor{1st}0.032\,{\tiny$\pm$0.0000} & \cellcolor{1st}0.067\,{\tiny$\pm$0.0004} & \cellcolor{1st}0.111\,{\tiny$\pm$0.0007} & \cellcolor{1st}0.075\,{\tiny$\pm$0.0004} & 0.056\,{\tiny$\pm$0.0000} & \cellcolor{1st}0.050\,{\tiny$\pm$0.0000} & \cellcolor{1st}0.088\,{\tiny$\pm$0.0005} & \cellcolor{1st}0.070\,{\tiny$\pm$0.0001} \\
SV & \cellcolor{2nd}0.047\,{\tiny$\pm$0.0000} & \cellcolor{2nd}0.110\,{\tiny$\pm$0.0005} & \cellcolor{1st}0.032\,{\tiny$\pm$0.0005} & \cellcolor{3rd}0.070\,{\tiny$\pm$0.0011} & \cellcolor{3rd}0.116\,{\tiny$\pm$0.0008} & \cellcolor{1st}0.075\,{\tiny$\pm$0.0000} & \cellcolor{1st}0.053\,{\tiny$\pm$0.0000} & \cellcolor{1st}0.050\,{\tiny$\pm$0.0000} & \cellcolor{1st}0.088\,{\tiny$\pm$0.0000} & \cellcolor{2nd}0.071\,{\tiny$\pm$0.0003} \\
NASG & 0.049\,{\tiny$\pm$0.0004} & \cellcolor{2nd}0.110\,{\tiny$\pm$0.0004} & \cellcolor{2nd}0.034\,{\tiny$\pm$0.0000} & \cellcolor{3rd}0.070\,{\tiny$\pm$0.0009} & \cellcolor{2nd}0.115\,{\tiny$\pm$0.0009} & \cellcolor{2nd}0.076\,{\tiny$\pm$0.0000} & \cellcolor{3rd}0.055\,{\tiny$\pm$0.0004} & \cellcolor{1st}0.050\,{\tiny$\pm$0.0000} & \cellcolor{1st}0.088\,{\tiny$\pm$0.0000} & \cellcolor{3rd}0.072\,{\tiny$\pm$0.0002} \\
NASGabor & \cellcolor{3rd}0.048\,{\tiny$\pm$0.0005} & \cellcolor{3rd}0.111\,{\tiny$\pm$0.0009} & \cellcolor{2nd}0.034\,{\tiny$\pm$0.0004} & \cellcolor{2nd}0.068\,{\tiny$\pm$0.0005} & \cellcolor{2nd}0.115\,{\tiny$\pm$0.0000} & \cellcolor{2nd}0.076\,{\tiny$\pm$0.0000} & \cellcolor{2nd}0.054\,{\tiny$\pm$0.0004} & \cellcolor{1st}0.050\,{\tiny$\pm$0.0000} & \cellcolor{1st}0.088\,{\tiny$\pm$0.0000} & \cellcolor{3rd}0.072\,{\tiny$\pm$0.0001} \\
Neural & 0.049\,{\tiny$\pm$0.0000} & 0.112\,{\tiny$\pm$0.0007} & \cellcolor{3rd}0.035\,{\tiny$\pm$0.0004} & 0.072\,{\tiny$\pm$0.0011} & 0.121\,{\tiny$\pm$0.0011} & \cellcolor{1st}0.075\,{\tiny$\pm$0.0005} & \cellcolor{3rd}0.055\,{\tiny$\pm$0.0004} & \cellcolor{1st}0.050\,{\tiny$\pm$0.0000} & \cellcolor{1st}0.088\,{\tiny$\pm$0.0004} & 0.073\,{\tiny$\pm$0.0001} \\
\bottomrule
\end{tabular}
\begin{tabular}{lccccc|cccc|c}
\toprule
\multicolumn{11}{c}{MILO$\uparrow$ on Mip-NeRF~360~\cite{barron2022mipnerf360}\vspace{2pt}}\\
Method & \textit{Bicycle} & \textit{Flowers} & \textit{Garden} & \textit{Stump} & \textit{Treehill} & \textit{Bonsai} & \textit{Counter} & \textit{Kitchen} & \textit{Room} & \textit{Average} \\
\midrule
None & 3.403\,{\tiny$\pm$0.0038} & 2.579\,{\tiny$\pm$0.0035} & 3.660\,{\tiny$\pm$0.0019} & 3.185\,{\tiny$\pm$0.0011} & 2.868\,{\tiny$\pm$0.0031} & 3.641\,{\tiny$\pm$0.0018} & 3.100\,{\tiny$\pm$0.0041} & 3.626\,{\tiny$\pm$0.0026} & 3.468\,{\tiny$\pm$0.0020} & 3.281\,{\tiny$\pm$0.0015} \\
SH & \cellcolor{1st}3.495\,{\tiny$\pm$0.0034} & \cellcolor{1st}2.637\,{\tiny$\pm$0.0055} & \cellcolor{1st}3.730\,{\tiny$\pm$0.0025} & \cellcolor{2nd}3.281\,{\tiny$\pm$0.0039} & \cellcolor{1st}2.925\,{\tiny$\pm$0.0038} & 3.784\,{\tiny$\pm$0.0040} & 3.259\,{\tiny$\pm$0.0068} & \cellcolor{3rd}3.706\,{\tiny$\pm$0.0074} & 3.512\,{\tiny$\pm$0.0020} & \cellcolor{3rd}3.370\,{\tiny$\pm$0.0016} \\
SV & \cellcolor{2nd}3.476\,{\tiny$\pm$0.0046} & 2.630\,{\tiny$\pm$0.0051} & \cellcolor{2nd}3.728\,{\tiny$\pm$0.0054} & 3.246\,{\tiny$\pm$0.0037} & 2.865\,{\tiny$\pm$0.0076} & \cellcolor{1st}3.823\,{\tiny$\pm$0.0029} & \cellcolor{1st}3.363\,{\tiny$\pm$0.0100} & \cellcolor{1st}3.723\,{\tiny$\pm$0.0037} & \cellcolor{2nd}3.523\,{\tiny$\pm$0.0027} & \cellcolor{1st}3.375\,{\tiny$\pm$0.0013} \\
NASG & \cellcolor{3rd}3.475\,{\tiny$\pm$0.0021} & \cellcolor{2nd}2.633\,{\tiny$\pm$0.0011} & 3.720\,{\tiny$\pm$0.0049} & \cellcolor{3rd}3.278\,{\tiny$\pm$0.0023} & \cellcolor{2nd}2.901\,{\tiny$\pm$0.0034} & 3.795\,{\tiny$\pm$0.0049} & \cellcolor{3rd}3.311\,{\tiny$\pm$0.0099} & 3.703\,{\tiny$\pm$0.0050} & 3.516\,{\tiny$\pm$0.0043} & \cellcolor{3rd}3.370\,{\tiny$\pm$0.0011} \\
NASGabor & \cellcolor{3rd}3.475\,{\tiny$\pm$0.0033} & \cellcolor{3rd}2.632\,{\tiny$\pm$0.0063} & \cellcolor{3rd}3.723\,{\tiny$\pm$0.0041} & \cellcolor{1st}3.282\,{\tiny$\pm$0.0022} & \cellcolor{3rd}2.895\,{\tiny$\pm$0.0040} & \cellcolor{3rd}3.796\,{\tiny$\pm$0.0019} & \cellcolor{2nd}3.318\,{\tiny$\pm$0.0069} & 3.702\,{\tiny$\pm$0.0053} & \cellcolor{3rd}3.517\,{\tiny$\pm$0.0057} & \cellcolor{2nd}3.371\,{\tiny$\pm$0.0020} \\
Neural & 3.438\,{\tiny$\pm$0.0047} & 2.606\,{\tiny$\pm$0.0072} & 3.720\,{\tiny$\pm$0.0033} & 3.233\,{\tiny$\pm$0.0040} & 2.881\,{\tiny$\pm$0.0015} & \cellcolor{2nd}3.801\,{\tiny$\pm$0.0044} & 3.293\,{\tiny$\pm$0.0065} & \cellcolor{2nd}3.709\,{\tiny$\pm$0.0033} & \cellcolor{1st}3.524\,{\tiny$\pm$0.0052} & 3.356\,{\tiny$\pm$0.0011} \\
\bottomrule
\end{tabular}%
}
\caption{
Per-scene image quality metrics on Mip-NeRF~360~\cite{barron2022mipnerf360}, averaged over five training runs with the standard deviation across runs.
The three best results are highlighted in \textcolor{1stText}{\textbf{green}} in descending order of saturation.
}\label{tab:per_scene_m360}%
\end{table*}

\begin{table*}[p]
\centering
\setlength\tabcolsep{23.7pt}
{\scriptsize
\begin{tabular}{lcc|cc|c}
\toprule
\multicolumn{6}{c}{PSNR$\uparrow$ on Tanks \& Temples~\cite{Knapitsch2017}~$\slash$~Deep Blending~\cite{hedman2018deep}\vspace{2pt}}\\
Method & \textit{Train} & \textit{Truck} & \textit{Drjohnson} & \textit{Playroom} & \textit{Average} \\
\midrule
None & 24.10\,{\tiny$\pm$0.1286} & 26.78\,{\tiny$\pm$0.0207} & 29.34\,{\tiny$\pm$0.0503} & 30.39\,{\tiny$\pm$0.1534} & 27.65\,{\tiny$\pm$0.0207} \\
SH & 24.48\,{\tiny$\pm$0.1266} & 27.13\,{\tiny$\pm$0.0268} & 29.43\,{\tiny$\pm$0.0867} & \cellcolor{2nd}30.80\,{\tiny$\pm$0.0714} & 27.96\,{\tiny$\pm$0.0294} \\
SV & \cellcolor{1st}24.54\,{\tiny$\pm$0.0993} & \cellcolor{1st}27.37\,{\tiny$\pm$0.0263} & 29.33\,{\tiny$\pm$0.0875} & 30.68\,{\tiny$\pm$0.1853} & \cellcolor{2nd}27.98\,{\tiny$\pm$0.0357} \\
NASG & 24.39\,{\tiny$\pm$0.2059} & 27.21\,{\tiny$\pm$0.0164} & \cellcolor{2nd}29.55\,{\tiny$\pm$0.1153} & \cellcolor{3rd}30.73\,{\tiny$\pm$0.1417} & \cellcolor{3rd}27.97\,{\tiny$\pm$0.0396} \\
NASGabor & \cellcolor{2nd}24.53\,{\tiny$\pm$0.1562} & \cellcolor{3rd}27.22\,{\tiny$\pm$0.0288} & \cellcolor{3rd}29.51\,{\tiny$\pm$0.0779} & 30.60\,{\tiny$\pm$0.3347} & 27.96\,{\tiny$\pm$0.0855} \\
Neural & \cellcolor{3rd}24.49\,{\tiny$\pm$0.0959} & \cellcolor{2nd}27.23\,{\tiny$\pm$0.0056} & \cellcolor{1st}29.72\,{\tiny$\pm$0.2136} & \cellcolor{1st}31.08\,{\tiny$\pm$0.1384} & \cellcolor{1st}28.13\,{\tiny$\pm$0.0702} \\
\bottomrule
\end{tabular}
\begin{tabular}{lcc|cc|c}
\toprule
\multicolumn{6}{c}{SSIM$\uparrow$ on Tanks \& Temples~\cite{Knapitsch2017}~$\slash$~Deep Blending~\cite{hedman2018deep}\vspace{2pt}}\\
Method & \textit{Train} & \textit{Truck} & \textit{Drjohnson} & \textit{Playroom} & \textit{Average} \\
\midrule
None & 0.846\,{\tiny$\pm$0.0017} & \cellcolor{3rd}0.898\,{\tiny$\pm$0.0002} & 0.900\,{\tiny$\pm$0.0004} & 0.906\,{\tiny$\pm$0.0017} & \cellcolor{2nd}0.888\,{\tiny$\pm$0.0004} \\
SH & \cellcolor{1st}0.856\,{\tiny$\pm$0.0005} & \cellcolor{2nd}0.904\,{\tiny$\pm$0.0002} & \cellcolor{3rd}0.902\,{\tiny$\pm$0.0006} & \cellcolor{3rd}0.910\,{\tiny$\pm$0.0002} & \cellcolor{1st}0.893\,{\tiny$\pm$0.0001} \\
SV & \cellcolor{1st}0.856\,{\tiny$\pm$0.0010} & \cellcolor{2nd}0.904\,{\tiny$\pm$0.0001} & 0.900\,{\tiny$\pm$0.0006} & \cellcolor{3rd}0.910\,{\tiny$\pm$0.0007} & \cellcolor{1st}0.893\,{\tiny$\pm$0.0002} \\
NASG & \cellcolor{3rd}0.853\,{\tiny$\pm$0.0015} & \cellcolor{1st}0.905\,{\tiny$\pm$0.0001} & \cellcolor{1st}0.904\,{\tiny$\pm$0.0004} & \cellcolor{1st}0.912\,{\tiny$\pm$0.0002} & \cellcolor{1st}0.893\,{\tiny$\pm$0.0003} \\
NASGabor & \cellcolor{3rd}0.853\,{\tiny$\pm$0.0010} & \cellcolor{1st}0.905\,{\tiny$\pm$0.0001} & \cellcolor{1st}0.904\,{\tiny$\pm$0.0004} & \cellcolor{2nd}0.911\,{\tiny$\pm$0.0015} & \cellcolor{1st}0.893\,{\tiny$\pm$0.0004} \\
Neural & \cellcolor{2nd}0.854\,{\tiny$\pm$0.0015} & \cellcolor{2nd}0.904\,{\tiny$\pm$0.0001} & \cellcolor{2nd}0.903\,{\tiny$\pm$0.0004} & \cellcolor{3rd}0.910\,{\tiny$\pm$0.0020} & \cellcolor{1st}0.893\,{\tiny$\pm$0.0008} \\
\bottomrule
\end{tabular}
\begin{tabular}{lcc|cc|c}
\toprule
\multicolumn{6}{c}{LPIPS$\downarrow$ on Tanks \& Temples~\cite{Knapitsch2017}~$\slash$~Deep Blending~\cite{hedman2018deep}\vspace{2pt}}\\
Method & \textit{Train} & \textit{Truck} & \textit{Drjohnson} & \textit{Playroom} & \textit{Average} \\
\midrule
None & 0.211\,{\tiny$\pm$0.0018} & 0.127\,{\tiny$\pm$0.0001} & 0.309\,{\tiny$\pm$0.0002} & 0.300\,{\tiny$\pm$0.0007} & 0.237\,{\tiny$\pm$0.0005} \\
SH & \cellcolor{1st}0.197\,{\tiny$\pm$0.0008} & \cellcolor{1st}0.117\,{\tiny$\pm$0.0003} & 0.305\,{\tiny$\pm$0.0004} & 0.290\,{\tiny$\pm$0.0005} & \cellcolor{2nd}0.228\,{\tiny$\pm$0.0003} \\
SV & \cellcolor{2nd}0.199\,{\tiny$\pm$0.0006} & \cellcolor{3rd}0.121\,{\tiny$\pm$0.0002} & 0.311\,{\tiny$\pm$0.0005} & 0.296\,{\tiny$\pm$0.0010} & \cellcolor{3rd}0.232\,{\tiny$\pm$0.0003} \\
NASG & 0.204\,{\tiny$\pm$0.0009} & \cellcolor{1st}0.117\,{\tiny$\pm$0.0002} & \cellcolor{2nd}0.301\,{\tiny$\pm$0.0004} & \cellcolor{2nd}0.288\,{\tiny$\pm$0.0007} & \cellcolor{1st}0.227\,{\tiny$\pm$0.0002} \\
NASGabor & 0.203\,{\tiny$\pm$0.0004} & \cellcolor{1st}0.117\,{\tiny$\pm$0.0003} & \cellcolor{1st}0.300\,{\tiny$\pm$0.0005} & \cellcolor{1st}0.287\,{\tiny$\pm$0.0008} & \cellcolor{1st}0.227\,{\tiny$\pm$0.0004} \\
Neural & \cellcolor{3rd}0.202\,{\tiny$\pm$0.0013} & \cellcolor{2nd}0.118\,{\tiny$\pm$0.0002} & \cellcolor{3rd}0.303\,{\tiny$\pm$0.0018} & \cellcolor{3rd}0.289\,{\tiny$\pm$0.0001} & \cellcolor{2nd}0.228\,{\tiny$\pm$0.0004} \\
\bottomrule
\end{tabular}
\begin{tabular}{lcc|cc|c}
\toprule
\multicolumn{6}{c}{\FLIP$\downarrow$ on Tanks \& Temples~\cite{Knapitsch2017}~$\slash$~Deep Blending~\cite{hedman2018deep}\vspace{2pt}}\\
Method & \textit{Train} & \textit{Truck} & \textit{Drjohnson} & \textit{Playroom} & \textit{Average} \\
\midrule
None & 0.218\,{\tiny$\pm$0.0053} & 0.117\,{\tiny$\pm$0.0000} & 0.117\,{\tiny$\pm$0.0007} & 0.140\,{\tiny$\pm$0.0043} & 0.148\,{\tiny$\pm$0.0008} \\
SH & \cellcolor{1st}0.209\,{\tiny$\pm$0.0037} & \cellcolor{3rd}0.110\,{\tiny$\pm$0.0011} & 0.119\,{\tiny$\pm$0.0011} & \cellcolor{2nd}0.134\,{\tiny$\pm$0.0013} & \cellcolor{2nd}0.143\,{\tiny$\pm$0.0007} \\
SV & \cellcolor{2nd}0.211\,{\tiny$\pm$0.0038} & \cellcolor{1st}0.108\,{\tiny$\pm$0.0005} & 0.120\,{\tiny$\pm$0.0015} & \cellcolor{2nd}0.134\,{\tiny$\pm$0.0021} & \cellcolor{2nd}0.143\,{\tiny$\pm$0.0010} \\
NASG & 0.214\,{\tiny$\pm$0.0073} & \cellcolor{2nd}0.109\,{\tiny$\pm$0.0011} & \cellcolor{3rd}0.116\,{\tiny$\pm$0.0005} & \cellcolor{3rd}0.135\,{\tiny$\pm$0.0030} & \cellcolor{3rd}0.144\,{\tiny$\pm$0.0015} \\
NASGabor & \cellcolor{1st}0.209\,{\tiny$\pm$0.0050} & \cellcolor{2nd}0.109\,{\tiny$\pm$0.0005} & \cellcolor{2nd}0.115\,{\tiny$\pm$0.0013} & 0.138\,{\tiny$\pm$0.0064} & \cellcolor{2nd}0.143\,{\tiny$\pm$0.0016} \\
Neural & \cellcolor{3rd}0.213\,{\tiny$\pm$0.0038} & \cellcolor{3rd}0.110\,{\tiny$\pm$0.0005} & \cellcolor{1st}0.114\,{\tiny$\pm$0.0040} & \cellcolor{1st}0.129\,{\tiny$\pm$0.0013} & \cellcolor{1st}0.141\,{\tiny$\pm$0.0014} \\
\bottomrule
\end{tabular}
\begin{tabular}{lcc|cc|c}
\toprule
\multicolumn{6}{c}{DISTS$\downarrow$ on Tanks \& Temples~\cite{Knapitsch2017}~$\slash$~Deep Blending~\cite{hedman2018deep}\vspace{2pt}}\\
Method & \textit{Train} & \textit{Truck} & \textit{Drjohnson} & \textit{Playroom} & \textit{Average} \\
\midrule
None & 0.073\,{\tiny$\pm$0.0005} & \cellcolor{3rd}0.035\,{\tiny$\pm$0.0000} & 0.121\,{\tiny$\pm$0.0000} & \cellcolor{3rd}0.113\,{\tiny$\pm$0.0008} & 0.086\,{\tiny$\pm$0.0001} \\
SH & \cellcolor{1st}0.067\,{\tiny$\pm$0.0004} & \cellcolor{1st}0.032\,{\tiny$\pm$0.0000} & \cellcolor{2nd}0.118\,{\tiny$\pm$0.0000} & \cellcolor{1st}0.107\,{\tiny$\pm$0.0000} & \cellcolor{1st}0.081\,{\tiny$\pm$0.0001} \\
SV & \cellcolor{2nd}0.068\,{\tiny$\pm$0.0004} & \cellcolor{2nd}0.033\,{\tiny$\pm$0.0005} & 0.121\,{\tiny$\pm$0.0005} & \cellcolor{2nd}0.111\,{\tiny$\pm$0.0004} & \cellcolor{3rd}0.083\,{\tiny$\pm$0.0002} \\
NASG & 0.071\,{\tiny$\pm$0.0005} & \cellcolor{1st}0.032\,{\tiny$\pm$0.0000} & \cellcolor{1st}0.117\,{\tiny$\pm$0.0000} & \cellcolor{1st}0.107\,{\tiny$\pm$0.0000} & \cellcolor{2nd}0.082\,{\tiny$\pm$0.0001} \\
NASGabor & 0.071\,{\tiny$\pm$0.0005} & \cellcolor{1st}0.032\,{\tiny$\pm$0.0000} & \cellcolor{1st}0.117\,{\tiny$\pm$0.0005} & \cellcolor{1st}0.107\,{\tiny$\pm$0.0008} & \cellcolor{2nd}0.082\,{\tiny$\pm$0.0003} \\
Neural & \cellcolor{3rd}0.070\,{\tiny$\pm$0.0007} & \cellcolor{1st}0.032\,{\tiny$\pm$0.0000} & \cellcolor{3rd}0.120\,{\tiny$\pm$0.0039} & \cellcolor{1st}0.107\,{\tiny$\pm$0.0004} & \cellcolor{2nd}0.082\,{\tiny$\pm$0.0008} \\
\bottomrule
\end{tabular}
\begin{tabular}{lcc|cc|c}
\toprule
\multicolumn{6}{c}{MILO$\uparrow$ on Tanks \& Temples~\cite{Knapitsch2017}~$\slash$~Deep Blending~\cite{hedman2018deep}\vspace{2pt}}\\
Method & \textit{Train} & \textit{Truck} & \textit{Drjohnson} & \textit{Playroom} & \textit{Average} \\
\midrule
None & 3.088\,{\tiny$\pm$0.0164} & 3.472\,{\tiny$\pm$0.0038} & 3.180\,{\tiny$\pm$0.0036} & 3.391\,{\tiny$\pm$0.0090} & 3.283\,{\tiny$\pm$0.0037} \\
SH & \cellcolor{2nd}3.169\,{\tiny$\pm$0.0115} & 3.544\,{\tiny$\pm$0.0050} & \cellcolor{3rd}3.212\,{\tiny$\pm$0.0021} & 3.426\,{\tiny$\pm$0.0061} & 3.338\,{\tiny$\pm$0.0032} \\
SV & \cellcolor{1st}3.187\,{\tiny$\pm$0.0068} & \cellcolor{1st}3.568\,{\tiny$\pm$0.0028} & 3.188\,{\tiny$\pm$0.0042} & 3.419\,{\tiny$\pm$0.0133} & \cellcolor{3rd}3.340\,{\tiny$\pm$0.0031} \\
NASG & 3.145\,{\tiny$\pm$0.0098} & \cellcolor{3rd}3.548\,{\tiny$\pm$0.0053} & \cellcolor{1st}3.234\,{\tiny$\pm$0.0024} & \cellcolor{1st}3.452\,{\tiny$\pm$0.0053} & \cellcolor{2nd}3.345\,{\tiny$\pm$0.0021} \\
NASGabor & 3.148\,{\tiny$\pm$0.0076} & \cellcolor{2nd}3.552\,{\tiny$\pm$0.0034} & \cellcolor{1st}3.234\,{\tiny$\pm$0.0045} & \cellcolor{2nd}3.451\,{\tiny$\pm$0.0109} & \cellcolor{1st}3.346\,{\tiny$\pm$0.0040} \\
Neural & \cellcolor{3rd}3.157\,{\tiny$\pm$0.0136} & 3.547\,{\tiny$\pm$0.0034} & \cellcolor{2nd}3.230\,{\tiny$\pm$0.0057} & \cellcolor{3rd}3.444\,{\tiny$\pm$0.0084} & \cellcolor{2nd}3.345\,{\tiny$\pm$0.0044} \\
\bottomrule
\end{tabular}%
}
\caption{
Per-scene image quality metrics on Tanks \& Temples~\cite{Knapitsch2017}~$\slash$~Deep Blending~\cite{hedman2018deep}, averaged over five training runs with the standard deviation across runs.
The three best results are highlighted in \textcolor{1stText}{\textbf{green}} in descending order of saturation.
}\label{tab:per_scene_tandt_db}%
\end{table*}

\begin{table*}[p]
\centering
\setlength\tabcolsep{7pt}
{\scriptsize
\begin{tabular}{lcccccccc|c}
\toprule
\multicolumn{10}{c}{PSNR$\uparrow$ on NeRF Synthetic~\cite{mildenhall2020nerf}\vspace{2pt}}\\
Method & \textit{Chair} & \textit{Drums} & \textit{Ficus} & \textit{Hotdog} & \textit{Lego} & \textit{Materials} & \textit{Mic} & \textit{Ship} & \textit{Average} \\
\midrule
None & 33.60\,{\tiny$\pm$0.0199} & 25.40\,{\tiny$\pm$0.0390} & 32.04\,{\tiny$\pm$0.0585} & 33.49\,{\tiny$\pm$0.4511} & 34.83\,{\tiny$\pm$0.0125} & 28.22\,{\tiny$\pm$0.0821} & 34.70\,{\tiny$\pm$0.0382} & 29.36\,{\tiny$\pm$0.1582} & 31.46\,{\tiny$\pm$0.0873} \\
SH & \cellcolor{2nd}35.98\,{\tiny$\pm$0.0111} & 26.28\,{\tiny$\pm$0.0259} & \cellcolor{2nd}35.46\,{\tiny$\pm$0.3506} & \cellcolor{1st}38.35\,{\tiny$\pm$0.0920} & \cellcolor{2nd}36.14\,{\tiny$\pm$0.0059} & \cellcolor{2nd}30.91\,{\tiny$\pm$0.0060} & \cellcolor{3rd}37.48\,{\tiny$\pm$0.0638} & \cellcolor{2nd}31.44\,{\tiny$\pm$0.2776} & \cellcolor{2nd}34.01\,{\tiny$\pm$0.0478} \\
SV & \cellcolor{1st}36.34\,{\tiny$\pm$0.0074} & \cellcolor{1st}27.00\,{\tiny$\pm$0.0316} & 30.65\,{\tiny$\pm$2.3306} & \cellcolor{3rd}36.53\,{\tiny$\pm$4.2332} & \cellcolor{1st}36.36\,{\tiny$\pm$0.0174} & \cellcolor{1st}31.08\,{\tiny$\pm$0.0142} & \cellcolor{1st}37.80\,{\tiny$\pm$0.0144} & \cellcolor{1st}31.88\,{\tiny$\pm$0.0505} & \cellcolor{3rd}33.45\,{\tiny$\pm$0.6306} \\
NASG & 34.77\,{\tiny$\pm$0.0090} & \cellcolor{3rd}26.34\,{\tiny$\pm$0.0389} & \cellcolor{3rd}33.40\,{\tiny$\pm$0.0802} & 36.35\,{\tiny$\pm$0.9041} & 35.60\,{\tiny$\pm$0.0560} & 28.94\,{\tiny$\pm$0.0239} & 35.25\,{\tiny$\pm$3.1155} & 30.87\,{\tiny$\pm$0.1773} & 32.69\,{\tiny$\pm$0.3724} \\
NASGabor & 34.79\,{\tiny$\pm$0.0145} & 26.32\,{\tiny$\pm$0.0196} & 33.35\,{\tiny$\pm$0.0367} & \cellcolor{2nd}37.03\,{\tiny$\pm$0.6098} & 35.70\,{\tiny$\pm$0.0144} & 28.96\,{\tiny$\pm$0.0865} & 36.59\,{\tiny$\pm$0.0413} & 30.89\,{\tiny$\pm$0.1202} & 32.95\,{\tiny$\pm$0.0763} \\
Neural & \cellcolor{3rd}35.70\,{\tiny$\pm$0.0637} & \cellcolor{2nd}26.75\,{\tiny$\pm$0.0596} & \cellcolor{1st}35.60\,{\tiny$\pm$0.0372} & \cellcolor{1st}38.35\,{\tiny$\pm$0.0588} & \cellcolor{3rd}36.10\,{\tiny$\pm$0.0389} & \cellcolor{3rd}30.88\,{\tiny$\pm$0.0439} & \cellcolor{2nd}37.68\,{\tiny$\pm$0.0339} & \cellcolor{3rd}31.28\,{\tiny$\pm$0.1358} & \cellcolor{1st}34.04\,{\tiny$\pm$0.0241} \\
\bottomrule
\end{tabular}
\begin{tabular}{lcccccccc|c}
\toprule
\multicolumn{10}{c}{SSIM$\uparrow$ on NeRF Synthetic~\cite{mildenhall2020nerf}\vspace{2pt}}\\
Method & \textit{Chair} & \textit{Drums} & \textit{Ficus} & \textit{Hotdog} & \textit{Lego} & \textit{Materials} & \textit{Mic} & \textit{Ship} & \textit{Average} \\
\midrule
None & 0.983\,{\tiny$\pm$0.0000} & 0.949\,{\tiny$\pm$0.0002} & \cellcolor{3rd}0.978\,{\tiny$\pm$0.0001} & 0.964\,{\tiny$\pm$0.0057} & \cellcolor{3rd}0.981\,{\tiny$\pm$0.0001} & 0.948\,{\tiny$\pm$0.0005} & \cellcolor{3rd}0.989\,{\tiny$\pm$0.0001} & 0.881\,{\tiny$\pm$0.0041} & 0.959\,{\tiny$\pm$0.0012} \\
SH & \cellcolor{1st}0.989\,{\tiny$\pm$0.0000} & \cellcolor{2nd}0.956\,{\tiny$\pm$0.0002} & \cellcolor{1st}0.987\,{\tiny$\pm$0.0008} & \cellcolor{1st}0.987\,{\tiny$\pm$0.0005} & \cellcolor{1st}0.984\,{\tiny$\pm$0.0000} & \cellcolor{2nd}0.964\,{\tiny$\pm$0.0001} & \cellcolor{1st}0.994\,{\tiny$\pm$0.0001} & \cellcolor{2nd}0.903\,{\tiny$\pm$0.0022} & \cellcolor{1st}0.971\,{\tiny$\pm$0.0003} \\
SV & \cellcolor{1st}0.989\,{\tiny$\pm$0.0000} & \cellcolor{1st}0.959\,{\tiny$\pm$0.0002} & 0.970\,{\tiny$\pm$0.0086} & 0.978\,{\tiny$\pm$0.0210} & \cellcolor{1st}0.984\,{\tiny$\pm$0.0000} & \cellcolor{1st}0.966\,{\tiny$\pm$0.0001} & \cellcolor{1st}0.994\,{\tiny$\pm$0.0000} & \cellcolor{1st}0.906\,{\tiny$\pm$0.0014} & \cellcolor{3rd}0.968\,{\tiny$\pm$0.0029} \\
NASG & \cellcolor{3rd}0.986\,{\tiny$\pm$0.0000} & \cellcolor{3rd}0.955\,{\tiny$\pm$0.0002} & \cellcolor{2nd}0.983\,{\tiny$\pm$0.0001} & \cellcolor{3rd}0.980\,{\tiny$\pm$0.0070} & \cellcolor{2nd}0.983\,{\tiny$\pm$0.0001} & 0.953\,{\tiny$\pm$0.0002} & 0.987\,{\tiny$\pm$0.0126} & 0.897\,{\tiny$\pm$0.0022} & 0.966\,{\tiny$\pm$0.0017} \\
NASGabor & \cellcolor{3rd}0.986\,{\tiny$\pm$0.0000} & \cellcolor{3rd}0.955\,{\tiny$\pm$0.0001} & \cellcolor{2nd}0.983\,{\tiny$\pm$0.0000} & \cellcolor{2nd}0.983\,{\tiny$\pm$0.0021} & \cellcolor{2nd}0.983\,{\tiny$\pm$0.0000} & 0.953\,{\tiny$\pm$0.0004} & \cellcolor{2nd}0.993\,{\tiny$\pm$0.0000} & 0.898\,{\tiny$\pm$0.0014} & 0.967\,{\tiny$\pm$0.0004} \\
Neural & \cellcolor{2nd}0.988\,{\tiny$\pm$0.0001} & \cellcolor{3rd}0.955\,{\tiny$\pm$0.0003} & \cellcolor{1st}0.987\,{\tiny$\pm$0.0001} & \cellcolor{1st}0.987\,{\tiny$\pm$0.0002} & \cellcolor{1st}0.984\,{\tiny$\pm$0.0001} & \cellcolor{3rd}0.963\,{\tiny$\pm$0.0001} & \cellcolor{1st}0.994\,{\tiny$\pm$0.0001} & \cellcolor{3rd}0.899\,{\tiny$\pm$0.0024} & \cellcolor{2nd}0.970\,{\tiny$\pm$0.0003} \\
\bottomrule
\end{tabular}
\begin{tabular}{lcccccccc|c}
\toprule
\multicolumn{10}{c}{LPIPS$\downarrow$ on NeRF Synthetic~\cite{mildenhall2020nerf}\vspace{2pt}}\\
Method & \textit{Chair} & \textit{Drums} & \textit{Ficus} & \textit{Hotdog} & \textit{Lego} & \textit{Materials} & \textit{Mic} & \textit{Ship} & \textit{Average} \\
\midrule
None & 0.020\,{\tiny$\pm$0.0000} & 0.052\,{\tiny$\pm$0.0001} & \cellcolor{3rd}0.021\,{\tiny$\pm$0.0001} & 0.042\,{\tiny$\pm$0.0022} & \cellcolor{3rd}0.023\,{\tiny$\pm$0.0001} & 0.060\,{\tiny$\pm$0.0005} & 0.014\,{\tiny$\pm$0.0001} & 0.148\,{\tiny$\pm$0.0020} & 0.048\,{\tiny$\pm$0.0006} \\
SH & \cellcolor{2nd}0.014\,{\tiny$\pm$0.0000} & \cellcolor{2nd}0.043\,{\tiny$\pm$0.0002} & \cellcolor{1st}0.014\,{\tiny$\pm$0.0010} & \cellcolor{1st}0.021\,{\tiny$\pm$0.0003} & \cellcolor{1st}0.019\,{\tiny$\pm$0.0000} & \cellcolor{2nd}0.042\,{\tiny$\pm$0.0001} & \cellcolor{2nd}0.008\,{\tiny$\pm$0.0001} & \cellcolor{2nd}0.130\,{\tiny$\pm$0.0006} & \cellcolor{1st}0.036\,{\tiny$\pm$0.0001} \\
SV & \cellcolor{1st}0.013\,{\tiny$\pm$0.0000} & \cellcolor{1st}0.038\,{\tiny$\pm$0.0001} & 0.034\,{\tiny$\pm$0.0095} & 0.029\,{\tiny$\pm$0.0192} & \cellcolor{1st}0.019\,{\tiny$\pm$0.0000} & \cellcolor{1st}0.041\,{\tiny$\pm$0.0001} & \cellcolor{1st}0.007\,{\tiny$\pm$0.0000} & \cellcolor{1st}0.126\,{\tiny$\pm$0.0011} & \cellcolor{2nd}0.038\,{\tiny$\pm$0.0028} \\
NASG & 0.017\,{\tiny$\pm$0.0000} & \cellcolor{3rd}0.046\,{\tiny$\pm$0.0002} & \cellcolor{2nd}0.018\,{\tiny$\pm$0.0000} & 0.029\,{\tiny$\pm$0.0048} & \cellcolor{2nd}0.021\,{\tiny$\pm$0.0001} & 0.055\,{\tiny$\pm$0.0004} & 0.018\,{\tiny$\pm$0.0191} & 0.137\,{\tiny$\pm$0.0009} & 0.043\,{\tiny$\pm$0.0024} \\
NASGabor & 0.017\,{\tiny$\pm$0.0000} & \cellcolor{3rd}0.046\,{\tiny$\pm$0.0001} & \cellcolor{2nd}0.018\,{\tiny$\pm$0.0001} & \cellcolor{3rd}0.026\,{\tiny$\pm$0.0024} & \cellcolor{2nd}0.021\,{\tiny$\pm$0.0000} & 0.055\,{\tiny$\pm$0.0005} & \cellcolor{3rd}0.009\,{\tiny$\pm$0.0001} & 0.136\,{\tiny$\pm$0.0006} & \cellcolor{3rd}0.041\,{\tiny$\pm$0.0003} \\
Neural & \cellcolor{3rd}0.015\,{\tiny$\pm$0.0001} & \cellcolor{2nd}0.043\,{\tiny$\pm$0.0001} & \cellcolor{1st}0.014\,{\tiny$\pm$0.0001} & \cellcolor{2nd}0.022\,{\tiny$\pm$0.0002} & \cellcolor{2nd}0.021\,{\tiny$\pm$0.0001} & \cellcolor{3rd}0.045\,{\tiny$\pm$0.0002} & \cellcolor{2nd}0.008\,{\tiny$\pm$0.0001} & \cellcolor{3rd}0.135\,{\tiny$\pm$0.0009} & \cellcolor{2nd}0.038\,{\tiny$\pm$0.0001} \\
\bottomrule
\end{tabular}
\begin{tabular}{lcccccccc|c}
\toprule
\multicolumn{10}{c}{\FLIP$\downarrow$ on NeRF Synthetic~\cite{mildenhall2020nerf}\vspace{2pt}}\\
Method & \textit{Chair} & \textit{Drums} & \textit{Ficus} & \textit{Hotdog} & \textit{Lego} & \textit{Materials} & \textit{Mic} & \textit{Ship} & \textit{Average} \\
\midrule
None & 0.032\,{\tiny$\pm$0.0004} & 0.064\,{\tiny$\pm$0.0005} & 0.039\,{\tiny$\pm$0.0005} & 0.039\,{\tiny$\pm$0.0039} & 0.032\,{\tiny$\pm$0.0000} & 0.053\,{\tiny$\pm$0.0004} & 0.018\,{\tiny$\pm$0.0000} & 0.072\,{\tiny$\pm$0.0016} & 0.044\,{\tiny$\pm$0.0007} \\
SH & \cellcolor{2nd}0.023\,{\tiny$\pm$0.0000} & \cellcolor{2nd}0.057\,{\tiny$\pm$0.0005} & \cellcolor{1st}0.027\,{\tiny$\pm$0.0009} & \cellcolor{1st}0.023\,{\tiny$\pm$0.0000} & \cellcolor{1st}0.028\,{\tiny$\pm$0.0000} & \cellcolor{2nd}0.035\,{\tiny$\pm$0.0000} & \cellcolor{1st}0.011\,{\tiny$\pm$0.0000} & \cellcolor{2nd}0.055\,{\tiny$\pm$0.0019} & \cellcolor{1st}0.032\,{\tiny$\pm$0.0003} \\
SV & \cellcolor{1st}0.022\,{\tiny$\pm$0.0000} & \cellcolor{1st}0.052\,{\tiny$\pm$0.0000} & 0.040\,{\tiny$\pm$0.0063} & 0.030\,{\tiny$\pm$0.0177} & \cellcolor{2nd}0.029\,{\tiny$\pm$0.0005} & \cellcolor{1st}0.034\,{\tiny$\pm$0.0000} & \cellcolor{1st}0.011\,{\tiny$\pm$0.0000} & \cellcolor{1st}0.052\,{\tiny$\pm$0.0005} & \cellcolor{2nd}0.034\,{\tiny$\pm$0.0024} \\
NASG & 0.027\,{\tiny$\pm$0.0000} & \cellcolor{2nd}0.057\,{\tiny$\pm$0.0004} & \cellcolor{3rd}0.034\,{\tiny$\pm$0.0004} & \cellcolor{3rd}0.028\,{\tiny$\pm$0.0046} & \cellcolor{3rd}0.030\,{\tiny$\pm$0.0000} & 0.047\,{\tiny$\pm$0.0005} & 0.017\,{\tiny$\pm$0.0080} & 0.060\,{\tiny$\pm$0.0015} & 0.038\,{\tiny$\pm$0.0011} \\
NASGabor & 0.027\,{\tiny$\pm$0.0000} & \cellcolor{3rd}0.058\,{\tiny$\pm$0.0000} & \cellcolor{3rd}0.034\,{\tiny$\pm$0.0005} & \cellcolor{2nd}0.026\,{\tiny$\pm$0.0016} & \cellcolor{3rd}0.030\,{\tiny$\pm$0.0000} & 0.047\,{\tiny$\pm$0.0007} & \cellcolor{3rd}0.013\,{\tiny$\pm$0.0000} & 0.060\,{\tiny$\pm$0.0013} & \cellcolor{3rd}0.037\,{\tiny$\pm$0.0003} \\
Neural & \cellcolor{3rd}0.026\,{\tiny$\pm$0.0004} & \cellcolor{3rd}0.058\,{\tiny$\pm$0.0005} & \cellcolor{2nd}0.028\,{\tiny$\pm$0.0000} & \cellcolor{1st}0.023\,{\tiny$\pm$0.0000} & \cellcolor{3rd}0.030\,{\tiny$\pm$0.0000} & \cellcolor{3rd}0.038\,{\tiny$\pm$0.0000} & \cellcolor{2nd}0.012\,{\tiny$\pm$0.0000} & \cellcolor{3rd}0.058\,{\tiny$\pm$0.0012} & \cellcolor{2nd}0.034\,{\tiny$\pm$0.0001} \\
\bottomrule
\end{tabular}
\begin{tabular}{lcccccccc|c}
\toprule
\multicolumn{10}{c}{DISTS$\downarrow$ on NeRF Synthetic~\cite{mildenhall2020nerf}\vspace{2pt}}\\
Method & \textit{Chair} & \textit{Drums} & \textit{Ficus} & \textit{Hotdog} & \textit{Lego} & \textit{Materials} & \textit{Mic} & \textit{Ship} & \textit{Average} \\
\midrule
None & 0.038\,{\tiny$\pm$0.0005} & 0.075\,{\tiny$\pm$0.0005} & 0.048\,{\tiny$\pm$0.0004} & 0.065\,{\tiny$\pm$0.0114} & 0.034\,{\tiny$\pm$0.0000} & 0.095\,{\tiny$\pm$0.0008} & 0.034\,{\tiny$\pm$0.0004} & 0.109\,{\tiny$\pm$0.0103} & 0.062\,{\tiny$\pm$0.0027} \\
SH & \cellcolor{2nd}0.032\,{\tiny$\pm$0.0004} & \cellcolor{2nd}0.057\,{\tiny$\pm$0.0004} & \cellcolor{1st}0.030\,{\tiny$\pm$0.0011} & \cellcolor{1st}0.039\,{\tiny$\pm$0.0000} & \cellcolor{2nd}0.030\,{\tiny$\pm$0.0000} & \cellcolor{2nd}0.068\,{\tiny$\pm$0.0004} & \cellcolor{2nd}0.020\,{\tiny$\pm$0.0000} & \cellcolor{2nd}0.089\,{\tiny$\pm$0.0016} & \cellcolor{1st}0.046\,{\tiny$\pm$0.0002} \\
SV & \cellcolor{1st}0.027\,{\tiny$\pm$0.0000} & \cellcolor{1st}0.047\,{\tiny$\pm$0.0004} & 0.096\,{\tiny$\pm$0.0271} & \cellcolor{3rd}0.045\,{\tiny$\pm$0.0195} & \cellcolor{1st}0.028\,{\tiny$\pm$0.0000} & \cellcolor{1st}0.063\,{\tiny$\pm$0.0004} & \cellcolor{1st}0.019\,{\tiny$\pm$0.0000} & \cellcolor{1st}0.084\,{\tiny$\pm$0.0004} & \cellcolor{3rd}0.051\,{\tiny$\pm$0.0045} \\
NASG & \cellcolor{3rd}0.037\,{\tiny$\pm$0.0004} & 0.065\,{\tiny$\pm$0.0004} & \cellcolor{3rd}0.037\,{\tiny$\pm$0.0005} & 0.050\,{\tiny$\pm$0.0077} & \cellcolor{3rd}0.032\,{\tiny$\pm$0.0000} & 0.088\,{\tiny$\pm$0.0009} & 0.045\,{\tiny$\pm$0.0483} & 0.095\,{\tiny$\pm$0.0016} & 0.056\,{\tiny$\pm$0.0060} \\
NASGabor & \cellcolor{3rd}0.037\,{\tiny$\pm$0.0000} & 0.066\,{\tiny$\pm$0.0004} & 0.038\,{\tiny$\pm$0.0004} & 0.046\,{\tiny$\pm$0.0022} & \cellcolor{3rd}0.032\,{\tiny$\pm$0.0000} & 0.088\,{\tiny$\pm$0.0008} & \cellcolor{3rd}0.023\,{\tiny$\pm$0.0000} & 0.094\,{\tiny$\pm$0.0012} & 0.053\,{\tiny$\pm$0.0004} \\
Neural & \cellcolor{2nd}0.032\,{\tiny$\pm$0.0000} & \cellcolor{3rd}0.059\,{\tiny$\pm$0.0005} & \cellcolor{2nd}0.031\,{\tiny$\pm$0.0004} & \cellcolor{2nd}0.040\,{\tiny$\pm$0.0004} & \cellcolor{2nd}0.030\,{\tiny$\pm$0.0005} & \cellcolor{3rd}0.074\,{\tiny$\pm$0.0005} & \cellcolor{2nd}0.020\,{\tiny$\pm$0.0000} & \cellcolor{3rd}0.093\,{\tiny$\pm$0.0012} & \cellcolor{2nd}0.047\,{\tiny$\pm$0.0002} \\
\bottomrule
\end{tabular}
\begin{tabular}{lcccccccc|c}
\toprule
\multicolumn{10}{c}{MILO$\uparrow$ on NeRF Synthetic~\cite{mildenhall2020nerf}\vspace{2pt}}\\
Method & \textit{Chair} & \textit{Drums} & \textit{Ficus} & \textit{Hotdog} & \textit{Lego} & \textit{Materials} & \textit{Mic} & \textit{Ship} & \textit{Average} \\
\midrule
None & 4.133\,{\tiny$\pm$0.0013} & 3.311\,{\tiny$\pm$0.0020} & 4.150\,{\tiny$\pm$0.0023} & 3.928\,{\tiny$\pm$0.0246} & 4.183\,{\tiny$\pm$0.0011} & 3.277\,{\tiny$\pm$0.0025} & 4.179\,{\tiny$\pm$0.0028} & 3.378\,{\tiny$\pm$0.0194} & 3.817\,{\tiny$\pm$0.0058} \\
SH & \cellcolor{2nd}4.255\,{\tiny$\pm$0.0019} & \cellcolor{3rd}3.461\,{\tiny$\pm$0.0022} & \cellcolor{2nd}4.262\,{\tiny$\pm$0.0147} & \cellcolor{1st}4.205\,{\tiny$\pm$0.0019} & \cellcolor{2nd}4.217\,{\tiny$\pm$0.0009} & \cellcolor{2nd}3.619\,{\tiny$\pm$0.0014} & \cellcolor{2nd}4.288\,{\tiny$\pm$0.0004} & \cellcolor{2nd}3.636\,{\tiny$\pm$0.0090} & \cellcolor{1st}3.993\,{\tiny$\pm$0.0017} \\
SV & \cellcolor{1st}4.260\,{\tiny$\pm$0.0004} & \cellcolor{1st}3.574\,{\tiny$\pm$0.0044} & 3.664\,{\tiny$\pm$0.2948} & 4.081\,{\tiny$\pm$0.2748} & \cellcolor{1st}4.218\,{\tiny$\pm$0.0010} & \cellcolor{1st}3.667\,{\tiny$\pm$0.0048} & \cellcolor{1st}4.292\,{\tiny$\pm$0.0013} & \cellcolor{1st}3.673\,{\tiny$\pm$0.0062} & \cellcolor{3rd}3.929\,{\tiny$\pm$0.0556} \\
NASG & 4.214\,{\tiny$\pm$0.0008} & 3.404\,{\tiny$\pm$0.0061} & \cellcolor{3rd}4.214\,{\tiny$\pm$0.0011} & 4.114\,{\tiny$\pm$0.0648} & 4.208\,{\tiny$\pm$0.0005} & 3.328\,{\tiny$\pm$0.0038} & 4.068\,{\tiny$\pm$0.4383} & 3.566\,{\tiny$\pm$0.0110} & 3.890\,{\tiny$\pm$0.0533} \\
NASGabor & 4.214\,{\tiny$\pm$0.0029} & 3.408\,{\tiny$\pm$0.0062} & 4.212\,{\tiny$\pm$0.0005} & \cellcolor{3rd}4.150\,{\tiny$\pm$0.0313} & 4.211\,{\tiny$\pm$0.0008} & 3.331\,{\tiny$\pm$0.0041} & 4.264\,{\tiny$\pm$0.0011} & 3.573\,{\tiny$\pm$0.0033} & 3.920\,{\tiny$\pm$0.0037} \\
Neural & \cellcolor{3rd}4.238\,{\tiny$\pm$0.0029} & \cellcolor{2nd}3.480\,{\tiny$\pm$0.0050} & \cellcolor{1st}4.268\,{\tiny$\pm$0.0008} & \cellcolor{2nd}4.204\,{\tiny$\pm$0.0023} & \cellcolor{3rd}4.212\,{\tiny$\pm$0.0011} & \cellcolor{3rd}3.610\,{\tiny$\pm$0.0031} & \cellcolor{3rd}4.287\,{\tiny$\pm$0.0021} & \cellcolor{3rd}3.603\,{\tiny$\pm$0.0048} & \cellcolor{2nd}3.988\,{\tiny$\pm$0.0017} \\
\bottomrule
\end{tabular}%
}
\caption{
Per-scene image quality metrics on NeRF Synthetic~\cite{mildenhall2020nerf}, averaged over five training runs with the standard deviation across runs.
The three best results are highlighted in \textcolor{1stText}{\textbf{green}} in descending order of saturation.
}\label{tab:per_scene_synthetic}%
\end{table*}

\cref{tab:per_scene_m360,tab:per_scene_tandt_db,tab:per_scene_synthetic} list per-scene results for all 21 scenes, averaged over five training runs together with the standard deviation across runs.
In addition to PSNR, SSIM~\cite{wang2004ssim}, and LPIPS~\cite{zhang2018lpips}, we report the more recent perceptual metrics \FLIP~\cite{andersson2020flip}, DISTS~\cite{ding2022dists}, and MILO~\cite{cogalan2025milo}.
Several patterns are hidden by the dataset averages in \cref{tab:training}.

\paragraph{Outdoor vs.\ Indoor}
On the outdoor scenes of Mip-NeRF~360, SH achieves the best PSNR on most scenes, and even the model without view-dependent appearance trails only slightly.
On the indoor scenes, the picture reverses: the expressive models clearly outperform SH, and the gap to the model without view-dependent appearance grows substantially.
View-dependent appearance thus matters primarily in indoor scenes with glossy surfaces and controlled lighting, whereas outdoor scenes are dominated by diffuse content and their remaining error is largely geometric.
This is consistent with our findings on decoder capacity in \cref{tab:neural_architecture}.
The neural model follows the expressive models indoors but ranks last among the view-dependent models in SSIM on most outdoor scenes, in line with the weaker regularization noted in \cref{tab:qualitative_analysis}.

\paragraph{Run-to-Run Variation}
On Mip-NeRF~360, dataset averages are stable to a few hundredths of a dB, whereas Tanks \& Temples and Deep Blending are noisier, in particular \emph{train} and \emph{playroom}, the two scenes with the strongest photometric variation.
Most notably, the synthetic scenes expose a robustness difference between models that the averages conceal.
Whereas SH and the neural model barely vary on every NeRF Synthetic scene, SV fluctuates by several dB on \emph{ficus} and \emph{hotdog}, and NASG on \emph{mic}.
In these cases, individual runs settle into noticeably worse local minima, which is what \cref{tab:qualitative_analysis} summarizes as robustness.
The band-limited SH basis is the most forgiving representation, whereas sharper angular primitives such as Voronoi cells and anisotropic lobes can lock into poor configurations early in the optimization from which gradient descent does not recover.

\paragraph{Perceptual Metrics}
\FLIP, DISTS, and MILO largely agree with the established metrics and do not change the conclusions of the main paper.
DISTS slightly favors SH and MILO the spherical models on real scenes, whereas \FLIP, like PSNR, favors the neural model on Tanks \& Temples and Deep Blending.
Across all six metrics and three datasets, no model is consistently superior and the differences remain within a narrow band.
This supports the conclusion of the main paper that memory footprint and rendering performance are the more discriminative axes when choosing an appearance model.

\end{document}